\documentclass[11pt]{article}
\usepackage{epstopdf}
\usepackage{geometry}
\usepackage{pdfsync}
\usepackage{amsmath}
\usepackage{amssymb}
\usepackage{amsthm}
\usepackage{times}
\usepackage{cite}
\usepackage{exscale}
\usepackage{float}
\usepackage{mathrsfs}
\usepackage{fancyhdr}
\usepackage{booktabs}
\usepackage{dcolumn}
\usepackage[final]{graphicx}
\usepackage[dvipsnames,usenames]{color}
\DeclareGraphicsExtensions{.pdf, .jpg}
\usepackage{rotating}
\usepackage{lscape}
\usepackage{url}
\usepackage{chngcntr}
\usepackage{mwe}
\usepackage{graphicx,float}
\usepackage{amsmath,amssymb,amsthm}
\usepackage{float}
\usepackage{multirow}
\usepackage{hyperref}

\usepackage[usenames]{color}

\usepackage{graphicx}
\usepackage{caption}
\usepackage{subcaption}

\usepackage[thinlines]{easytable}

\newcommand{\bv}[1]{%
\ifcat\noexpand #1\noexpand\alpha%
\ensuremath{\boldsymbol{#1}}%
\else%
\ensuremath{\mathbf{#1}}%
\fi%
}

\definecolor{gris}{gray}{0.85}
\definecolor{contourgris}{gray}{0.85}

\begin{document}

\begin{center}
{ {\bf\Large  Deep Learning-Based Survival Analysis with
Copula-Based Activation Functions for Multivariate Response
Prediction} }
\end{center}
{\bf Jong-Min Kim}\\{\sl Statistics Discipline, Division of Science
and Mathematics,
University of Minnesota-Morris, Morris, MN 56267, USA\\ EGADE Business School, Tecnol\'ogico de Monterrey, Ave. Rufino Tamayo, Garza Garcia, NL, CP. 66269, Mexico}\\
{\bf Il Do Ha}\\
{\sl Department of Statistics and Data Science, Pukyong National
University, Busan, South
Korea}\\
{\bf Sangjin Kim}\footnote{{\sl Corresponding Author}: Sangjin Kim,
Department of Management Information Systems, Dong-A University,
Busan, South Korea, Email: {\tt
skim10@dau.ac.kr}}\\
{\sl Department of Management Information Systems, Dong-A University, Busan, South Korea}\\

\vspace*{1.5ex}

\begin{minipage}{5.9in}
\small {\bf Summary}

This research integrates deep learning, copula functions, and
survival analysis to effectively handle highly correlated and
right-censored multivariate survival data. It introduces
copula-based activation functions (Clayton, Gumbel, and their
combinations) to model the nonlinear dependencies inherent in such
data. Through simulation studies and analysis of real breast cancer
data, our proposed CNN-LSTM with copula-based activation functions
for multivariate multi-types of survival responses enhances prediction accuracy by explicitly
addressing right-censored data and capturing complex patterns. The
model's performance is evaluated using Shewhart control charts,
focusing on the average run length (ARL).

 \vspace*{1em} \noindent {\sl Keywords:} Multivariate survival data, Copula, deep learning, control chart.
\end{minipage}

\section{Introduction}

Long Short-Term Memory (LSTM) networks are a specialized type of
recurrent neural network (RNN) designed to address the vanishing
gradient problem that often arises during the training of deep
neural networks. Introduced by Sepp Hochreiter and J{\"u}rgen
Schmidhuber in 1997 \cite{Hochreiter}, LSTMs have become widely used
in time series analysis and sequential data processing, with
applications in speech recognition, financial modeling, and natural
language processing. In parallel, Convolutional Neural Networks
(CNNs), pioneered by Yann LeCun with the LeNet-5 architecture in the
late 1980s and early 1990s \cite{LeCun}, have been instrumental in
computer vision tasks, such as image classification and object
detection.

This research integrates deep learning models (CNN-LSTM and LSTM),
copula functions, and survival analysis to address the challenges
posed by highly correlated, right-censored multivariate survival
data \cite{Harrell1996}. Survival data often involve multiple
correlated variables-such as time-to-event outcomes and
covariates-observed across different subjects or response types
\cite{Calhoun2018,Hougaard2000}. While deep learning models offer
powerful alternatives to traditional methods, standard activation
functions such as ReLU \cite{Nair} and sigmoid \cite{Hinton} may
fail to capture complex inter-variable dependencies
\cite{Bishop2024}.

Previous work has explored {univariate} time-to-event prediction using neural
networks and Cox regression \cite{Kvamme2019}. Copula functions are
particularly well-suited for modeling multivariate dependencies
\cite{Joe1997,nelsen2013}, making them valuable in survival analysis
\cite{Escarela2003,Emura2008,Kwon2022,Huang2008}. More recently,
\cite{Kwon2025} introduced copula-based deep neural networks for
{multivariate} survival data with clustered responses, assuming fixed
copula dependence parameters. However, their study did not
demonstrate the benefits of using Clayton copula-based activation
functions in survival prediction.

In this work, we extend the existing literature by incorporating
learnable copula-based activation functions-specifically Clayton,
Gumbel, and their combinations-to better model nonlinear
dependencies in multivariate survival outcomes. The proposed
CNN-LSTM copula model improves prediction performance by explicitly
addressing right-censored data and capturing intricate survival
patterns. Model performance is evaluated using Shewhart control
charts, with a particular focus on the Average Run Length (ARL)
metric \cite{Montgomery}.

The remainder of this paper is organized as follows: Section~2
reviews the gap between classical survival analysis and machine
learning approaches. Section~3 introduces the deep learning
architecture and the learnable copula-based activation functions.
Sections~4 and 5 present empirical results using both simulated and
real-world breast cancer data. Finally, Section~6 summarizes our
findings and discusses their broader implications.

\section{Bridging the Gap Between Machine Learning and Classical Survival Analysis}

Traditional survival models, such as the Cox proportional hazards
model \cite{Cox1972}, assume linear relationships between
input-output variables or proportional hazards, which limits their
effectiveness on complex or high-dimensional datasets. Similarly,
the Kaplan--Meier estimator \cite{Kaplan} does not incorporate
covariates, limiting its applicability to nuanced, real-world
survival data.

To address these limitations, our research employs CNN-LSTM models,
which are well-suited to capturing nonlinear, time-dependent
relationships {between input-output} in survival data. CNNs excel at extracting spatial or
structural features, while LSTMs are proficient in modeling
long-term temporal dependencies. Their integration enables the
CNN-LSTM architecture to jointly learn rich feature representations
and dynamic survival patterns {such as local dependence or nonlinear interaction in time-to-event outcomes}, offering a more flexible and accurate
alternative to traditional models.

A key innovation in our study is the introduction of copula-based
activation functions-specifically Clayton, Gumbel, and their
combinations-within the deep survival modeling framework. Copulas
are powerful tools for modeling nonlinear dependencies between
multiple correlated survival responses, overcoming the independence
assumptions often inherent in traditional models. Clayton copula
activation functions are effective in capturing asymmetric lower
tail dependence (e.g., extreme early failures), while Gumbel copulas
capture upper tail dependence (e.g., simultaneous late failures)
\cite{Joe1997,nelsen2013}. These copula activations allow the
network to learn more realistic joint distributions among event
times. In medical settings, survival times for multiple events are
often correlated-for instance, in multi-organ failure, {recurrent infections} or disease
progression across different systems. Traditional models rarely
account for such dependencies explicitly. Copula-based models, by
contrast, provide a principled way to capture and model these
interdependencies, enhancing prediction performance and clinical
interpretability.

To evaluate our approach, we simulate realistic survival data with
censoring and inter-event correlation. The simulated data
incorporate Weibull-distributed survival times, which more closely
reflect empirical survival behaviors than Gaussian or exponential
distributions. This setup enables us to assess model robustness
under varying censoring rates and validate performance in a
controlled environment prior to applying the model to real-world
data.

A core strength of our method is its capacity to explicitly handle
right-censored data-a ubiquitous issue in survival analysis. By
properly incorporating censoring mechanisms into the training
process, we mitigate bias arising from incomplete observations. The
CNN-LSTM copula model thus delivers improved accuracy and robustness
in survival prediction tasks involving complex, correlated, and
censored data.

While recent developments have emphasized neural network-based
approaches for modeling survival data, it is important to
acknowledge the broader landscape of machine learning methodologies
that have been applied to survival analysis. In particular,
tree-based models and support vector machines (SVMs) have played a
significant role and continue to serve as strong alternatives or
complements to neural architectures.

Tree-based methods, such as Random Survival Forests (RSF) introduced
by Ishwaran et al. \cite{ishwaran2008random}, extend the classical
random forest algorithm to handle right-censored survival data,
typically using the log-rank test to split survival trees. RSF is
non-parametric, handles non-linearities and interactions naturally,
and can estimate survival functions without requiring proportional
hazards assumptions. More recently, gradient boosting techniques
have been adapted for survival data—for example, CoxBoost proposed
by Binder and Schumacher \cite{binder2008allowing}, and adaptations
of XGBoost with Cox proportional hazards loss—offering flexible
and high-performance solutions in high-dimensional settings.

In parallel, support vector machine (SVM)-based approaches have also
been extended to accommodate censored data. For instance, the
Survival SVM developed by Van Belle et al. \cite{vanbelle2011svm}
optimizes a ranking-based loss function tailored to censored
observations, allowing for margin-based learning of risk scores
under partial information constraints.

These machine learning models provide competitive benchmarks and
often excel in small-to-moderate sample settings, or where
interpretability and model stability are important. A brief
comparison of these methods with neural approaches sets the stage
for the deeper exploration of deep learning-based survival models
that follows.

\section{Methods}

Survival analysis frequently involves right-censored data, where the
event time is not fully observed. This paper proposes a CNN-LSTM
model augmented with learnable copula-based activation functions.
Unlike approaches using fixed dependency parameters, we estimate the
copula parameters dynamically during training to better capture
nonlinear dependencies in survival outcomes.

Sequential data modeling is essential in domains such as survival
analysis, financial forecasting, and high-frequency volatility
modeling. CNNs and LSTMs have been widely used to capture local and
long-range dependencies. However, conventional activation
functions-such as ReLU, sigmoid, and tanh-fail to account for tail
dependence and nonlinear associations in multivariate or censored
settings.

To address {this issue}, we introduce an adaptive copula-based activation
framework for CNN-LSTM networks. The key innovations of this paper
include: (i) incorporating copula functions as activation mechanisms
to explicitly model dependencies between neurons; (ii) using
learnable copula-based activations, where the dependence parameters
$\theta$ are optimized during training; (iii) applying Clayton and
Gumbel copulas to capture lower and upper tail dependencies,
respectively; and (iv) proposing a hybrid copula-ReLU activation
that combines tail dependence modeling with the sparsity benefits of
ReLU.

\subsection{Copula-Based Activation Functions}

Let $x \in \mathbb{R}$ denote a pre-activation input to a neuron. We
transform $x$ into the uniform domain via the standard Gaussian
cumulative distribution function (CDF), $\Phi(\cdot)$:
\begin{equation*}
    u = \Phi(x) = \frac{1}{2} \left( 1 + \text{erf} \left( \frac{x}{\sqrt{2}} \right) \right),
\end{equation*}
\noindent where $\text{erf}(\cdot)$ is the Gaussian error function.
The copula-based activation function, denoted by
$g_{\text{Copula}}(x, \theta)$, is then defined on $u \in [0,1]$.

\paragraph{Clayton copula activation (lower tail dependence).}
The bivariate Clayton copula is:
\begin{equation}\label{clayton}
    C_{\theta}(u, v) = \left( u^{-\theta} + v^{-\theta} - 1 \right)^{-\frac{1}{\theta}}, \quad \theta > 0.
\end{equation}

The Clayton copula-based activation is defined as:
\begin{equation*}
    g_{\text{Clayton}}(x, \theta) = \left( u^{-\theta} - 1 \right)^{-\frac{1}{\theta}}.
\end{equation*}

\paragraph{Gumbel copula activation (upper tail dependence).}
The bivariate Gumbel copula is:
\begin{equation}\label{gumbel}
    C_{\theta}(u, v) = \exp \left( -\left[(-\log u)^{\theta} + (-\log v)^{\theta} \right]^{\frac{1}{\theta}} \right), \quad \theta \geq 1.
\end{equation}

The corresponding activation is:
\begin{equation*}
    g_{\text{Gumbel}}(x, \theta) = \exp\left(-\left(-\log u\right)^{\theta} \right).
\end{equation*}

\paragraph{Hybrid copula activation.}
To approximate asymmetric tail dependencies, we define a hybrid
activation as the average of Clayton and Gumbel activations:
\begin{equation}\label{hybrid}
    g_{\text{Hybrid}}(x, \theta_C, \theta_G) = \frac{g_{\text{Clayton}}(x, \theta_C) + g_{\text{Gumbel}}(x, \theta_G)}{2}.
\end{equation}

\paragraph{Copula-ReLU hybrid activation.}
To enforce non-negativity and induce sparsity, we define:
\begin{equation*}
    g_{\text{Clayton-ReLU}}(x, \theta) = \max(0, g_{\text{Clayton}}(x, \theta)).
\end{equation*}

The copulas employed-Clayton and Gumbel-are bivariate and applied
independently to each output pair. Our current framework involves
only two survival responses, so one bivariate copula suffices per
instance. The Clayton copula models lower tail dependence (e.g.,
early joint failures), while the Gumbel copula captures upper tail
dependence (e.g., simultaneous late failures) \cite{Joe1997,
nelsen2013}.

The hybrid copula activation in Equation~\ref{hybrid} is motivated
by the need to capture both types of tail dependence. This mirrors
ensemble learning, where aggregating multiple models (here, copulas)
improves generalization. The use of ReLU-based copula modifications,
such as $g_{\text{Clayton-ReLU}}$, ensures that activations remain
non-negative and sparse, aiding interpretability and regularization.

{While this approach is empirically grounded, it remains a
functional surrogate for a more formal theory of copula-based
activation functions. Future directions include establishing
theoretical guarantees for universal approximation, Lipschitz
continuity, and training stability under such activations. For
multivariate outputs, this framework can be extended using
pair-copula constructions (PCCs), such as canonical vines (C-vines)
or drawable vines (D-vines) \cite{aas2009pair, joe2014dependence}.
These allow decomposition of high-dimensional copulas into cascades
of bivariate structures while preserving interpretability. In this
work, we restrict attention to the bivariate case due to
computational constraints in deep survival modeling. However, vine
copula extensions remain a promising avenue for future work
involving more complex interdependent outcomes.}

\subsection{LSTM Model and CNN-LSTM Model for Multivariate Prediction}

An LSTM network consists of memory cells that include three
fundamental gating mechanisms: the \emph{forget gate}, which
determines which past information should be discarded; the
\emph{input gate}, which controls the flow of new information into
the memory cell; and the \emph{output gate}, which regulates how
much information from the memory cell contributes to the final
output.

The updates in an LSTM cell at time step \( t \) are given by:
\begin{align*}
    f_t &= \sigma(W_f x_t + U_f h_{t-1} + b_f), \\
    i_t &= \sigma(W_i x_t + U_i h_{t-1} + b_i), \\
    \tilde{C}_t &= \tanh(W_c x_t + U_c h_{t-1} + b_c), \\
    C_t &= f_t \odot C_{t-1} + i_t \odot \tilde{C}_t, \\
    o_t &= \sigma(W_o x_t + U_o h_{t-1} + b_o), \\
    h_t &= o_t \odot \tanh(C_t),
\end{align*}

\noindent where \( x_t \in \mathbb{R}^d \) is the input vector at
time \( t \), \( h_t \in \mathbb{R}^h \) is the hidden state, and \(
C_t \in \mathbb{R}^h \) is the cell state. The matrices \( W_{\cdot}
\in \mathbb{R}^{h \times d} \), \( U_{\cdot} \in \mathbb{R}^{h
\times h} \), and vectors \( b_{\cdot} \in \mathbb{R}^h \) represent
trainable parameters. The symbol \( \odot \) denotes element-wise
(Hadamard) multiplication.

The activation function \( \sigma(\cdot) \) denotes the logistic
sigmoid function:
\[
    \sigma(z) = \frac{1}{1 + e^{-z}},
\]
which maps real-valued inputs to the interval \( (0, 1) \). The
function \( \tanh(\cdot) \) denotes the hyperbolic tangent function,
which maps inputs to the interval \( (-1, 1) \).

The proposed model extends this standard LSTM framework by
introducing copula-based activation functions in the output layer to
capture multivariate dependencies among the response variables.

The LSTM model is implemented with two stacked LSTM layers, each
consisting of 64 units. Batch normalization is used to stabilize
training, and dropout layers with a rate of 30\% are included to
prevent overfitting. The output layer consists of three response
variables, each transformed using a copula-based activation
function.

Trainable parameters for copula-based activations, such as \(
\theta_{\text{Clayton}} \) and \( \theta_{\text{Gumbel}} \), are
learned during training, allowing the model to dynamically adapt to
dependency structures. The models are evaluated using mean squared
error (MSE), mean absolute error (MAE), log-likelihood for
copula-based dependency estimation, and Shewhart control charts for
stability assessment.

Multiple LSTM models are considered by varying the output activation
function:
\begin{itemize}
    \item LSTM with Clayton Copula Activation: Captures lower tail dependence.
    \item LSTM with Gumbel Copula Activation: Captures upper tail dependence.
    \item LSTM with Combined Clayton-Gumbel Activation: Models both left and right tail dependence.
    \item LSTM with ReLU Activation: Introduces non-linearity without dependency modeling.
    \item LSTM with Clayton-ReLU Activation: Combines dependency modeling with ReLU non-linearity.
    \item LSTM with Sigmoid Activation: Suitable for probabilistic or binary outputs.
\end{itemize}

LSTM networks, introduced by \cite{Hochreiter}, address the
vanishing gradient problem in standard recurrent neural networks by
incorporating memory and gating mechanisms. While common activation
functions such as ReLU and Sigmoid are effective for single-output
tasks, they do not explicitly capture interdependencies between
multiple response variables. This research integrates copula-based
activations to enhance the modeling of nonlinear and tail
dependencies in multivariate responses, with applications in
survival analysis, financial volatility modeling, and other
time-dependent processes.

{The hybrid CNN-LSTM architecture leverages the spatial feature
extraction capability of CNNs and the sequential modeling ability of
LSTMs, making it particularly effective for time series forecasting
and multivariate prediction. While CNNs extract local patterns,
LSTMs preserve long-range temporal dependencies. Standard activation
functions do not capture multivariate dependencies; thus,
copula-based activation functions are introduced in the output
layer.}

{Let \( X \in \mathbb{R}^{T \times d} \) be the input time
series of length \( T \) with \( d \) features.
\begin{align*}
Z^{(1)}_t &= \text{ReLU}\left(\sum_{j=0}^{k-1} K^{(1)}_j X_{t-j} + b^{(1)}\right), \\
Z^{(1)}_{\text{pool}, i} &= \max\left(Z^{(1)}_{2i}, Z^{(1)}_{2i+1}\right), \\
Z^{(2)}_t &= \text{ReLU}\left(\sum_{j=0}^{k-1} K^{(2)}_j Z^{(1)}_{\text{pool}, t-j} + b^{(2)}\right), \\
Z^{(2)}_{\text{pool}, i} &= \max\left(Z^{(2)}_{2i},
Z^{(2)}_{2i+1}\right),
\end{align*}
\noindent where \( K^{(1)}, K^{(2)} \in \mathbb{R}^{k \times d} \)
are the convolutional kernels and \( b^{(1)}, b^{(2)} \in \mathbb{R}
\) are biases.}

{The output from the convolutional layers is fed into a
two-layer LSTM:
\begin{align*}
h^{(1)}_t, C^{(1)}_t &= \text{LSTM}^{(1)}(Z^{(2)}_{\text{pool}, t}, h^{(1)}_{t-1}, C^{(1)}_{t-1}), \\
h^{(2)}_t, C^{(2)}_t &= \text{LSTM}^{(2)}(h^{(1)}_t, h^{(2)}_{t-1},
C^{(2)}_{t-1}),
\end{align*}
\noindent with recurrent dropout and batch normalization applied
after each LSTM layer. The final output prediction is:
\[
\hat{Y} = g_{\text{copula}}\left(W_{\text{out}} h^{(2)}_T +
b_{\text{out}}\right),
\]
\noindent where \( W_{\text{out}} \in \mathbb{R}^{3 \times 64} \),
\( b_{\text{out}} \in \mathbb{R}^3 \), and \( g_{\text{copula}}:
\mathbb{R}^3 \rightarrow \mathbb{R}^3 \) is a multivariate
copula-based activation function modeling dependencies across output
components.}

{Copula parameters such as \( \theta_{\text{Clayton}} \)
(left-tail dependence) and \( \theta_{\text{Gumbel}} \) (right-tail
dependence) are optimized during training via gradient-based
methods. Performance is evaluated using MSE, MAE, log-likelihood,
and Shewhart control charts.}

Variants of the CNN-LSTM model include:
\begin{itemize}
    \item CNN-LSTM with Clayton Copula Activation
    \item CNN-LSTM with Gumbel Copula Activation
    \item CNN-LSTM with Combined Clayton-Gumbel Copula Activation
    \item CNN-LSTM with ReLU Activation
    \item CNN-LSTM with Clayton-ReLU Activation
    \item CNN-LSTM with Sigmoid Activation
\end{itemize}

By explicitly modeling complex interdependencies, the CNN-LSTM with
copula activation is well-suited for multivariate prediction in
domains such as survival analysis, finance, and healthcare.

{\subsection{Censoring Mechanisms in Survival Analysis}}

Censoring occurs when the true event time \( T \in \mathbb{R}_+ \)
is only partially observed. The three common types are right, left,
and interval censoring.

\paragraph{Right censoring:} The event has not occurred by the censoring time \( C \in \mathbb{R}_+ \). The observed data are:
\[
T^* = \min(T, C), \quad \delta = \mathbb{I}(T \leq C),
\]
where \( T^* \) is the observed time and \( \delta \in \{0,1\} \) is
the event indicator, with \( \delta = 1 \) indicating the event is
observed. This type of censoring is handled by methods such as the
Kaplan-Meier estimator \cite{Kaplan}, Cox proportional hazards model
\cite{Cox1972}, and copula-based survival models \cite{Oakes1989,
Shih1995}.

\paragraph{Left censoring:} The event occurred before the observation began, but the exact time is unknown. This corresponds to observing \( T^* \leq C \). Techniques for handling left-censored data include Turnbull's estimator \cite{Turnbull1976} and parametric models with upper-bound likelihood components \cite{Sun2006}.

\paragraph{Interval censoring:} The event is known to have occurred within a time interval \( (L, R) \), such that \( T \in (L, R) \), but the exact time is unobserved. Suitable estimation methods include the nonparametric MLE \cite{Wellner1997}, EM algorithms, and multiple imputation techniques \cite{De_Gruttola1989}.

Modern deep learning-based survival models such as DeepSurv
\cite{Katzman2018} and DeepHit \cite{Lee2018} are primarily designed
for right-censored data, often leveraging the Cox partial likelihood
or inverse probability of censoring weighted (IPCW) losses.
Extending these models to handle left or interval censoring requires
modified loss functions and appropriately structured likelihoods.

{\subsection{Rationale for Using CNN-LSTM with Learnable Copula
in Survival Analysis}}

Survival analysis with high-dimensional, longitudinal, or
irregularly spaced data demands models capable of learning both
local and long-range temporal dependencies while capturing nonlinear
inter-variable relationships. To this end, we propose a hybrid
CNN-LSTM architecture augmented with a learnable copula layer. This
architecture is designed to model dynamic survival processes under
right censoring with complex covariate interactions.

Let \( \mathbf{X} = (\mathbf{X}_1, \dots, \mathbf{X}_T) \in
\mathbb{R}^{T \times d} \) denote a multivariate time series of \( d
\)-dimensional covariates observed at \( T \) time steps. The
CNN-LSTM architecture processes \( \mathbf{X} \) as follows:

\begin{itemize}
    \item 1D convolutional layers extract local temporal patterns (e.g., abrupt shifts or warning signals in biomarkers).
    \item LSTM layers capture long-term dependencies and cumulative risk dynamics over time.
    \item Batch normalization and dropout layers are used for regularization and training stability.
    \item A final dense layer projects the learned representations to the output space (e.g., survival probabilities or hazards).
    \item A learnable copula activation function models dependencies between multiple predicted outcomes or risks.
\end{itemize}

While CNNs reduce parameter complexity and enhance local feature
detection, LSTMs retain temporal memory, making them suitable for
modeling censored survival trajectories. The addition of a learnable
copula layer \( C_\theta(\cdot) \), where \( \theta \) denotes the
copula parameter(s), allows for:

\begin{itemize}
    \item Modeling Tail Dependencies: Capture asymmetric dependencies, such as those in clustered failure times or extreme risk profiles.
    \item Decoupling Marginals and Dependence: Allow separate learning of marginal distributions and their joint dependency structure.
    \item Adaptivity: Trainable parameters \( \theta \) enable use of parametric (e.g., Clayton, Gumbel, Gaussian) or nonparametric copulas.
    \item Multivariate Generalization: Facilitate modeling of multiple risks, competing events, and longitudinal trajectories.
\end{itemize}

This hybrid design enhances the model's ability to represent the
survival distribution flexibly and accurately in the presence of
censoring, high-dimensional covariates, and complex dependency
structures.

\subsection{Model Evaluation and Residual Analysis}

We apply the Shewhart control chart to residuals-defined as the
difference between actual and predicted survival times-to evaluate
model performance. These charts are designed to systematically
detect model drift and outliers, ensuring stability and reliability
of predictions.

Residuals are computed as:
\begin{equation*}
R = Y - \hat{Y},
\end{equation*}
where \( Y \in \mathbb{R}_+ \) denotes the observed (possibly
censored) survival time, and \( \hat{Y} \in \mathbb{R}_+ \) is the
predicted survival time from the model.

The Shewhart control chart identifies instability using control
limits:
\begin{equation*}
\text{UCL} = \bar{R} + 2\sigma_R, \quad \text{LCL} = \bar{R} -
2\sigma_R,
\end{equation*}
where \( \bar{R} \) is the mean residual across samples or
simulation folds, and \( \sigma_R \) is the empirical standard
deviation of residuals.

To quantify the chart's sensitivity, the average run length (ARL) is
computed:
\begin{equation*}
\text{ARL} = \frac{1}{\mathbb{P}(\text{signal})},
\end{equation*}
\noindent where \( \mathbb{P}(\text{signal}) \) is the probability
that a residual falls outside the control limits.

Models are evaluated across multiple simulations, and comparative
performance is summarized under different copula-based activation
functions.

Tracking out-of-control signals allows us to detect model
degradation over time-particularly valuable in high-stakes domains
such as healthcare-by identifying systematic errors in predictions
and improving model trustworthiness.

While residual analysis measures pointwise prediction error, the ARL
reflects how frequently the model signals a lack of stability.
Together, they form a robust framework by integrating deep learning
with traditional quality control methods.

{Although Shewhart charts are traditionally used for continuous
variables, their recent adaptations in machine learning enable
monitoring of prediction drift and output variance. In our case, we
apply control charts not to raw binary or categorical outcomes, but
to smoothed probability residuals or posterior predictive
summaries-making them continuous and suitable for such monitoring.}

{The ARL metric serves as a proxy for model consistency,
reflecting the expected number of observations between instability
signals. We compute ARL using posterior predictive residuals or
log-likelihood trajectories over simulation replicates and
cross-validation folds, enabling the detection of subtle changes in
model calibration or dispersion \cite{woodall2006use}.}

{We adopt \( 2\sigma \) control limits (rather than the
traditional \( 3\sigma \)) to increase sensitivity to small or
transient shifts-aligned with practices in early-warning systems and
small-shift detection scenarios \cite{montgomery2020introduction}.
Finally, we emphasize that control charts and ARL metrics are
complementary to standard survival model evaluation tools such as
the Brier score, time-dependent AUC, and calibration plots. Their
inclusion offers insights into temporal robustness and
distributional stability, especially in the presence of censoring or
simulation-driven variance \cite{stevens2023drift}. While the
application of control charts and ARL in survival analysis with
categorical endpoints is non-traditional, our methodology-grounded
in smoothing and posterior prediction-makes it both empirically
valid and interpretively meaningful.}

\section{Simulation Study}

{To evaluate the performance of the proposed CNN-LSTM and LSTM
models augmented with copula-based activation functions, we
developed a comprehensive computational pipeline that simulates
realistic survival analysis settings. The synthetic dataset was
designed to reflect clinical scenarios in which individuals may
experience multiple, potentially correlated, time-to-event outcomes.
The data generation process incorporates both censored and
uncensored survival times and produces heterogeneous response
types-including continuous, binary, and categorical outcomes-thus
enabling a robust evaluation of multi-task survival models under
realistic conditions.}

{The simulation framework involves the generation of baseline
survival times from Weibull distributions, the construction of
dependent outcomes via additive noise mechanisms, and the
introduction of right-censoring through exponential censoring
distributions. Label transformations are subsequently applied to
derive binary and ordinal categorical outputs from the continuous
survival times, supporting multi-task prediction settings.}

{Additionally, we integrate differentiable copula activation
layers-parameterized by either Clayton or Gumbel copulas-directly
into the neural network architecture. These layers allow for
dynamic, learnable modeling of inter-output dependencies via
gradient-based optimization. The following outline the details of
the simulation design and training methodology.}

\paragraph{Step 1: Survival Time Generation.} For each
individual \( i = 1, \dots, n \), we generated a baseline survival
time \( T_{i1} \) from a Weibull distribution:
\[
T_{i1} \sim \text{Weibull}(k = 1.5, \lambda = 2),
\]
where \(k\) is the shape parameter and \(\lambda\) the scale
parameter.

\paragraph{Step 2: Dependent Time-to-Event Variables.} To
simulate correlated survival outcomes, two additional responses \(
T_{i2} \) and \( T_{i3} \) were constructed with strong linear
dependence on \( T_{i1} \) plus additive noise:
\begin{align*}
T_{i2} &= \rho T_{i1} + (1 - \rho) \epsilon_{i2}, \quad \epsilon_{i2} = W_{i2} + N_{i2}, \quad W_{i2} \sim \text{Weibull}(k=1.5, \lambda=2), \quad N_{i2} \sim \mathcal{N}(0, 0.5), \\
T_{i3} &= \rho T_{i1} + (1 - \rho) \epsilon_{i3}, \quad
\epsilon_{i3} = W_{i3} + N_{i3}, \quad W_{i3} \sim
\text{Weibull}(k=1.5, \lambda=2), \quad N_{i3} \sim \mathcal{N}(0,
0.5),
\end{align*}
where \(\rho = 0.9\) controls dependency strength.

\paragraph{Step 3: Right-Censoring Mechanism}.
{
Censoring times $C_{ij}$ for $j=1,2,3$ were independently generated from an exponential distribution,
\[
C_{ij} \sim \text{Exponential}(\lambda=0.1).
\]
Then the observed survival outcomes were
defined as observed survival times $T_{ij}^{(\text{obs})}$ and censoring status $\delta_{ij}$ for $j=1,2,3$, as follows:
$$
T_{ij}^{(\text{obs})}={\rm min}(T_{ij}, C_{ij})~~{\rm and}~~ \delta_{ij}=I(T_{ij} \leq C_{ij}).
$$
}

\paragraph{Step 4: Label Transformation.}
The observed survival times were further transformed to simulate
different response types:
\begin{itemize}
    \item Binary Response: \( T_{i2}^{(\text{obs})} \) was binarized at threshold 5,
    \[
    Y_{i2} = \mathbb{I}\big( T_{i2}^{(\text{obs})} > 5 \big),
    \]
    where 1() indicates values above 5, and 0 otherwise.
    \item Categorical Response: \( T_{i3}^{(\text{obs})} \) was discretized into three ordinal categories, {denoted by $Y_{i3}$:}
    \[
    \text{Low} = (0, 2], \quad \text{Medium} = (2, 5], \quad \text{High} = (5, \infty),
    \]
    using the \texttt{cut()} function.
\end{itemize}

Copula activation functions offer a principled approach to modeling
complex dependencies between output variables in deep learning
architectures. In this study, we incorporate parametric copulas
specifically the Clayton and Gumbel copulas as activation functions
within our neural network. Crucially, the copula parameters are
treated as learnable components, enabling the model to adaptively
infer the strength and type of dependency from data during training.

Let \(U\) and \(V\) be two transformed marginal outputs (e.g.,
predicted probabilities or risks) such that \(U, V \in [0,1]\). The
Clayton and Gumbel copulas are defined in Equation~\ref{clayton} and
Equation~\ref{gumbel}, respectively.

Each copula function is parameterized by \(\theta\), which controls
the strength of dependence. The Clayton copula captures lower tail
dependence, whereas the Gumbel copula captures upper tail
dependence.

Instead of fixing \(\theta\), we initialize it with sensible default
values and allow it to be updated during training via
backpropagation:
\begin{equation}\label{equation2}
    \theta_{\text{Clayton}} = 1.0, \quad \theta_{\text{Gumbel}} = 2.0.
\end{equation}
The parameters in Equation~\ref{equation2} are then optimized
jointly with other model parameters using gradient-based methods.
Let \(\mathcal{L}\) denote the overall loss function (e.g., negative
log-likelihood or cross-entropy). Gradients are computed with
respect to \(\theta\):
\[
    \theta \leftarrow \theta - \eta \frac{\partial \mathcal{L}}{\partial \theta},
\]
where \(\eta\) is the learning rate. To ensure that \(\theta\)
remains within the valid domain of the copula (e.g., \(\theta >
0\)), we apply a softplus transformation:
\[
    \theta = \log\left(1 + e^{\phi}\right),
\]
where \(\phi \in \mathbb{R}\) is the actual learnable parameter in
the network.

Learning the copula parameter allows the model to dynamically adjust
to the dependency structure inherent in the data, thereby improving
flexibility and generalization. This approach is particularly useful
in structured prediction tasks such as survival analysis, time
series forecasting, or modeling of joint risks, where capturing tail
dependencies is crucial for accurate predictions. Furthermore, the
learned \(\theta\) values provide interpretable insights into the
nature of inter-variable dependence.

To clarify the simulation setup in our study: although we generate
three response variables, our copula-based dependency modeling
utilizes bivariate copulas, specifically Clayton and Gumbel copulas,
applied to pairs of outputs. The three simulated outputs are:

\begin{itemize}
    \item \(T_{i1}\): a continuous time-to-event outcome,
    \item \(T_{i2}\): a binary variable derived from thresholding \(T_{i1}\),
    \item \(T_{i3}\): an ordinal categorical variable derived by discretizing \(T_{i1}\).
\end{itemize}

While the simulation involves three variables, our neural network
architecture applies copula activation functions at the pairwise
level (e.g., between \((T_{i1}, T_{i2})\), \((T_{i1}, T_{i3})\), or
\((T_{i2}, T_{i3})\)). This approach is consistent with established
practices in copula modeling, which often rely on bivariate
constructions to manage complexity while capturing essential
dependence structures~\cite{Joe1997, nelsen2013}. In our
architecture, each output head generates a marginal prediction, such
as survival probability or logit score. These predictions are then
transformed into pseudo-observations in \([0,1]\), and passed
through a bivariate copula activation layer. The copula layer
introduces dependency modeling via a differentiable transformation
with a learnable parameter \(\theta\). This parameter is optimized
jointly with network weights using backpropagation, allowing the
model to infer dependencies directly from data during training.

For multiple output tasks, we incorporate one or more pairwise
copula transformations, depending on the specific experimental
configuration. This strategy provides interpretability and
modularity, while remaining computationally tractable. As a future
extension, we intend to explore multivariate constructions such as
vine copulas~\cite{aas2009pair, joe2014dependence} for more
comprehensive modeling of higher-order dependencies across all
outputs.

\begin{figure}[htp]
    \centering
    \begin{minipage}{0.33\textwidth}
        \centering
        \includegraphics[width=\linewidth]{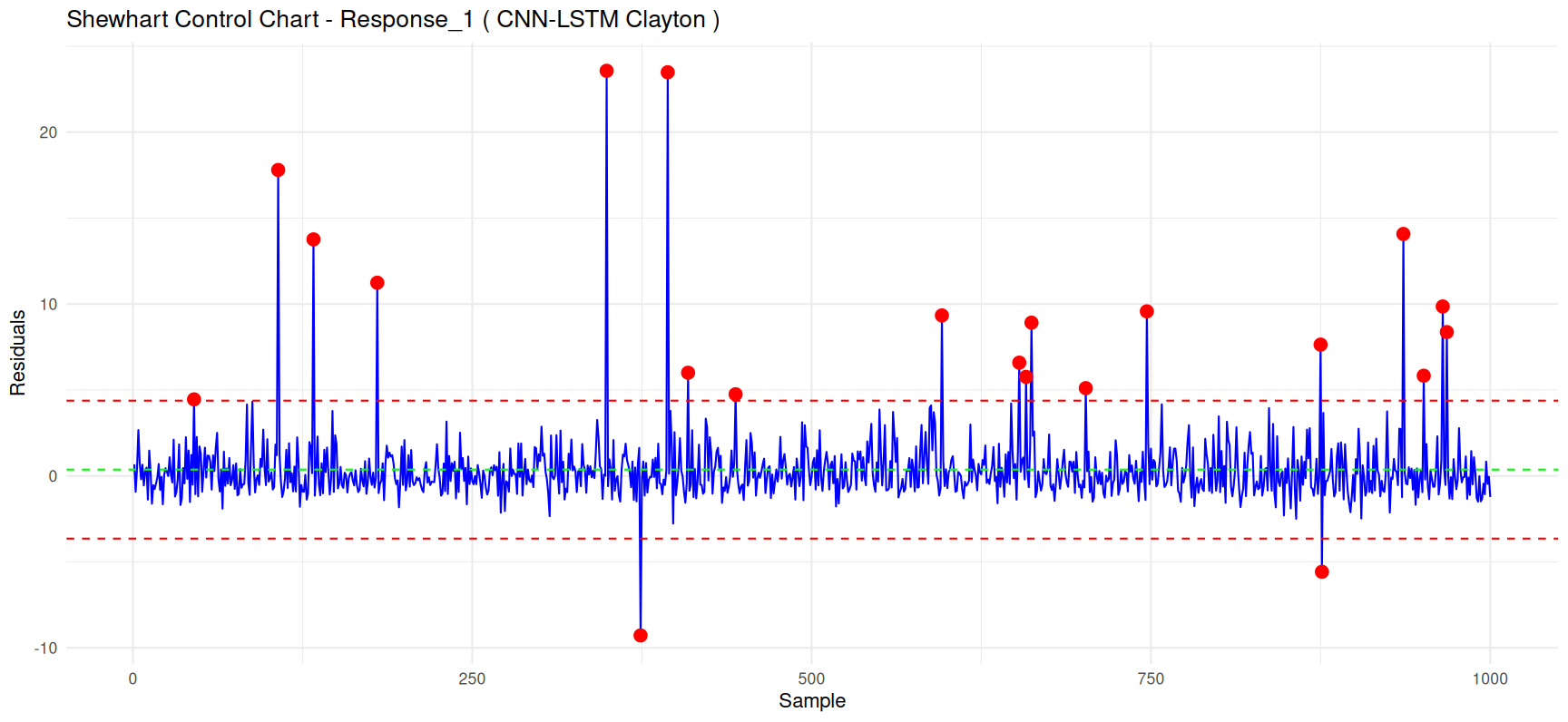}
        \subcaption{CNN-LSTM Clayton $Y_1$}\label{fig:fig7}
    \end{minipage}%
    \begin{minipage}{0.33\textwidth}
        \centering
        \includegraphics[width=\linewidth]{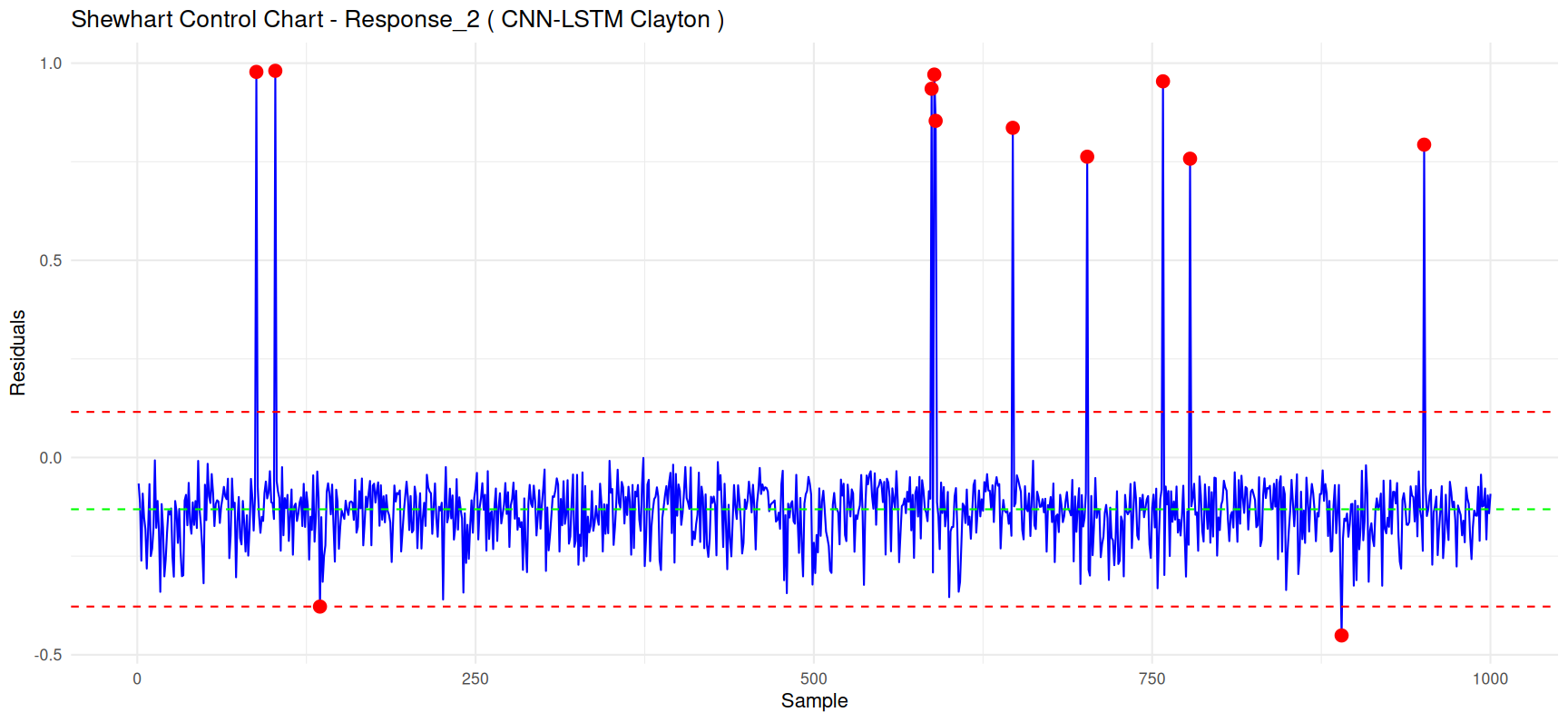}
        \subcaption{CNN-LSTM Clayton $Y_2$}\label{fig:fig8}
    \end{minipage}%
    \begin{minipage}{0.33\textwidth}
        \centering
        \includegraphics[width=\linewidth]{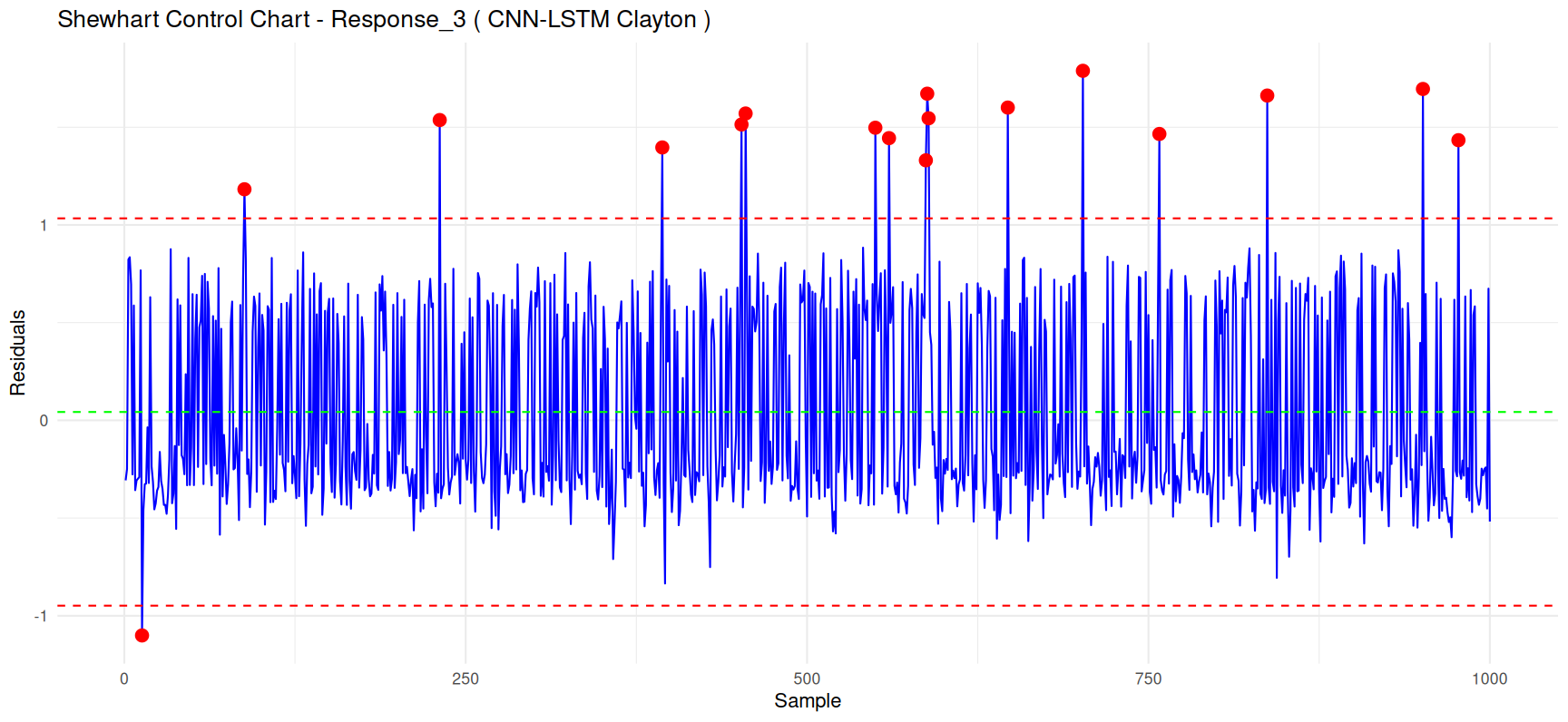}
        \subcaption{CNN-LSTM Clayton $Y_3$}\label{fig:fig9}
    \end{minipage} \\[1ex]
    \begin{minipage}{0.33\textwidth}
        \centering
        \includegraphics[width=\linewidth]{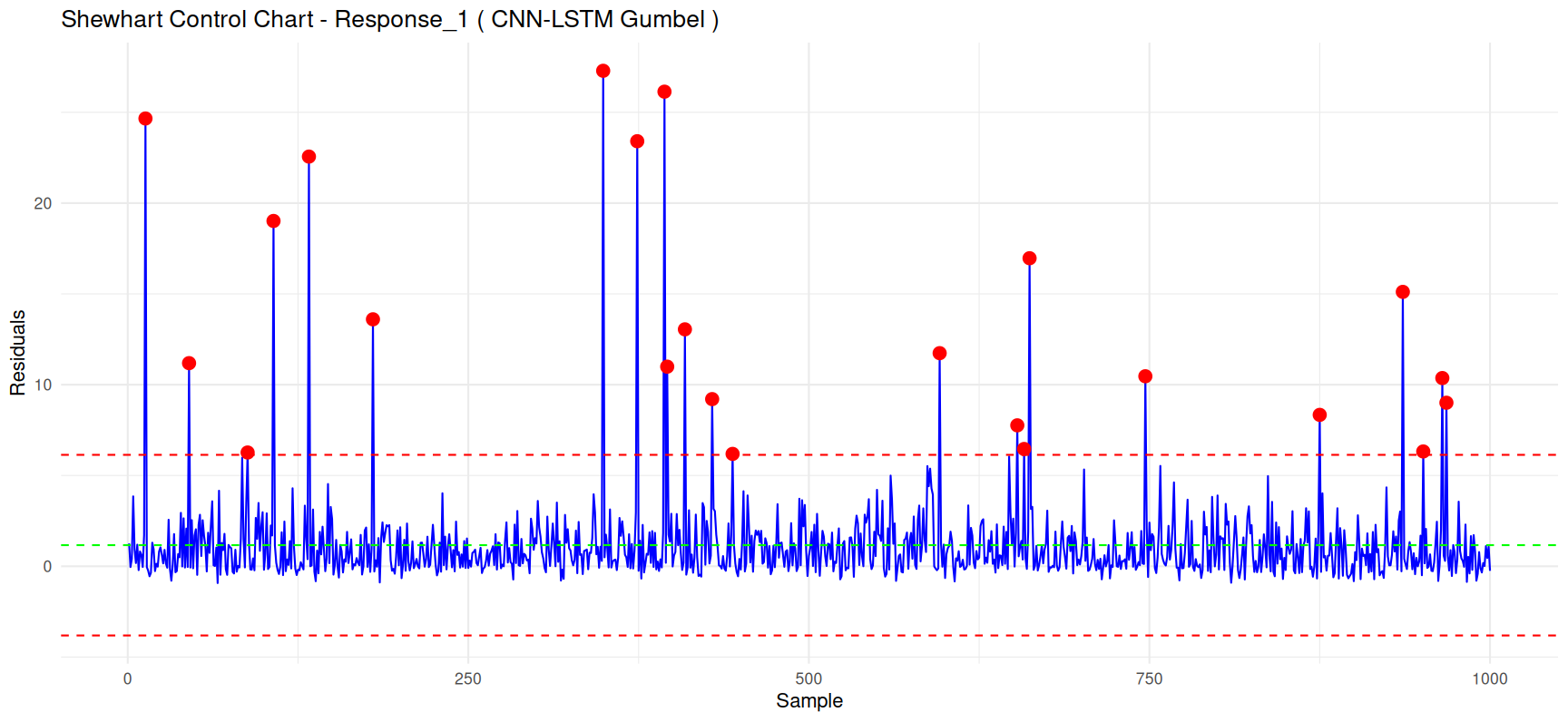}
        \subcaption{CNN-LSTM Gumbel $Y_1$}\label{fig:fig10}
    \end{minipage}%
    \begin{minipage}{0.33\textwidth}
        \centering
        \includegraphics[width=\linewidth]{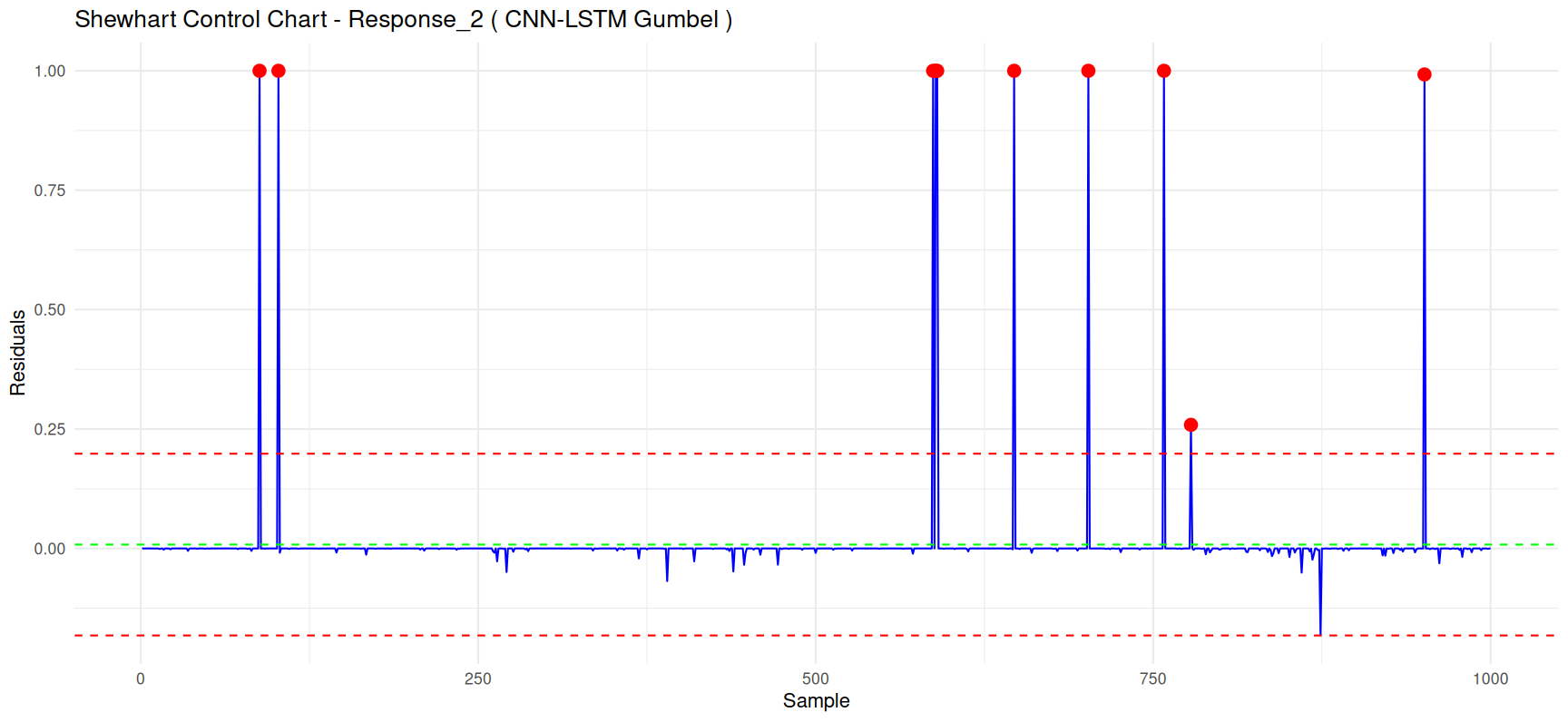}
        \subcaption{CNN-LSTM Gumbel $Y_2$}\label{fig:fig11}
    \end{minipage}%
    \begin{minipage}{0.33\textwidth}
        \centering
        \includegraphics[width=\linewidth]{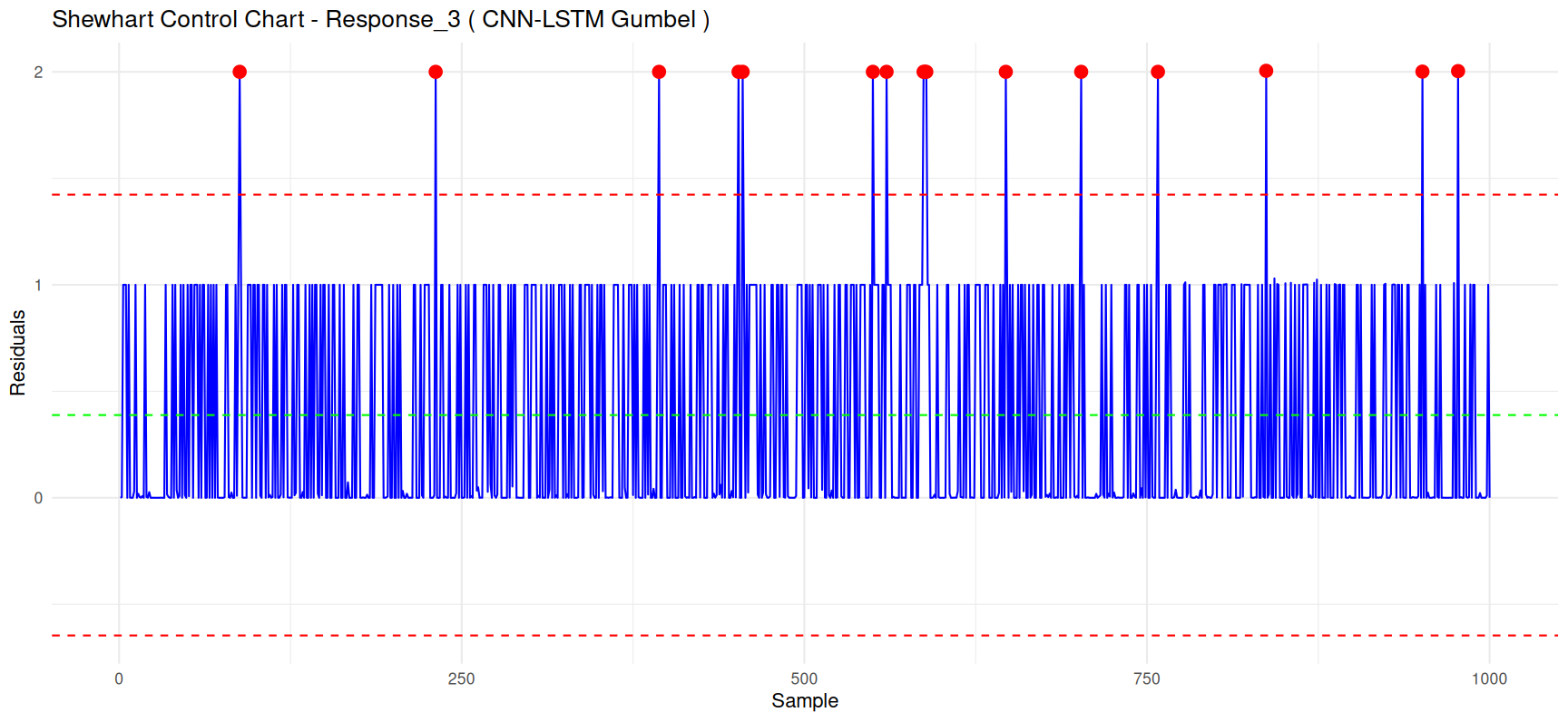}
        \subcaption{CNN-LSTM Gumbel $Y_3$}\label{fig:fig12}
    \end{minipage}
    \begin{minipage}{0.33\textwidth}
        \centering
        \includegraphics[width=\linewidth]{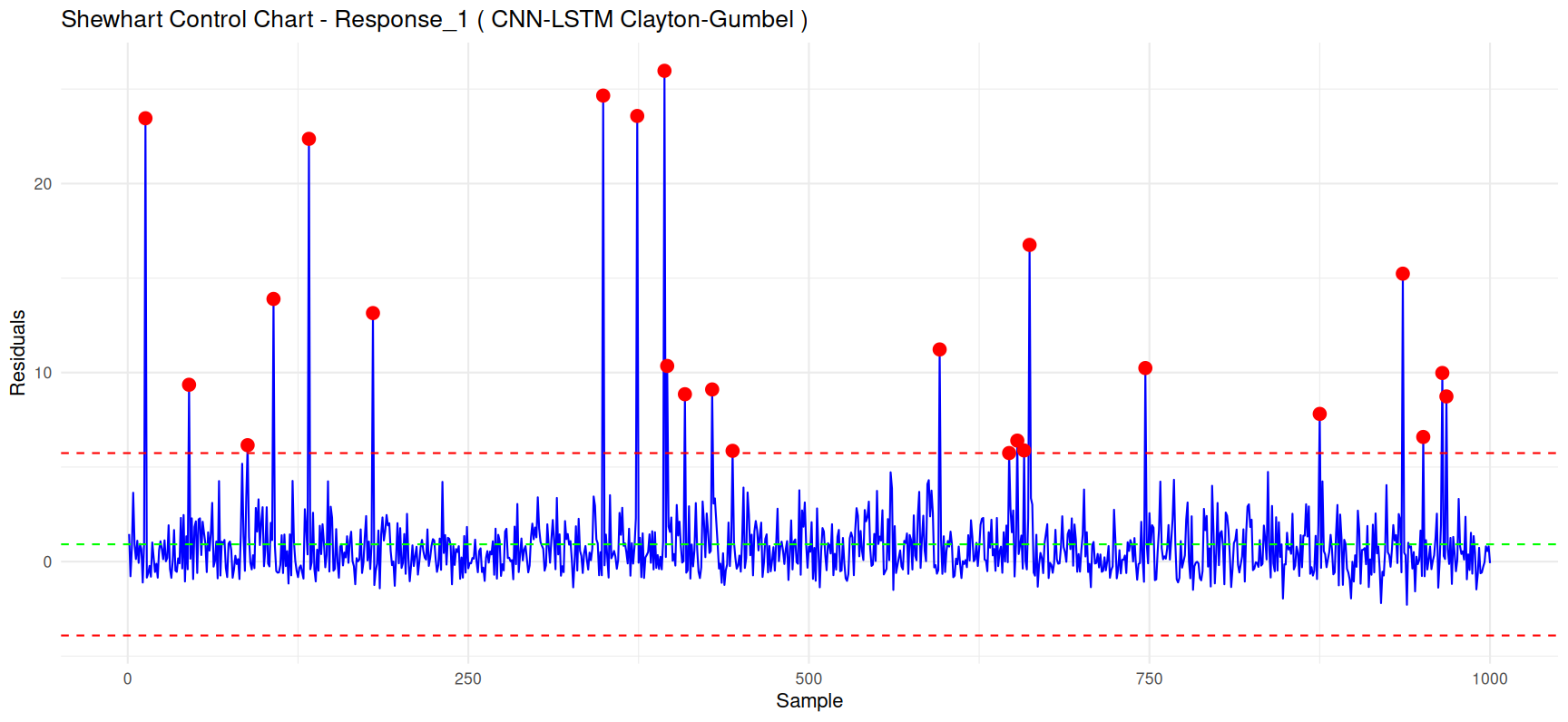}
        \subcaption{CNN-LSTM Clayton-Gumbel $Y_1$}\label{fig:fig4}
    \end{minipage}%
    \begin{minipage}{0.33\textwidth}
        \centering
        \includegraphics[width=\linewidth]{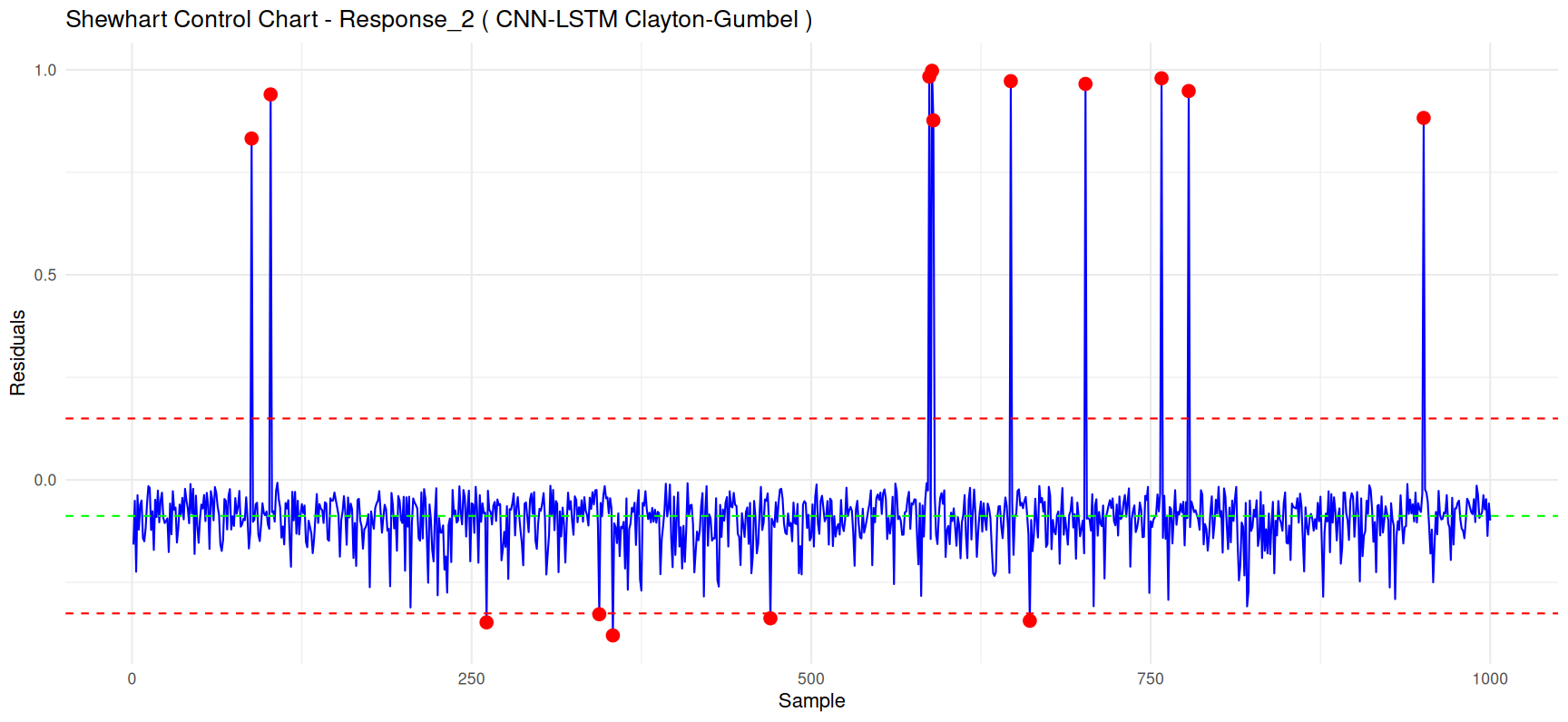}
        \subcaption{CNN-LSTM Clayton-Gumbel $Y_2$}\label{fig:fig5}
    \end{minipage}%
    \begin{minipage}{0.33\textwidth}
        \centering
        \includegraphics[width=\linewidth]{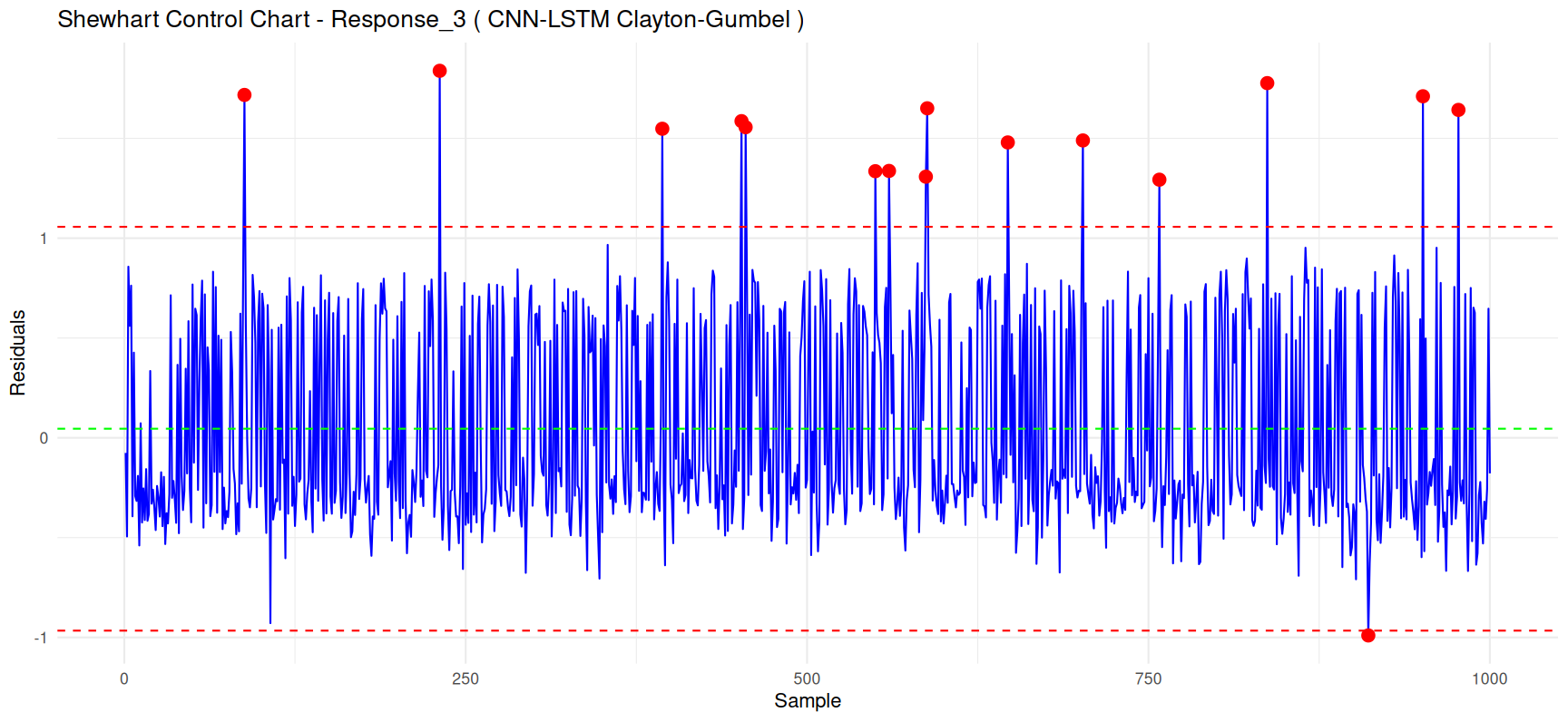}
        \subcaption{CNN-LSTM Clayton-Gumbel $Y_3$}\label{fig:fig6}
    \end{minipage} \\[1ex]
    \begin{minipage}{0.33\textwidth}
        \centering
        \includegraphics[width=\linewidth]{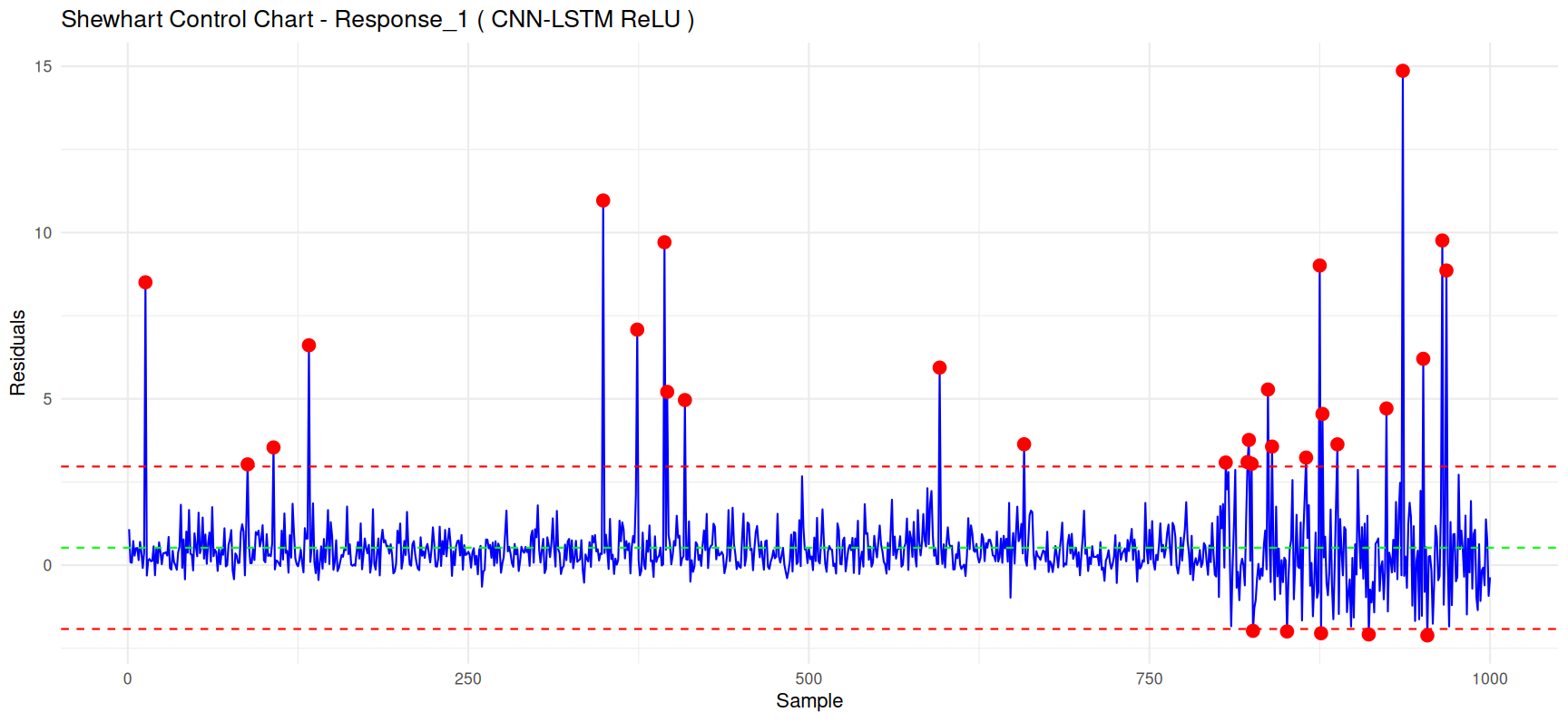}
        \subcaption{CNN-LSTM ReLU $Y_1$}\label{fig:fig4}
    \end{minipage}%
    \begin{minipage}{0.33\textwidth}
        \centering
        \includegraphics[width=\linewidth]{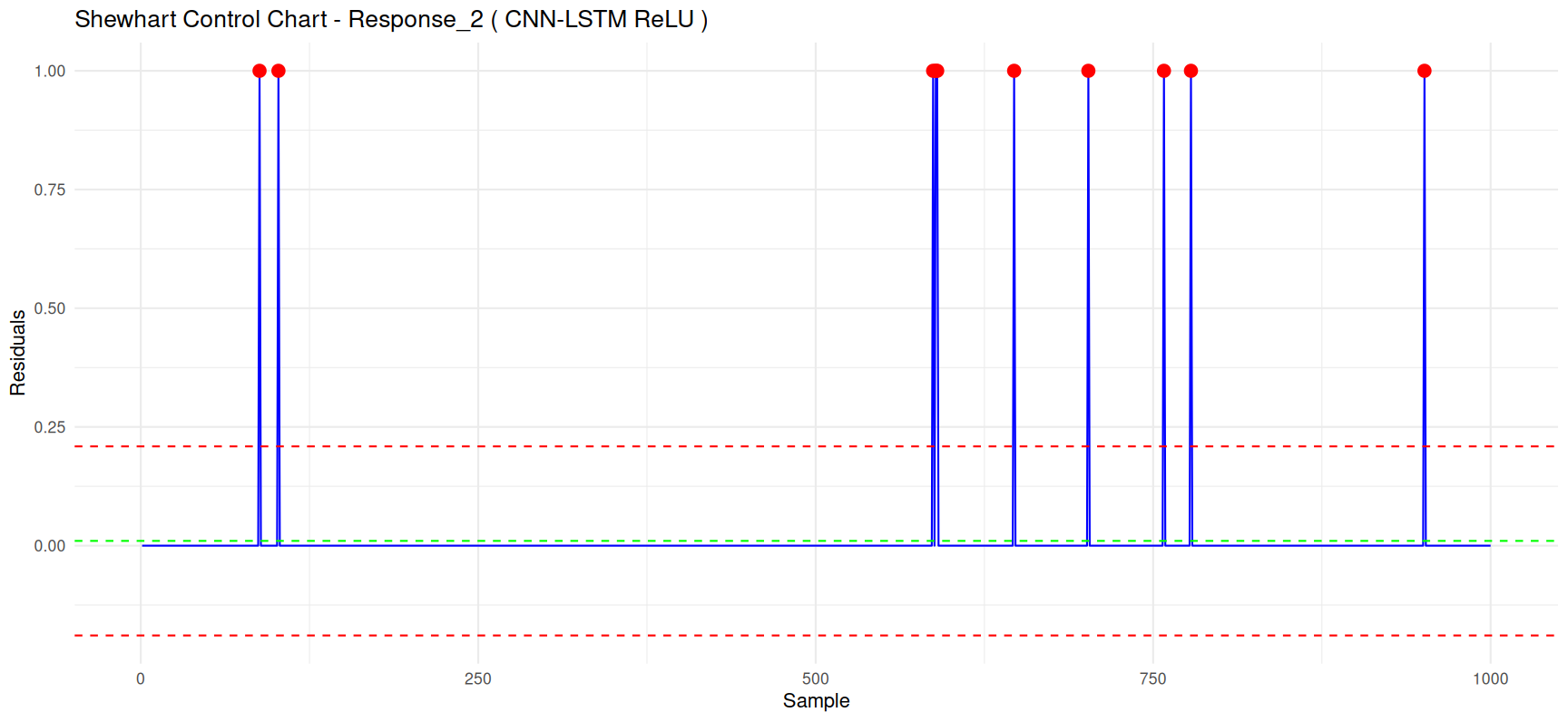}
        \subcaption{CNN-LSTM ReLU $Y_2$}\label{fig:fig5}
    \end{minipage}%
    \begin{minipage}{0.33\textwidth}
        \centering
        \includegraphics[width=\linewidth]{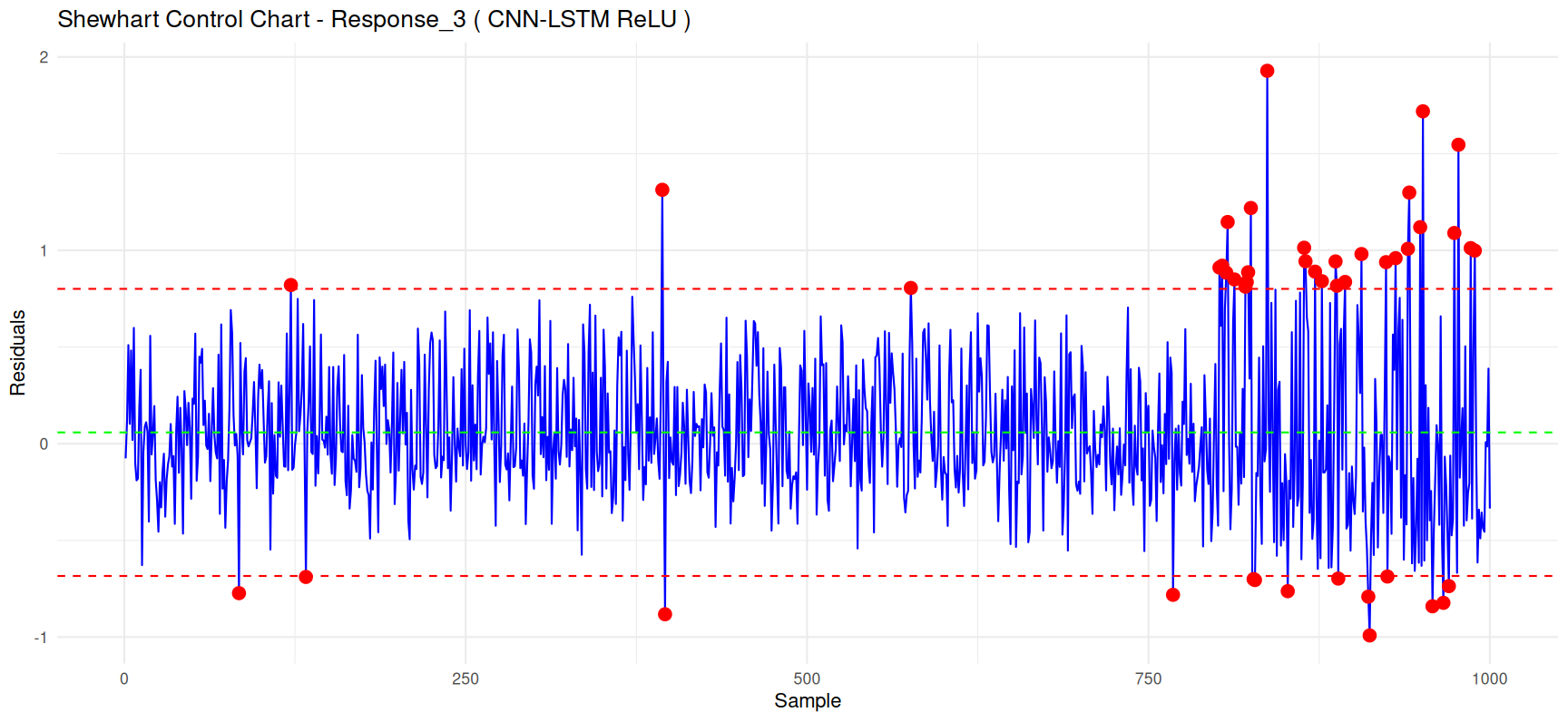}
        \subcaption{CNN-LSTM ReLU $Y_3$}\label{fig:fig6}
    \end{minipage} \\[1ex]
    \begin{minipage}{0.33\textwidth}
        \centering
        \includegraphics[width=\linewidth]{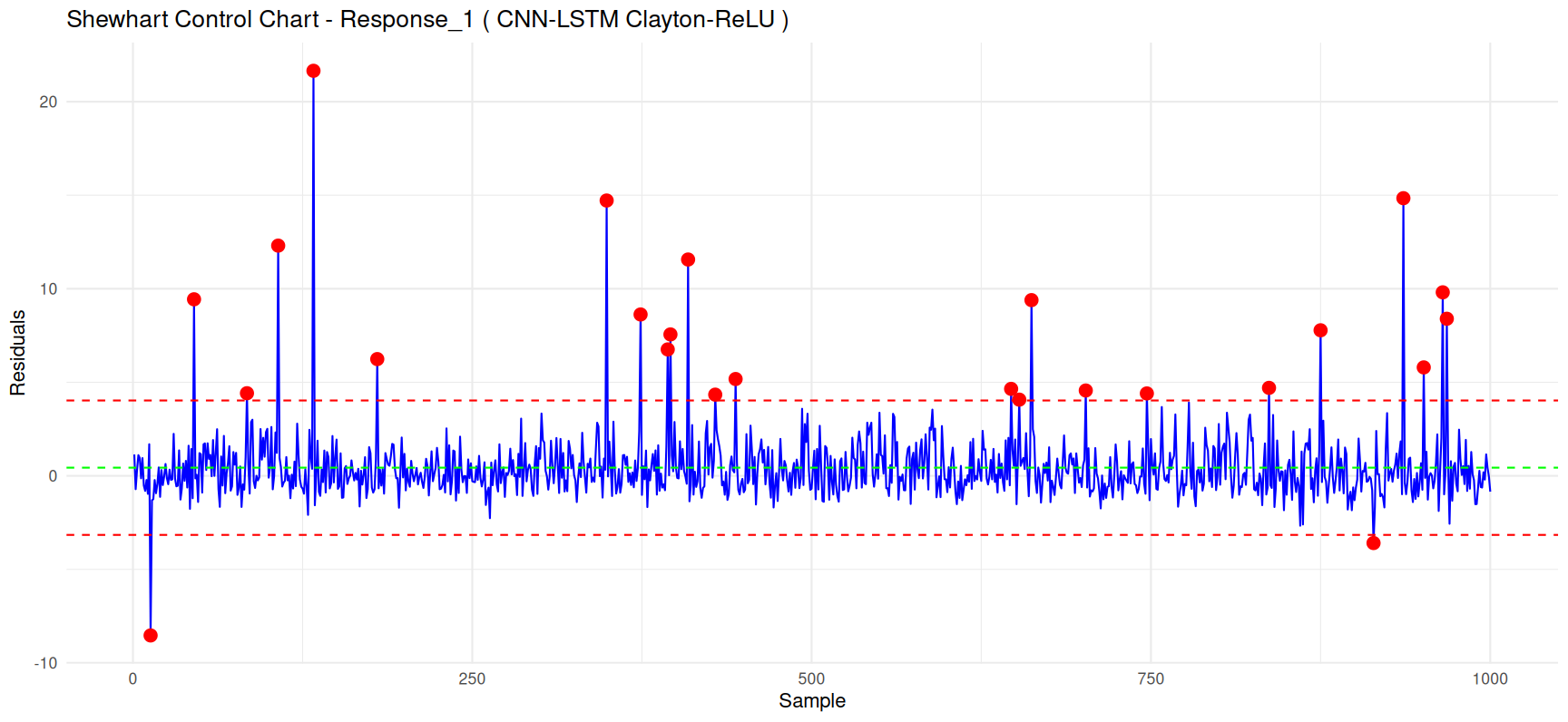}
        \subcaption{CNN-LSTM Clayton-ReLU $Y_1$}\label{fig:fig4}
    \end{minipage}%
    \begin{minipage}{0.33\textwidth}
        \centering
        \includegraphics[width=\linewidth]{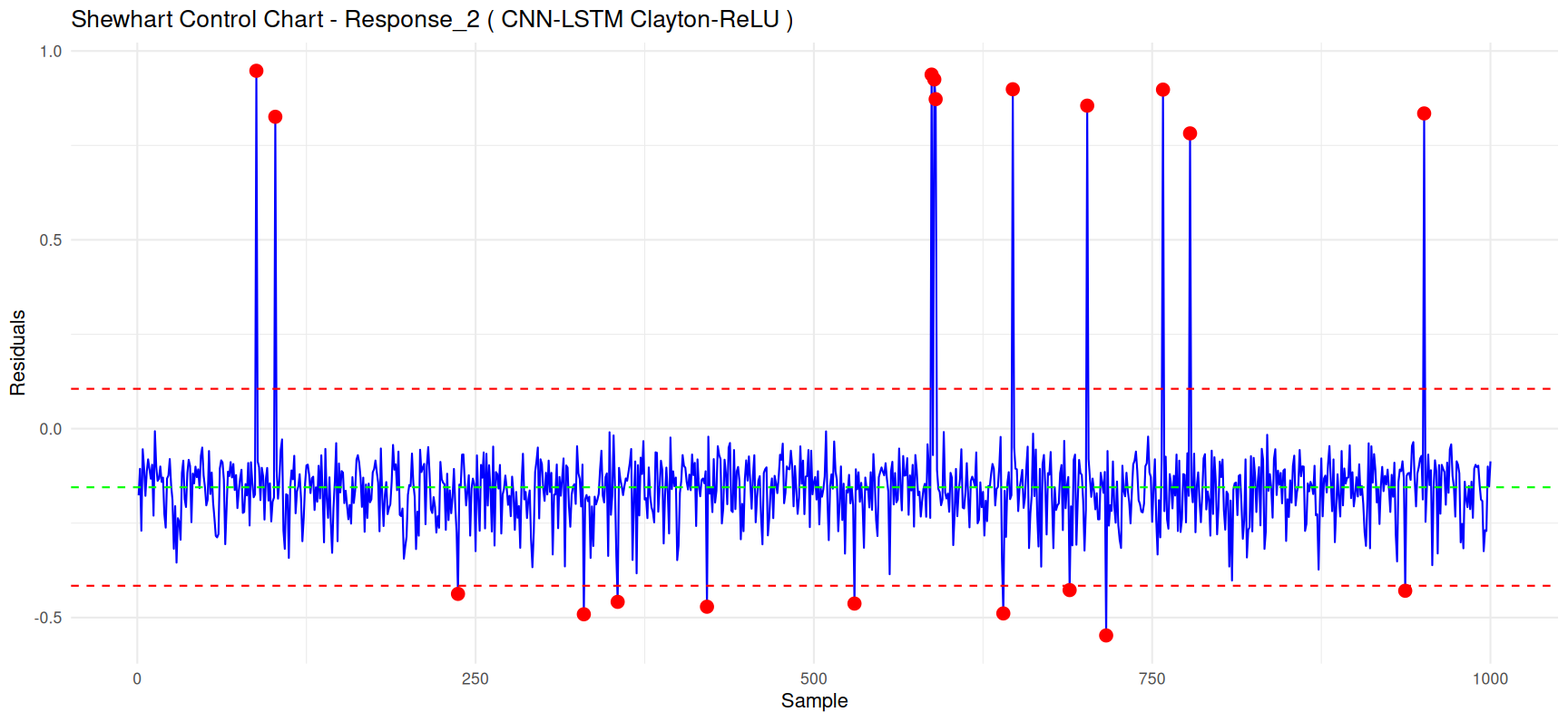}
        \subcaption{CNN-LSTM Clayton-ReLU $Y_2$}\label{fig:fig5}
    \end{minipage}%
    \begin{minipage}{0.33\textwidth}
        \centering
        \includegraphics[width=\linewidth]{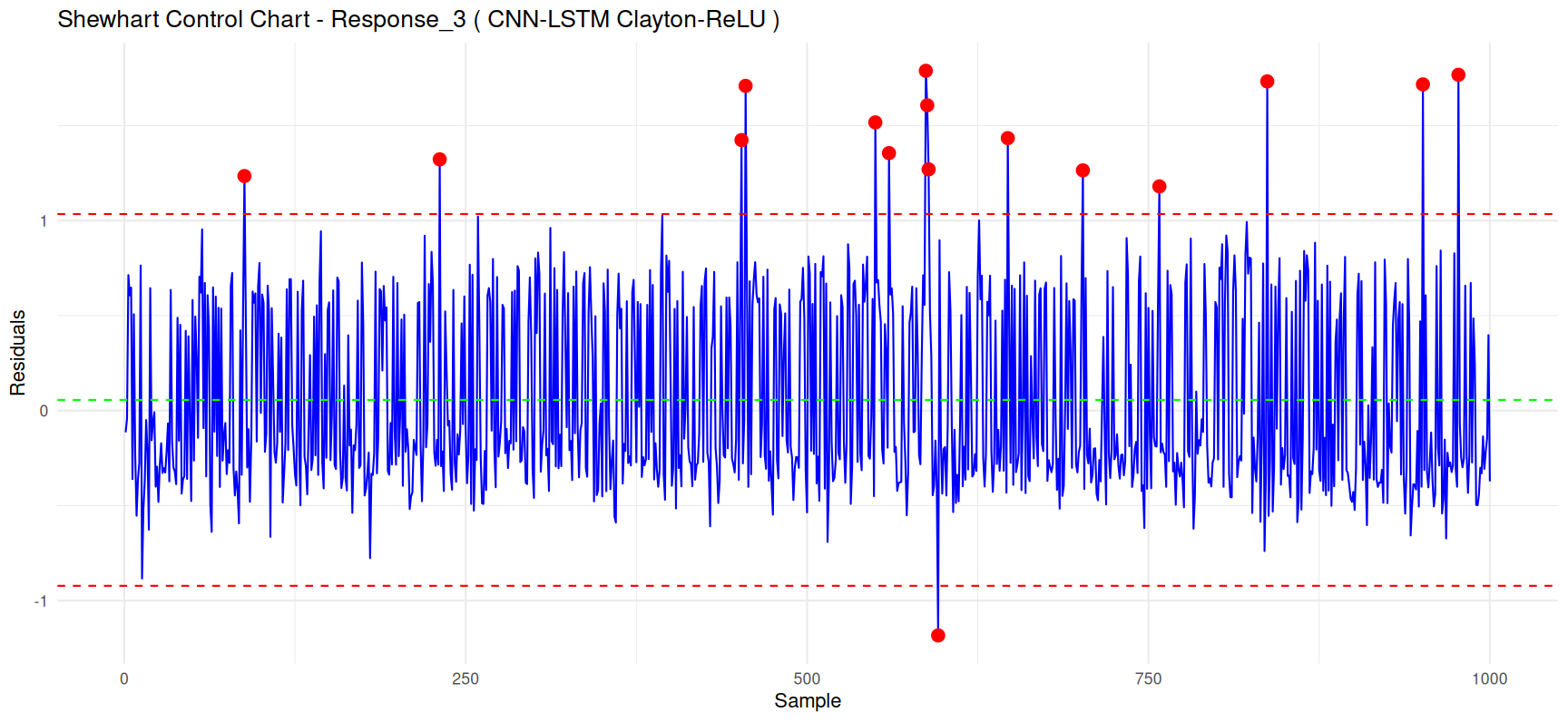}
        \subcaption{CNN-LSTM Clayton-ReLU $Y_3$}\label{fig:fig6}
    \end{minipage} \\[1ex]
    \begin{minipage}{0.33\textwidth}
        \centering
        \includegraphics[width=\linewidth]{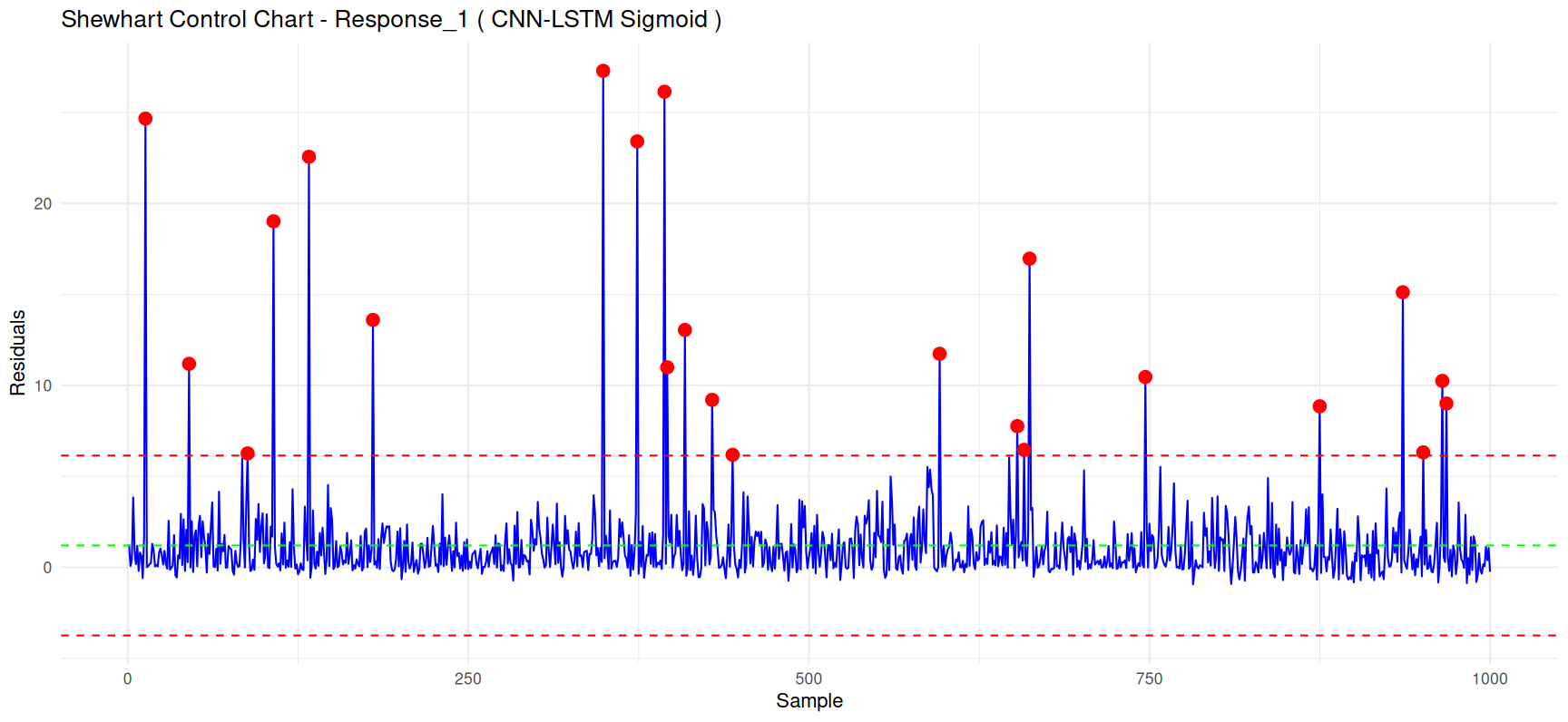}
        \subcaption{CNN-LSTM Sigmoid $Y_1$}\label{fig:fig4}
    \end{minipage}%
    \begin{minipage}{0.33\textwidth}
        \centering
        \includegraphics[width=\linewidth]{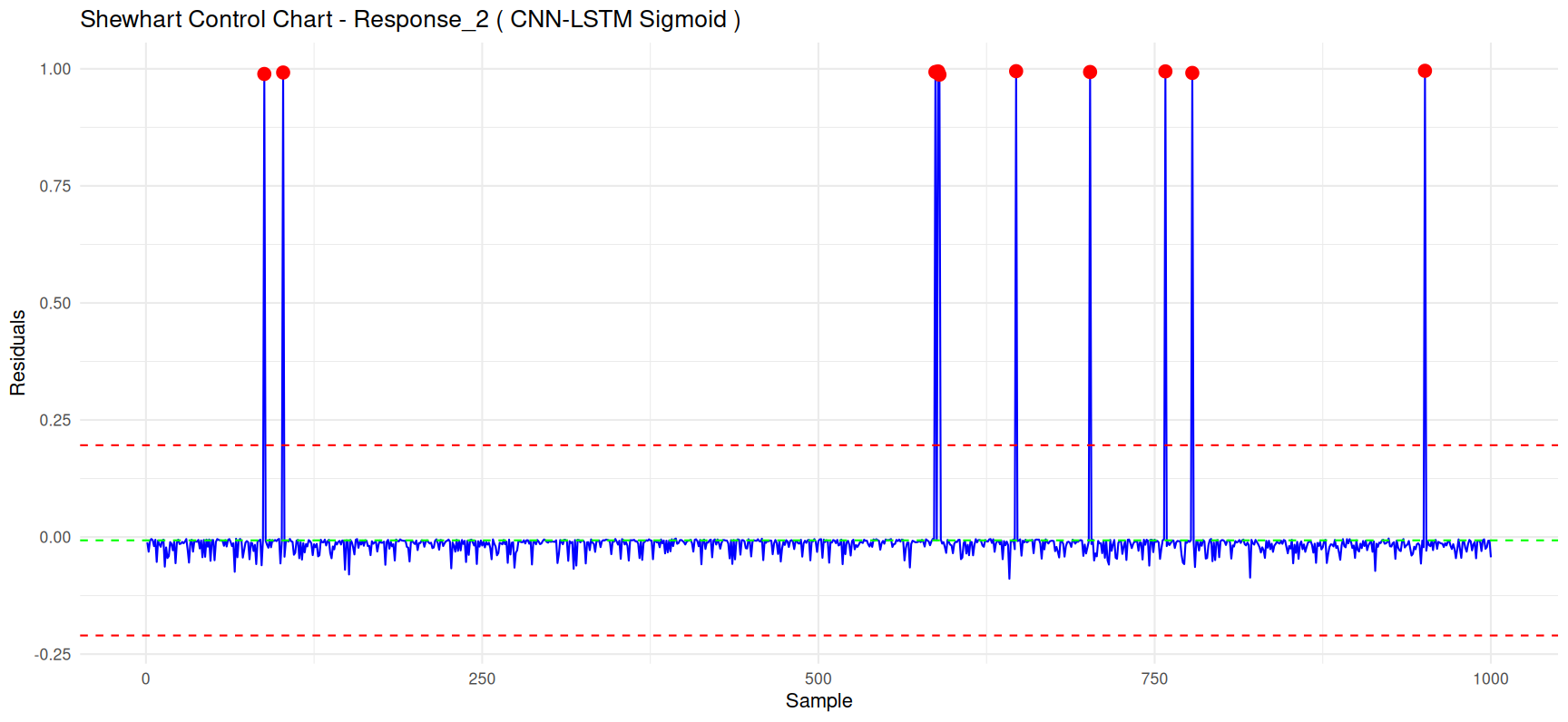}
        \subcaption{CNN-LSTM Sigmoid $Y_2$}\label{fig:fig5}
    \end{minipage}%
    \begin{minipage}{0.33\textwidth}
        \centering
        \includegraphics[width=\linewidth]{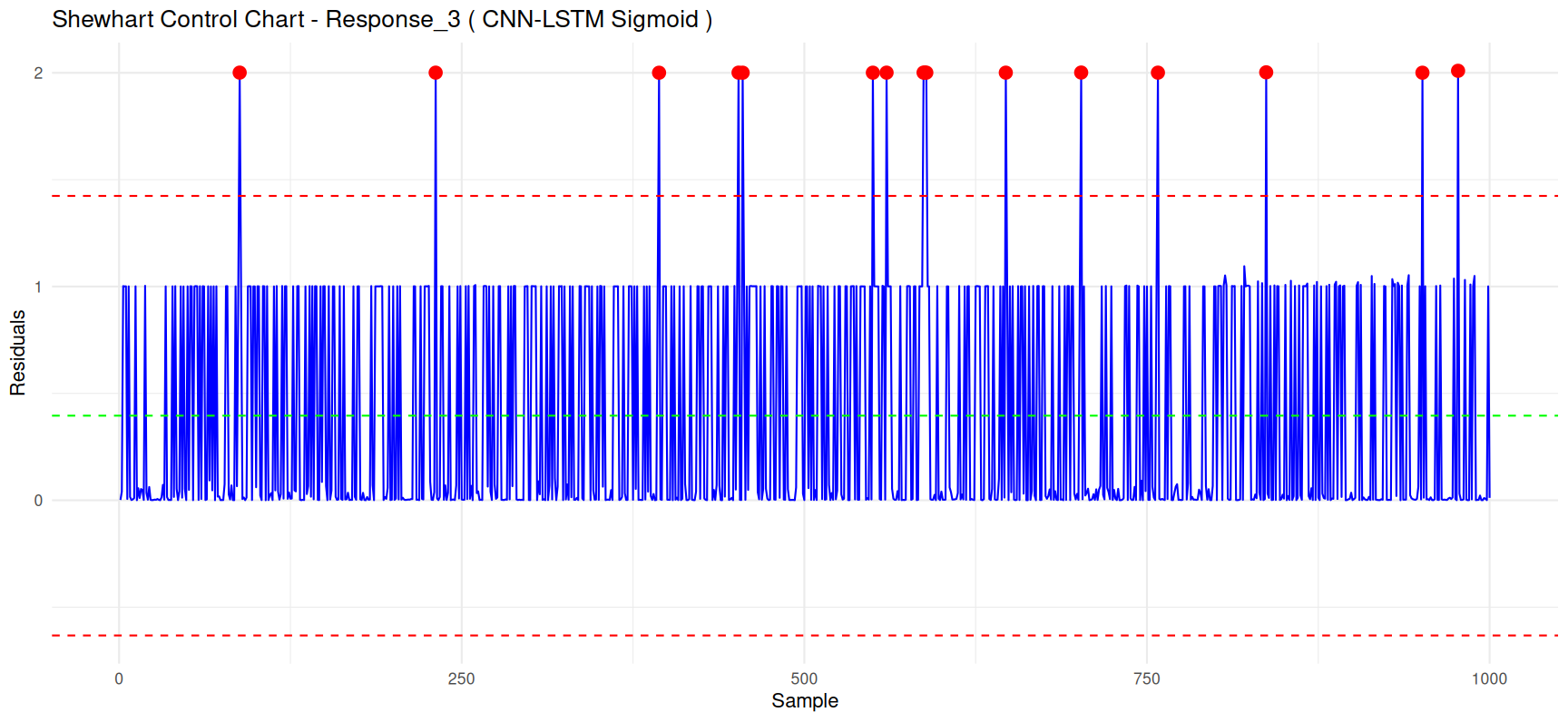}
        \subcaption{CNN-LSTM Sigmoid $Y_3$}\label{fig:fig6}
    \end{minipage} \\[1ex]
    \caption{Residual Shewhart Control Charts of CNN-LSTM Models with Simulated Data.}
    \label{fig:allfigures1}
\end{figure}

\begin{figure}[htp]
    \centering
    \begin{minipage}{0.33\textwidth}
        \centering
        \includegraphics[width=\linewidth]{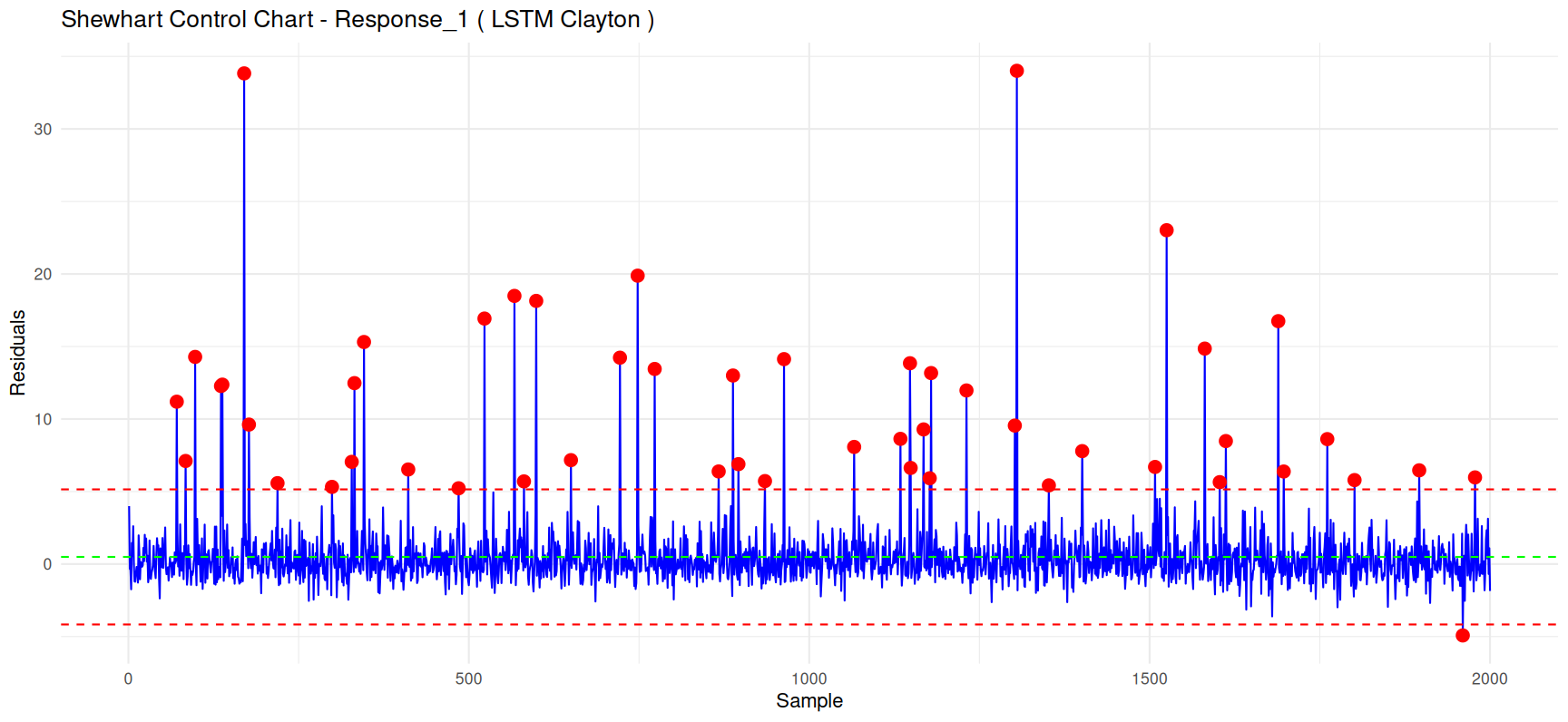}
        \subcaption{LSTM Clayton $Y_1$}\label{fig:fig7}
    \end{minipage}%
    \begin{minipage}{0.33\textwidth}
        \centering
        \includegraphics[width=\linewidth]{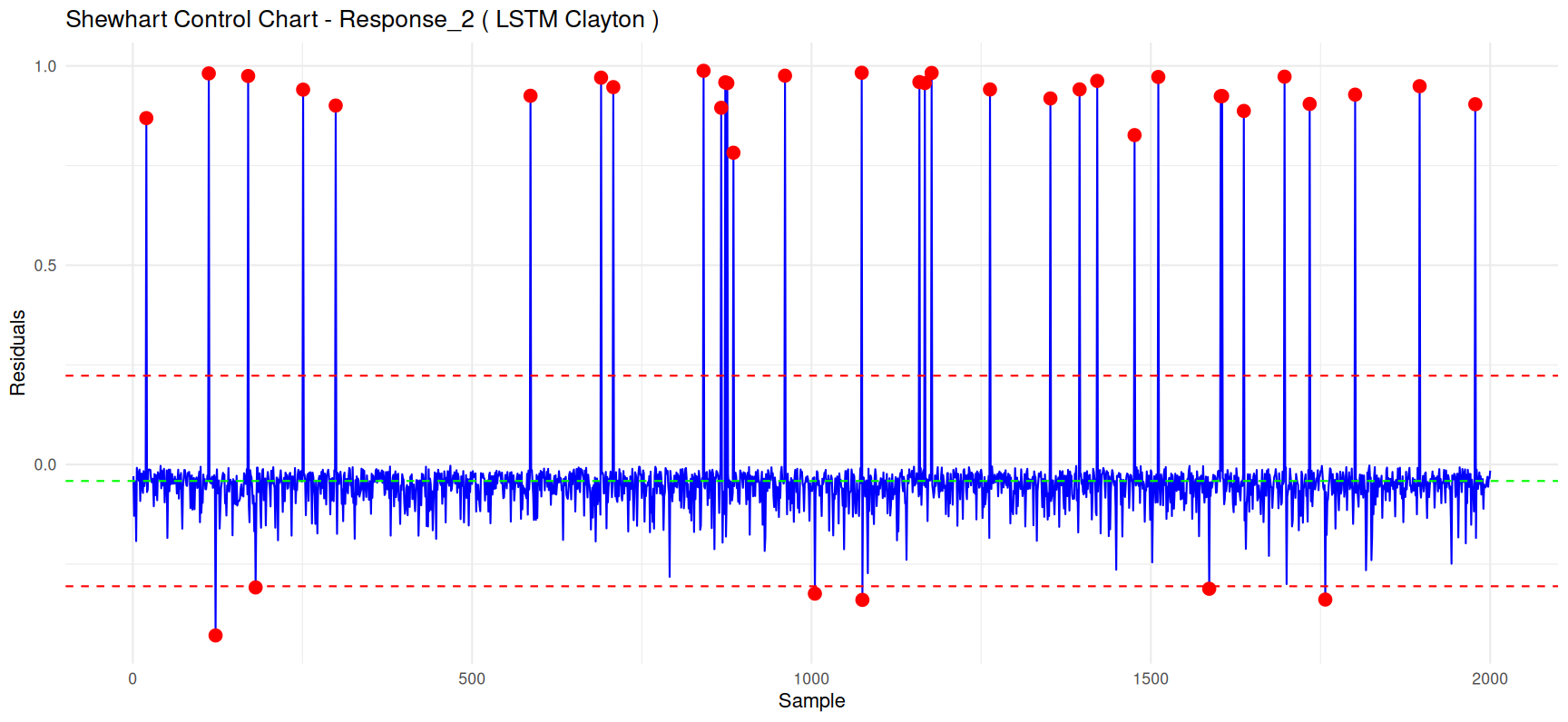}
        \subcaption{LSTM Clayton $Y_2$}\label{fig:fig8}
    \end{minipage}%
    \begin{minipage}{0.33\textwidth}
        \centering
        \includegraphics[width=\linewidth]{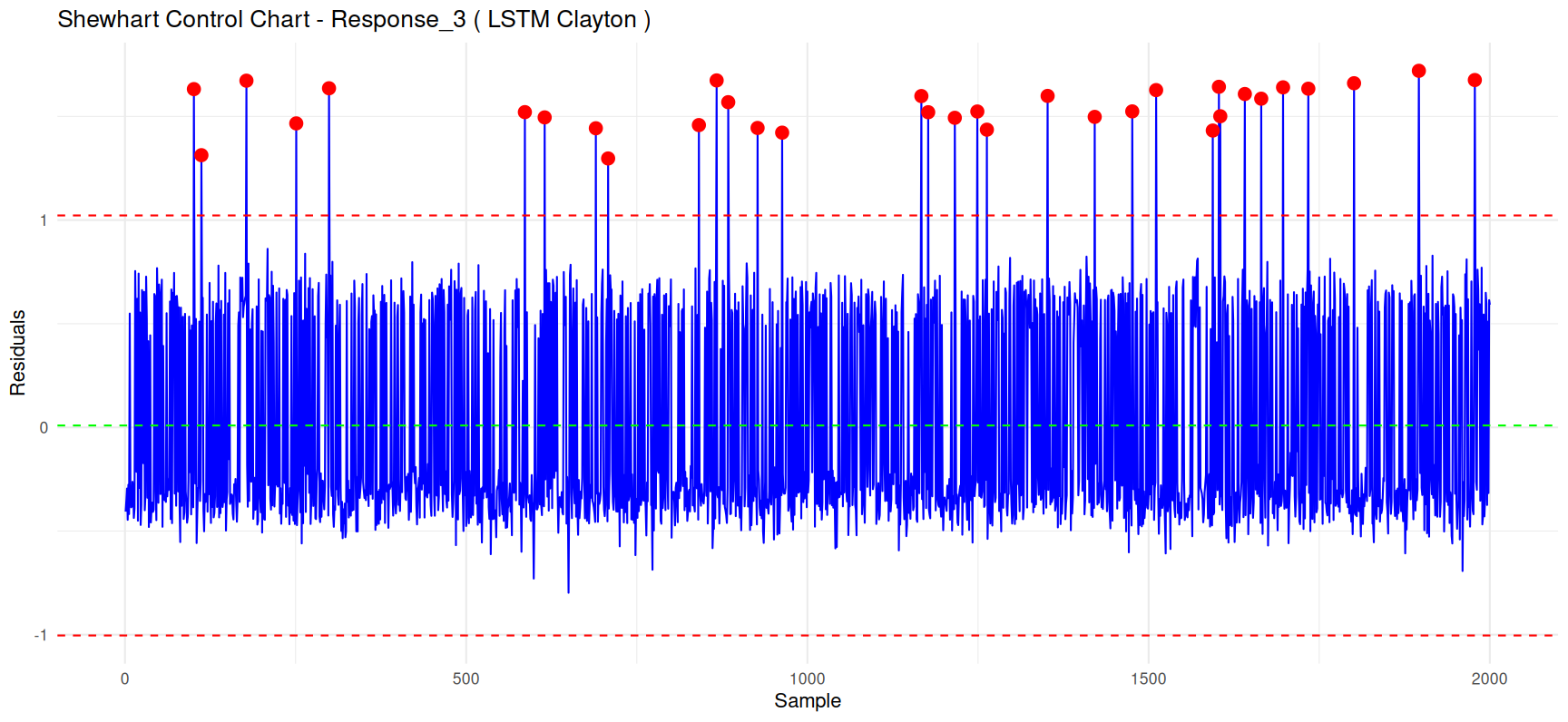}
        \subcaption{LSTM Clayton $Y_3$}\label{fig:fig9}
    \end{minipage} \\[1ex]
    \begin{minipage}{0.33\textwidth}
        \centering
        \includegraphics[width=\linewidth]{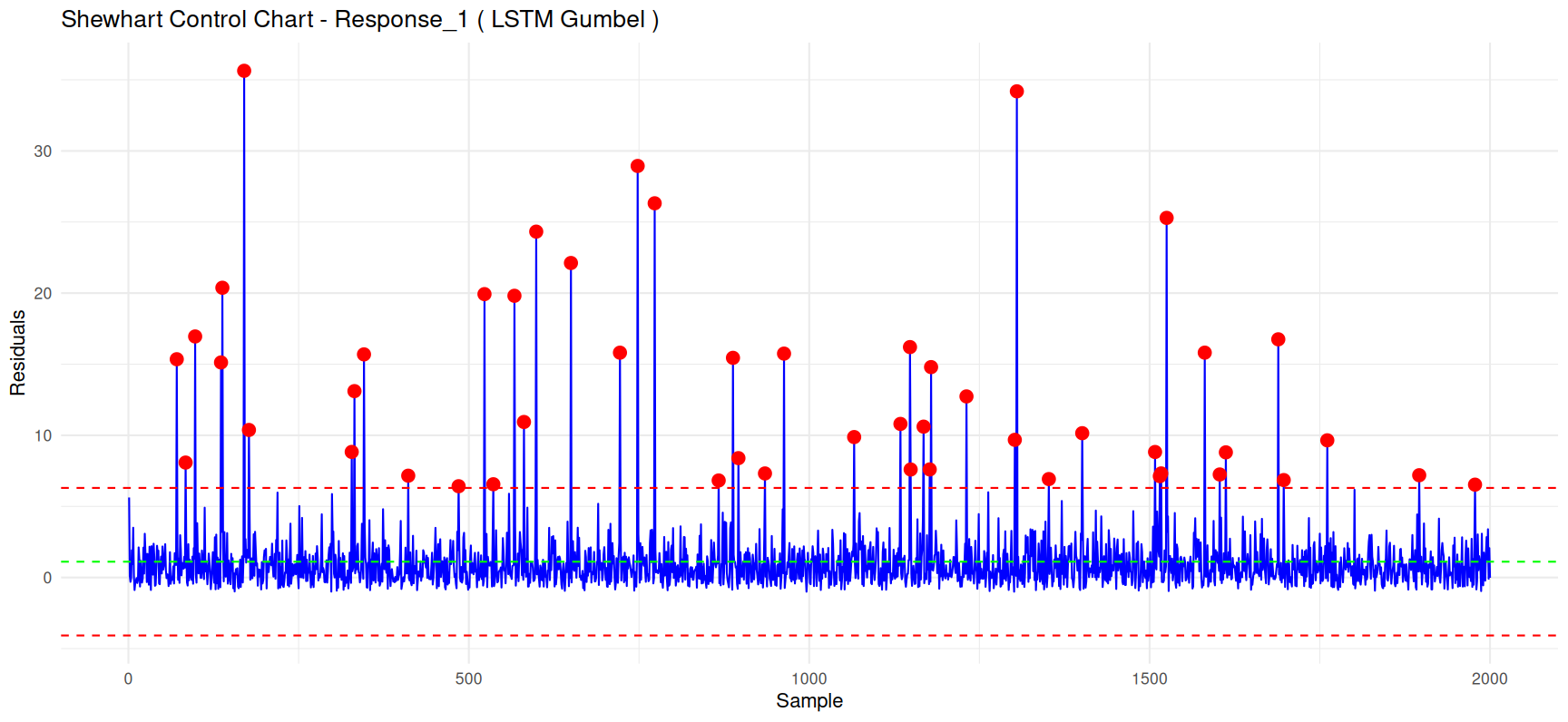}
        \subcaption{LSTM Gumbel $Y_1$}\label{fig:fig10}
    \end{minipage}%
    \begin{minipage}{0.33\textwidth}
        \centering
        \includegraphics[width=\linewidth]{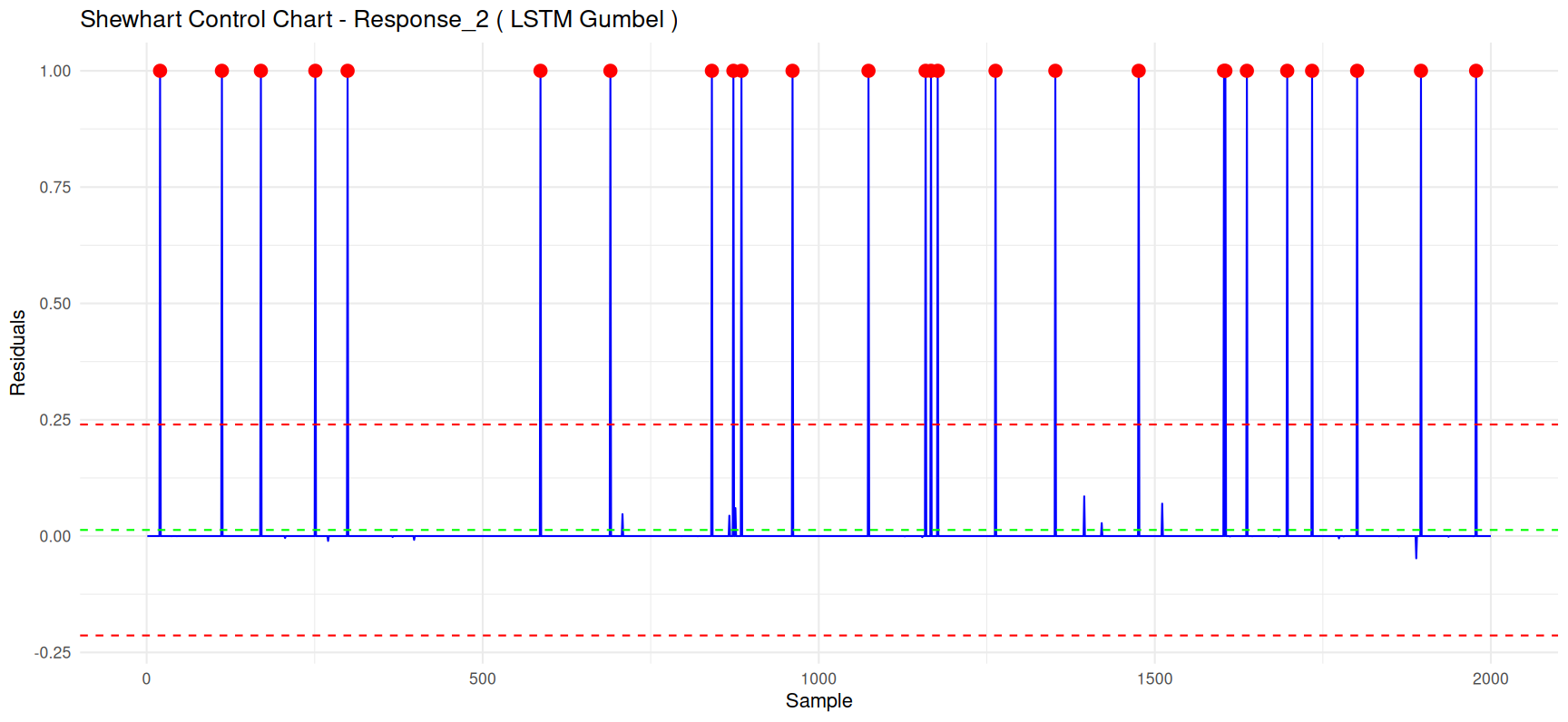}
        \subcaption{LSTM Gumbel $Y_2$}\label{fig:fig11}
    \end{minipage}%
    \begin{minipage}{0.33\textwidth}
        \centering
        \includegraphics[width=\linewidth]{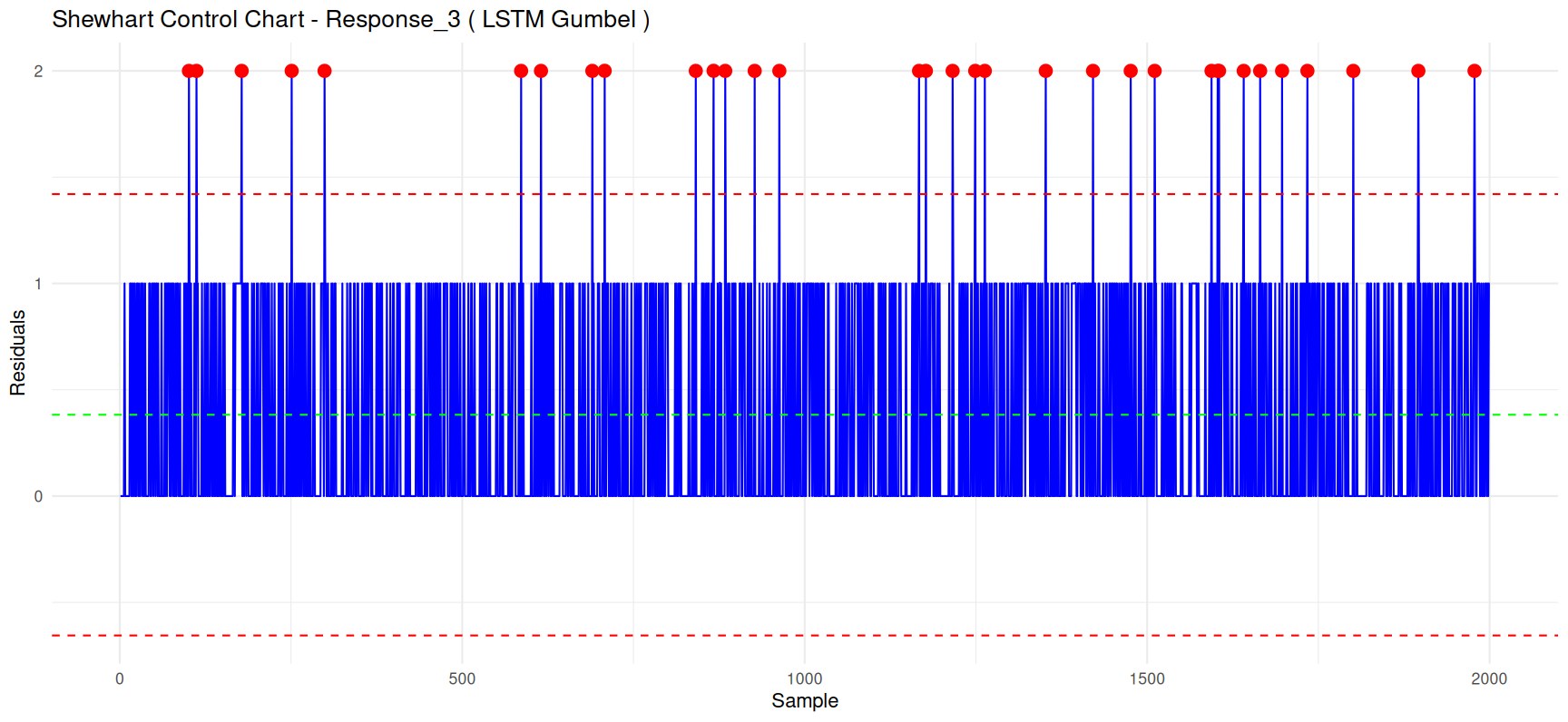}
        \subcaption{LSTM Gumbel $Y_3$}\label{fig:fig12}
    \end{minipage}
    \begin{minipage}{0.33\textwidth}
        \centering
        \includegraphics[width=\linewidth]{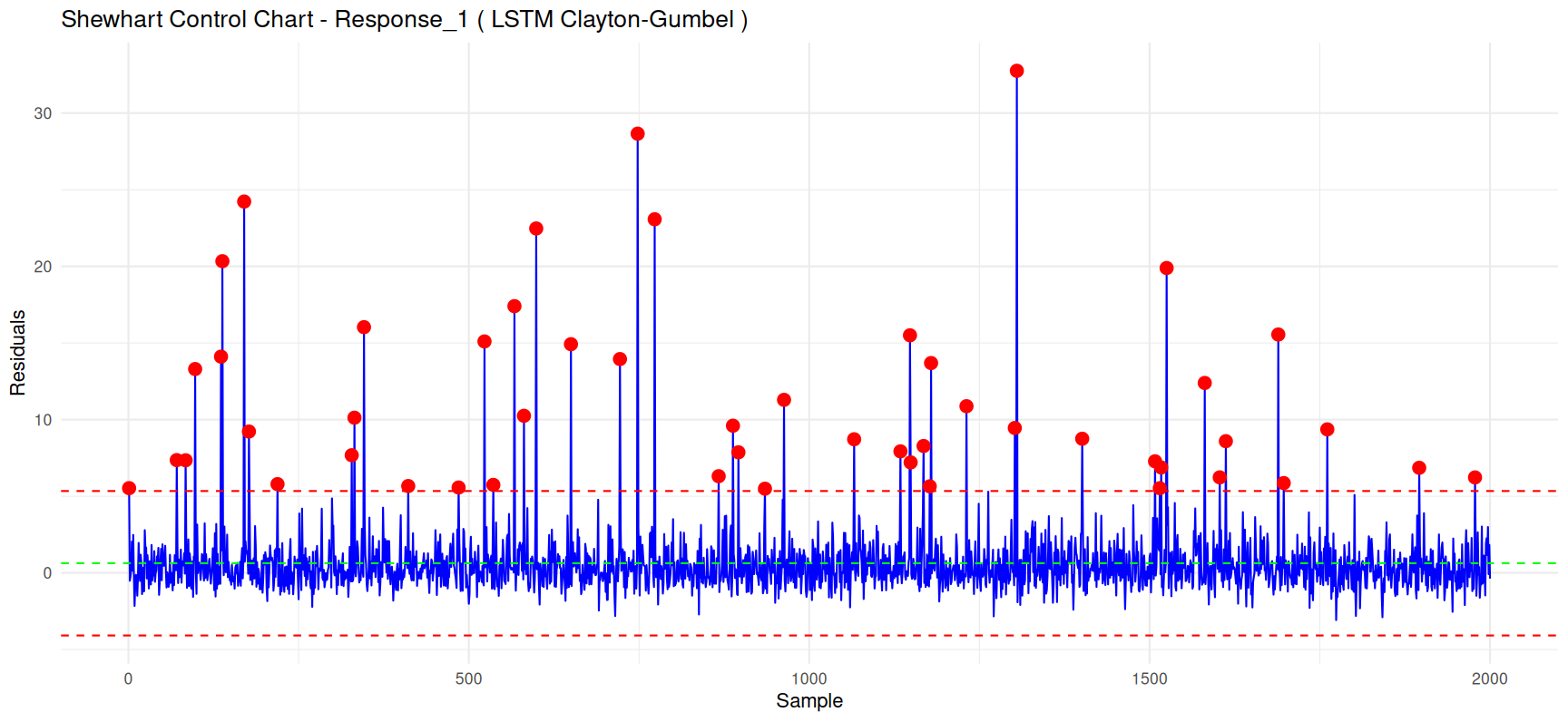}
        \subcaption{LSTM Clayton-Gumbel $Y_1$}\label{fig:fig4}
    \end{minipage}%
    \begin{minipage}{0.33\textwidth}
        \centering
        \includegraphics[width=\linewidth]{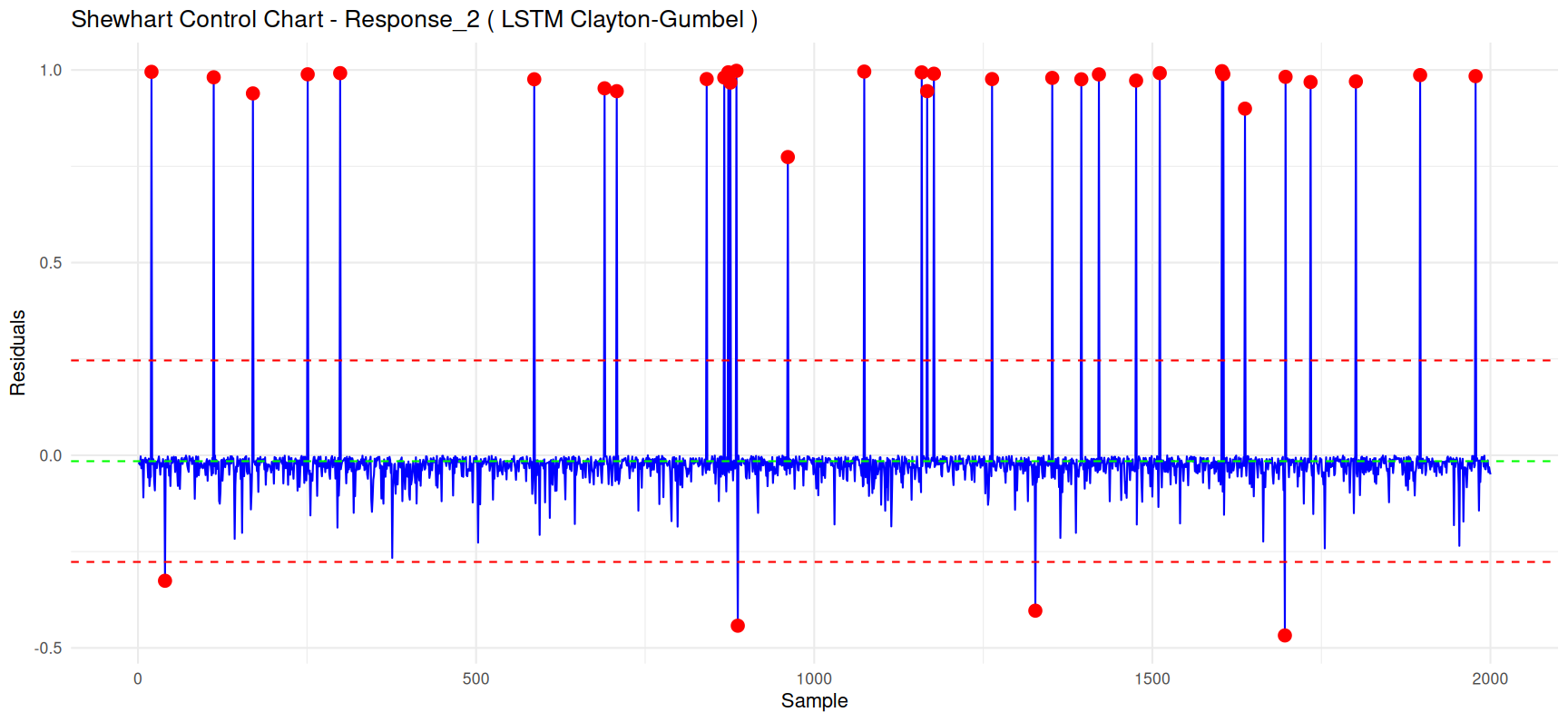}
        \subcaption{LSTM Clayton-Gumbel $Y_2$}\label{fig:fig5}
    \end{minipage}%
    \begin{minipage}{0.33\textwidth}
        \centering
        \includegraphics[width=\linewidth]{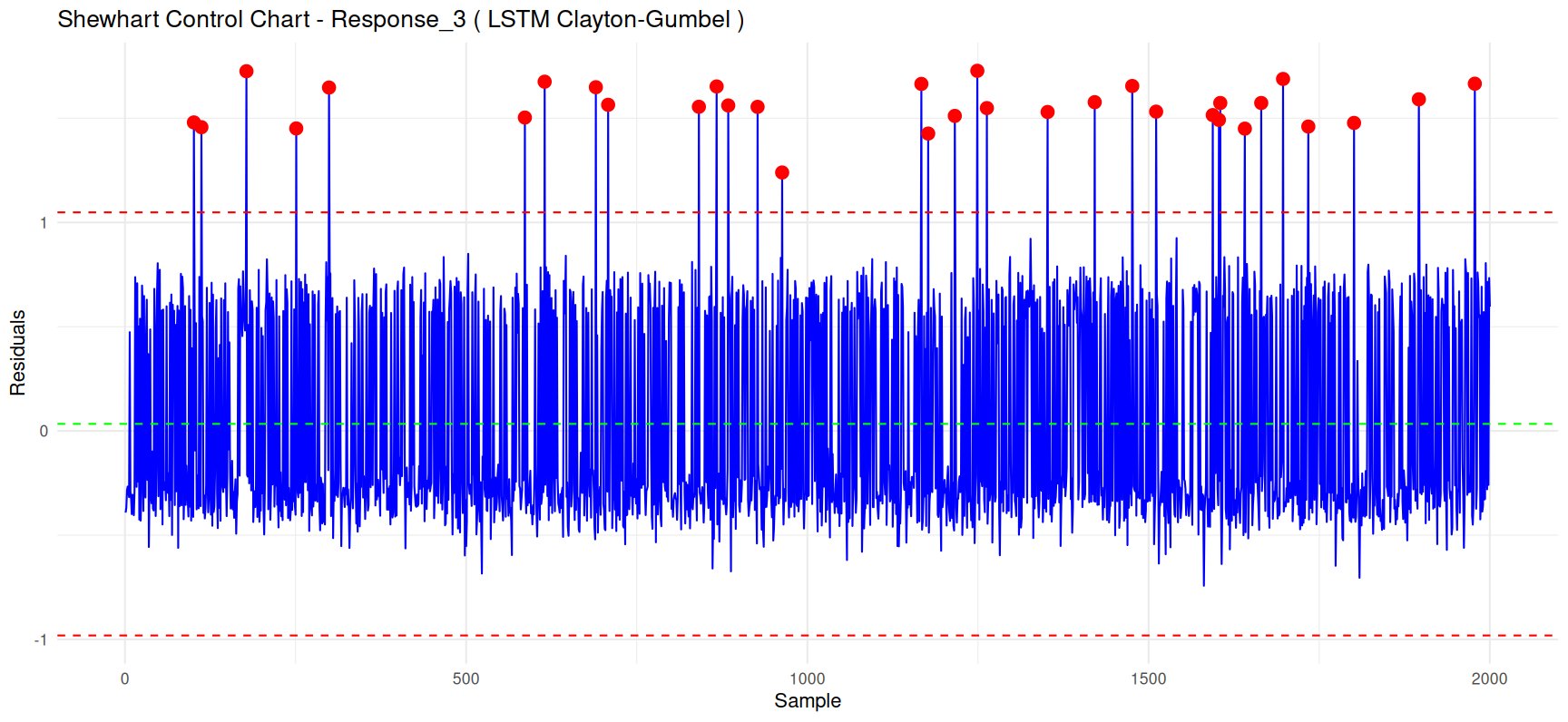}
        \subcaption{LSTM Clayton-Gumbel $Y_3$}\label{fig:fig6}
    \end{minipage} \\[1ex]
    \begin{minipage}{0.33\textwidth}
        \centering
        \includegraphics[width=\linewidth]{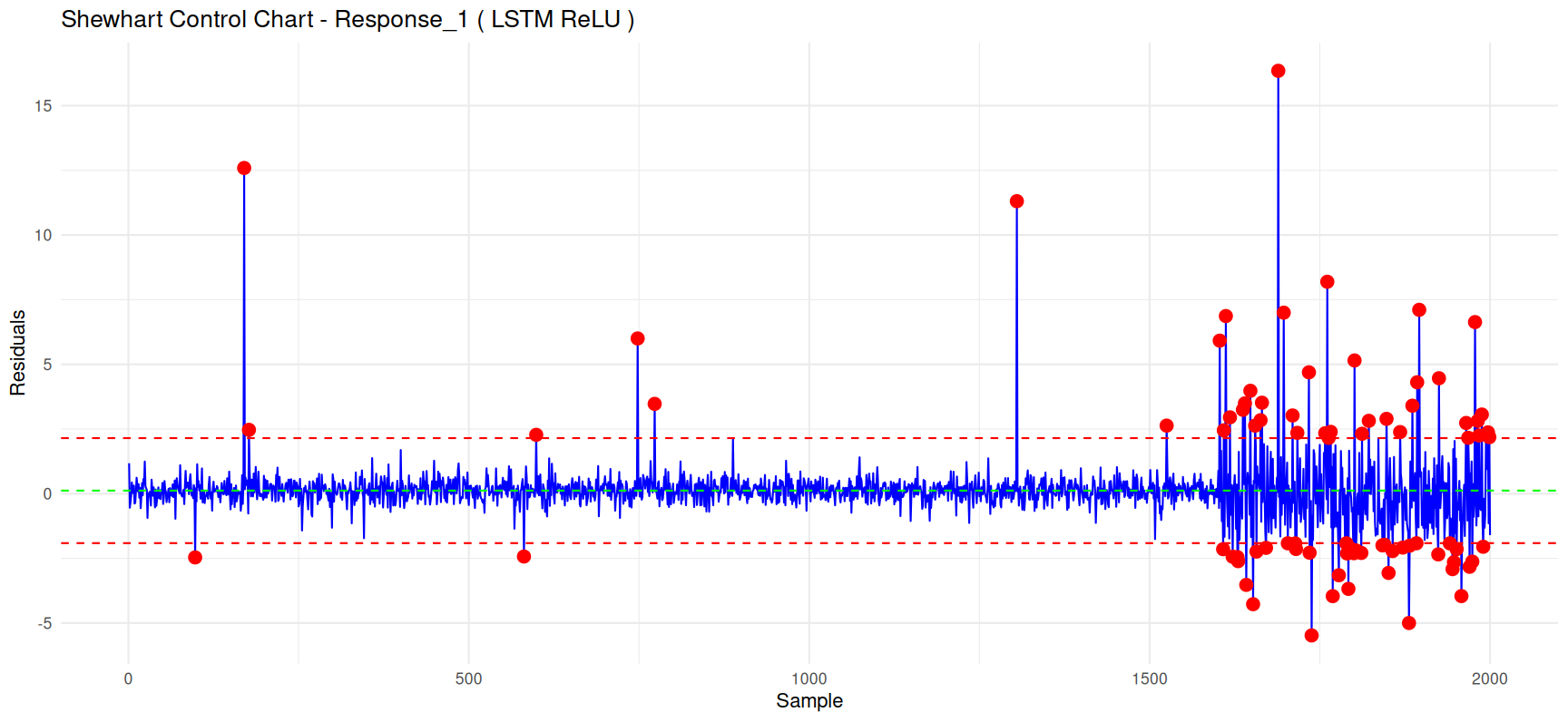}
        \subcaption{LSTM ReLU $Y_1$}\label{fig:fig4}
    \end{minipage}%
    \begin{minipage}{0.33\textwidth}
        \centering
        \includegraphics[width=\linewidth]{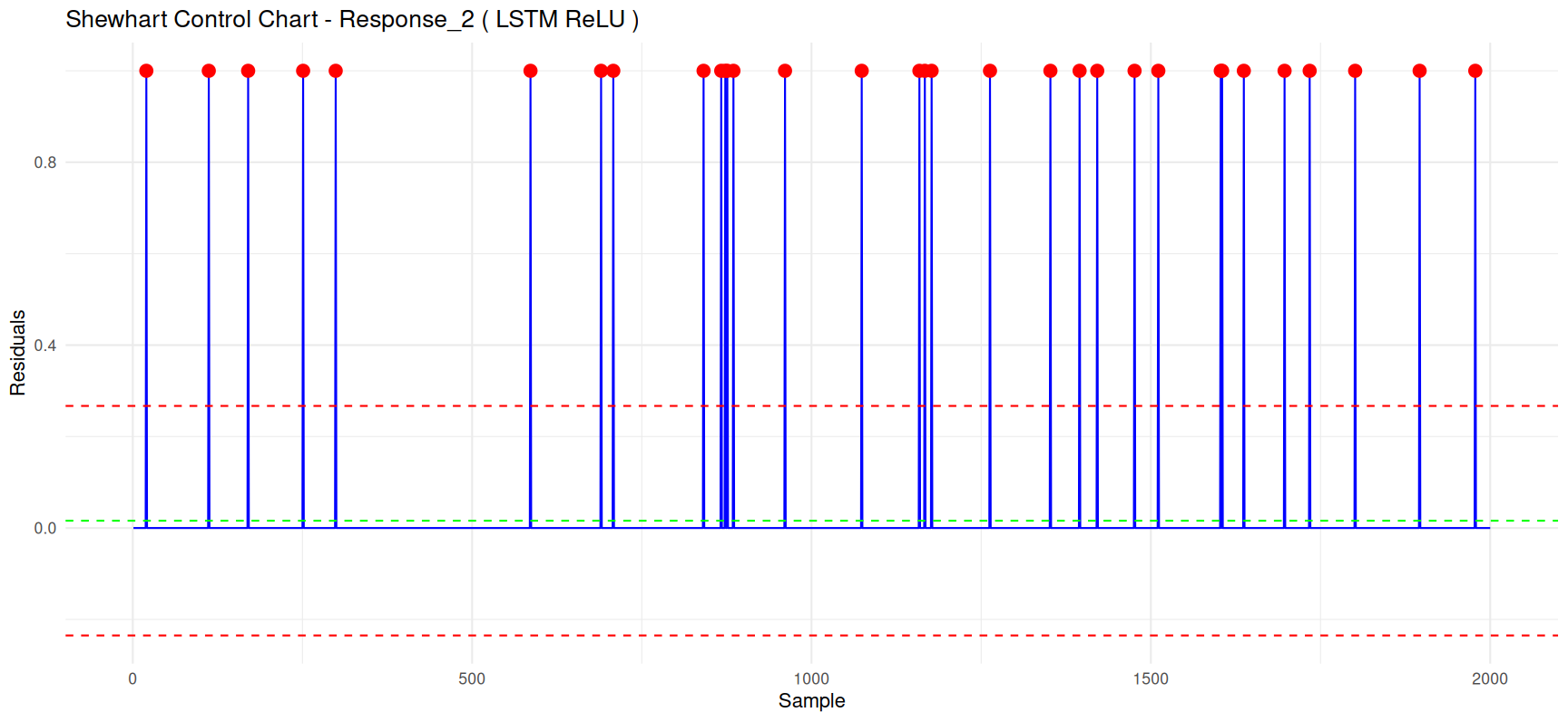}
        \subcaption{LSTM ReLU $Y_2$}\label{fig:fig5}
    \end{minipage}%
    \begin{minipage}{0.33\textwidth}
        \centering
        \includegraphics[width=\linewidth]{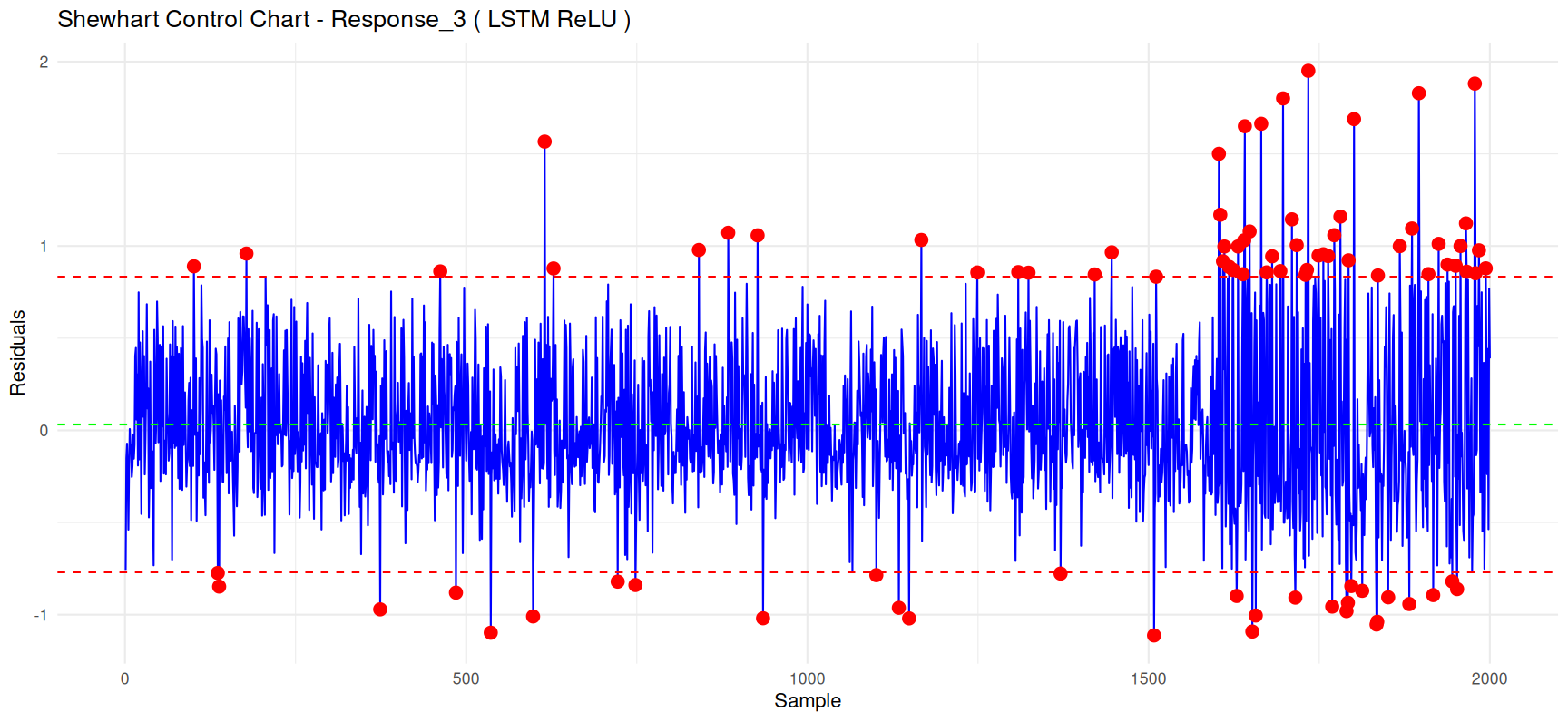}
        \subcaption{LSTM ReLU $Y_3$}\label{fig:fig6}
    \end{minipage} \\[1ex]
    \begin{minipage}{0.33\textwidth}
        \centering
        \includegraphics[width=\linewidth]{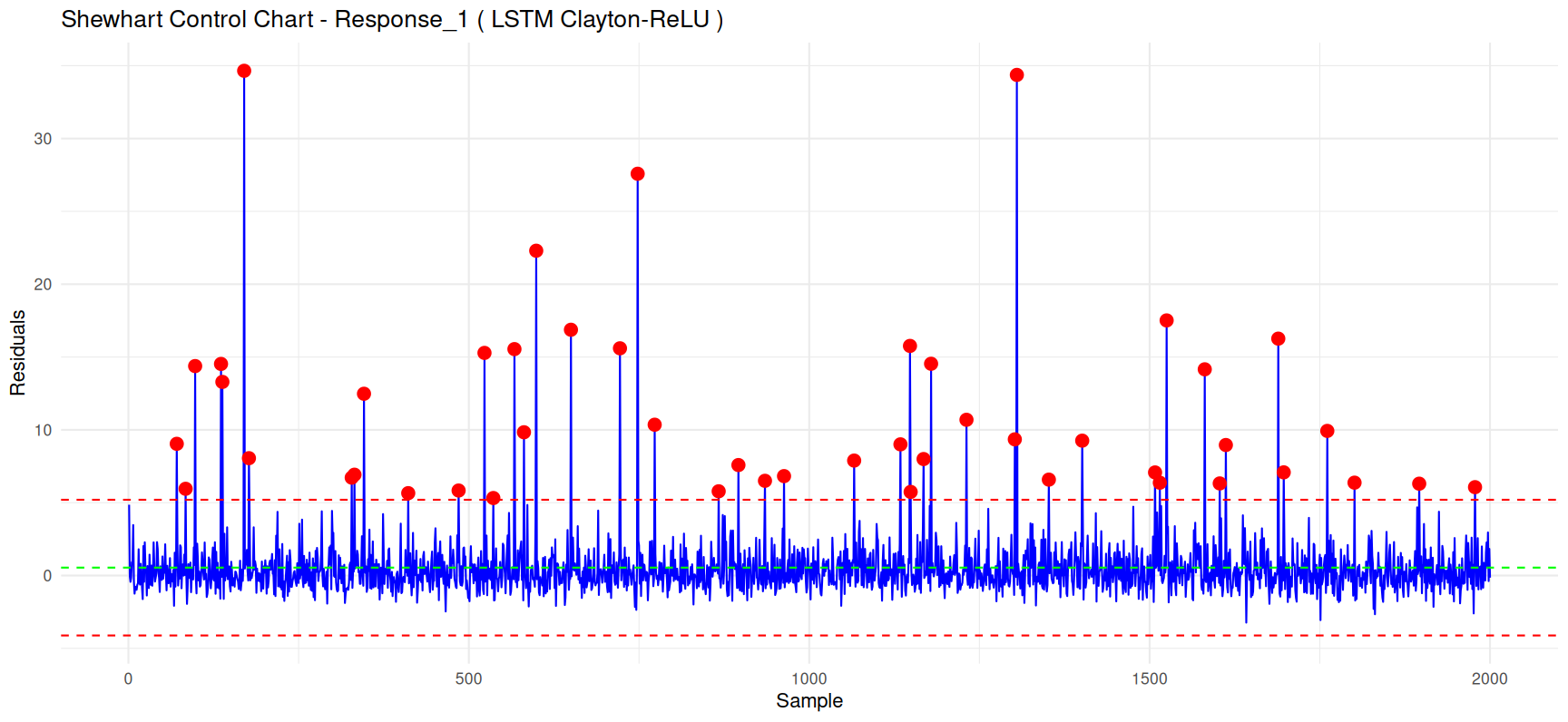}
        \subcaption{LSTM Clayton-ReLU $Y_1$}\label{fig:fig4}
    \end{minipage}%
    \begin{minipage}{0.33\textwidth}
        \centering
        \includegraphics[width=\linewidth]{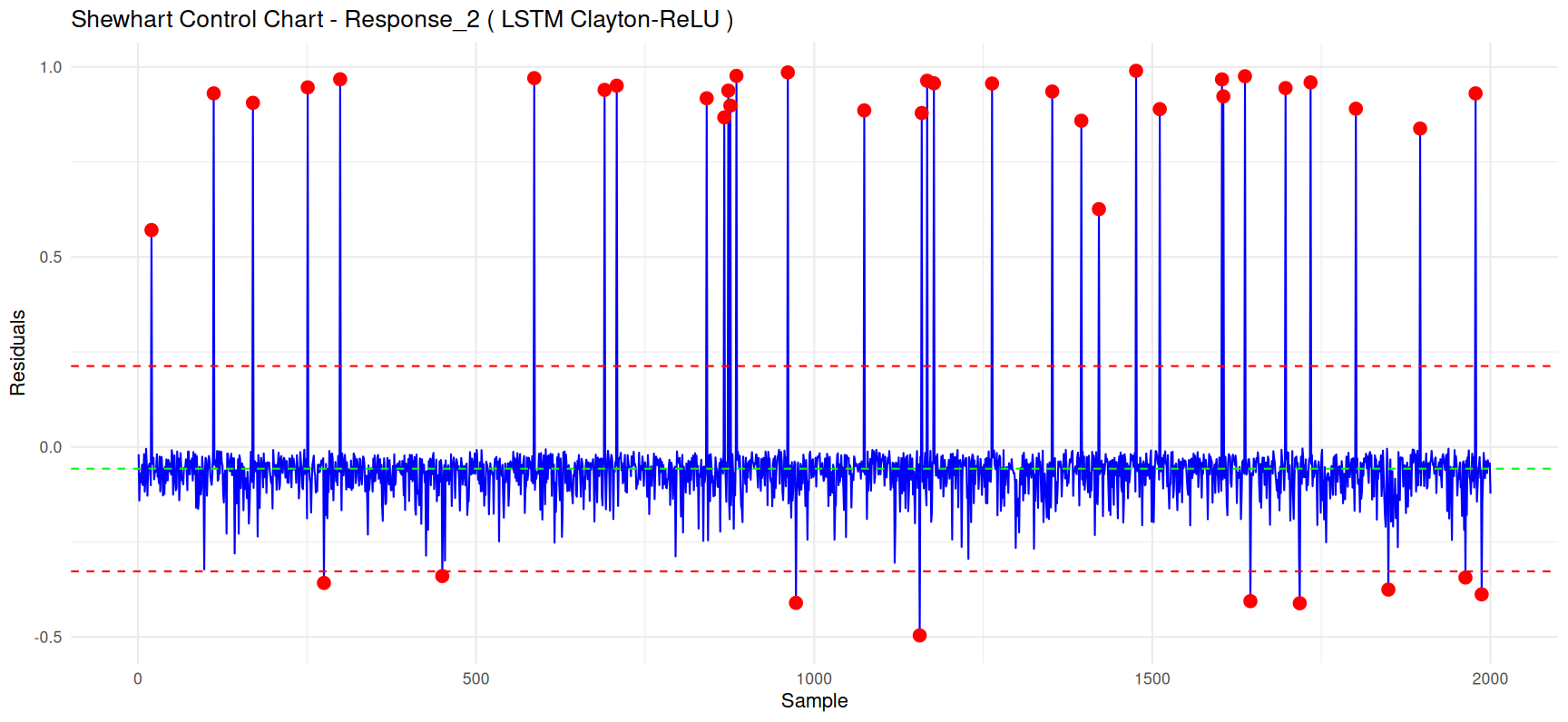}
        \subcaption{LSTM Clayton-ReLU $Y_2$}\label{fig:fig5}
    \end{minipage}%
    \begin{minipage}{0.33\textwidth}
        \centering
        \includegraphics[width=\linewidth]{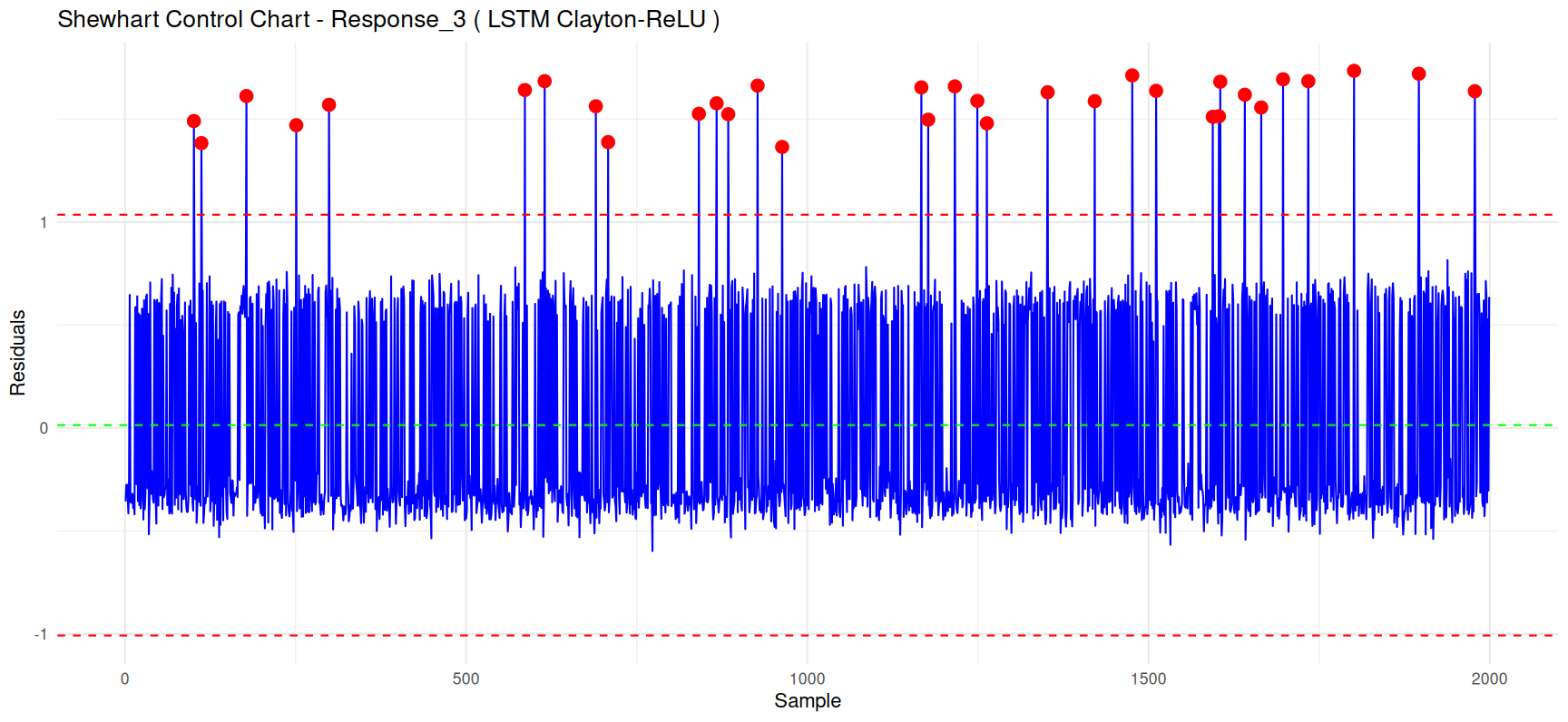}
        \subcaption{LSTM Clayton-ReLU $Y_3$}\label{fig:fig6}
    \end{minipage} \\[1ex]
    \begin{minipage}{0.33\textwidth}
        \centering
        \includegraphics[width=\linewidth]{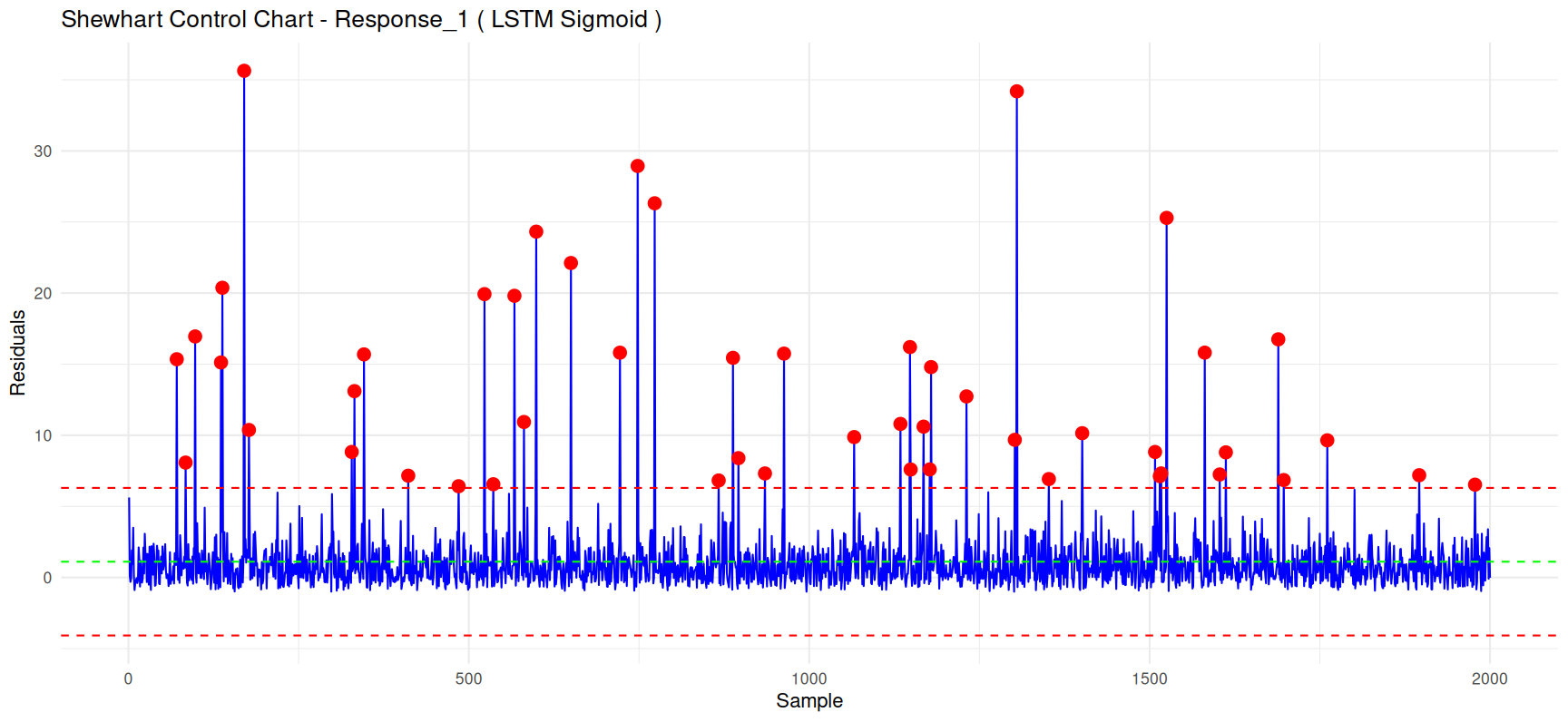}
        \subcaption{LSTM Sigmoid $Y_1$}\label{fig:fig4}
    \end{minipage}%
    \begin{minipage}{0.33\textwidth}
        \centering
        \includegraphics[width=\linewidth]{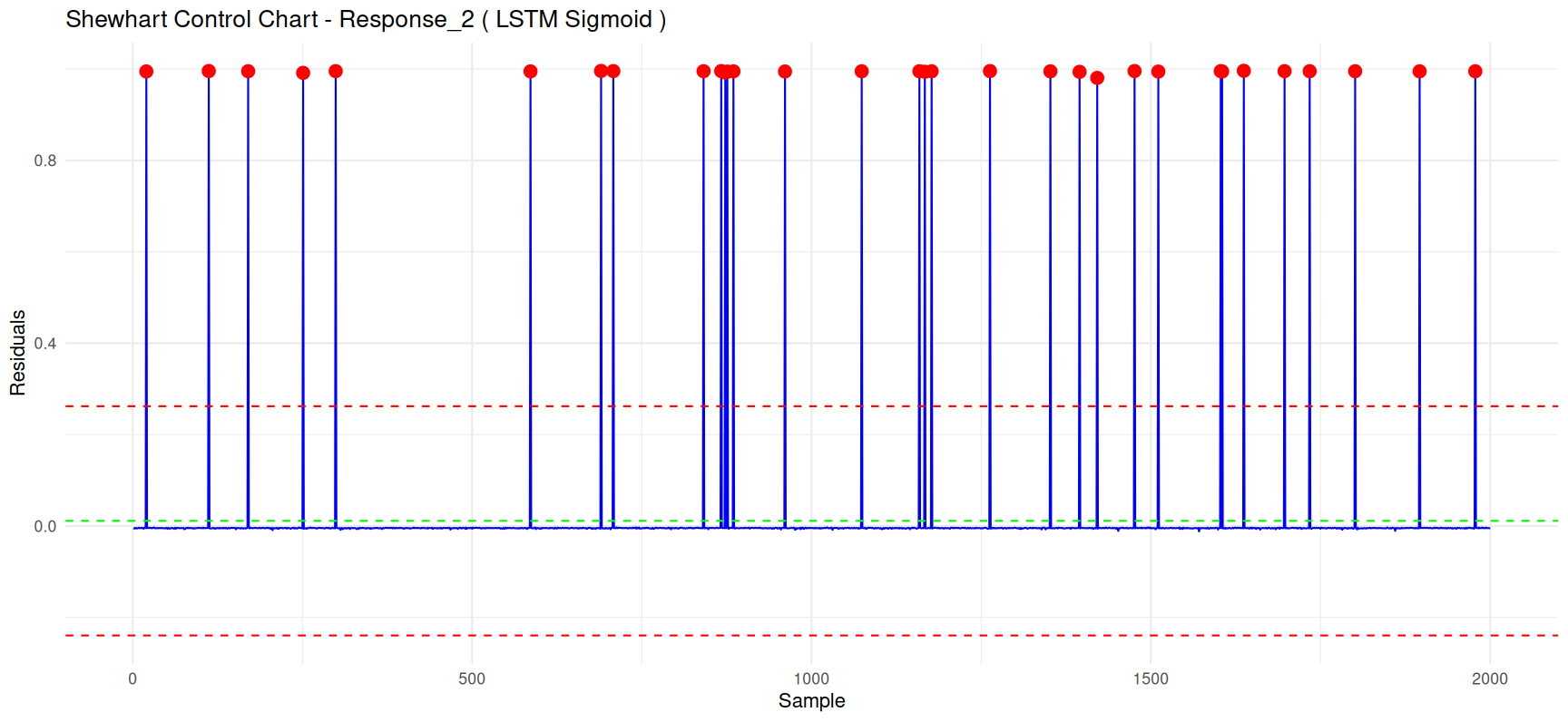}
        \subcaption{LSTM Sigmoid $Y_2$}\label{fig:fig5}
    \end{minipage}%
    \begin{minipage}{0.33\textwidth}
        \centering
        \includegraphics[width=\linewidth]{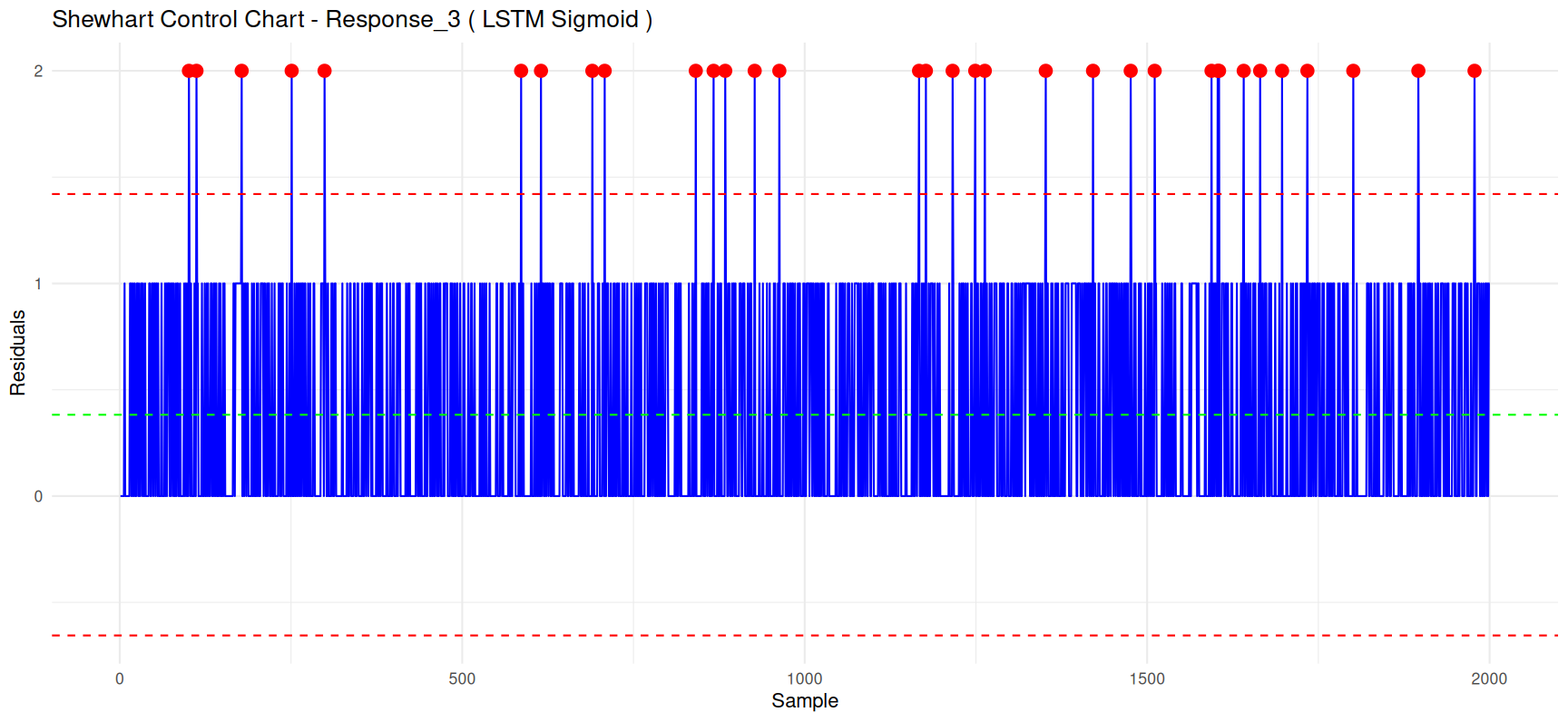}
        \subcaption{LSTM Sigmoid $Y_3$}\label{fig:fig6}
    \end{minipage} \\[1ex]
    \caption{Residual Shewhart Control Charts of LSTM Models with Simulated Data.}
    \label{fig:allfigures2}
\end{figure}

From the LSTM and CNN-LSTM models based on residual Shewhart control
charts in Figure \ref{fig:allfigures1} and Figure
\ref{fig:allfigures2}, we notice that the CNN-LSTM models with
Copula-based activations outperformed LSTM models with copula-based
activations in terms of residual variability and the frequency of
outliers in Response\_1, Response\_2, and Response\_3. However, both
CNN-LSTM and LSTM models with ReLU and Sigmoid activations exhibited
instability in the binary variable $Y_2$ and categorical variable
$Y_3$. Notably, CNN-LSTM with Clayton-ReLU activation improved
residual stability, reducing extreme shifts in the control chart in
$Y_2$ and $Y_3$. Our findings suggest that hybrid copula-based
Clayton-ReLU activation mitigates instability issues more
effectively than models using ReLU or Sigmoid activations for binary
variable Response\_2 and categorical variable Response\_3. Based on
this simulation study, we recommend the CNN-LSTM model with
Clayton-ReLU activation for analyzing multivariate response data,
including continuous, binary, and categorical variables.

\begin{table}[]
\center     \caption{Model Comparison with Simulated Data.}
 \label{table:1} \resizebox{1.0\textwidth}{!}{
\begin{tabular}{|c|c|c|c|c|c|}
\hline \textbf{Model}          & \textbf{Response} &
\textbf{Mean\_Residual} & \textbf{SD\_Residual} & \textbf{Mean\_ARL}
& \textbf{SD\_ARL} \\ \hline CNN-LSTM Clayton        & Response\_1 &
0.3674                  & 2.0058                & 46.0952 & 16.8473
\\ \hline CNN-LSTM Clayton        & Response\_2 & -0.1311
& 0.1234                & 79.2500 & 16.8473          \\ \hline
CNN-LSTM Clayton        & Response\_3 & 0.0427                  &
0.4955                & 57.4706 & 16.8473          \\ \hline
CNN-LSTM Gumbel         & Response\_1 & 1.1632                  &
2.4861                & 42.0870 & 26.8608          \\ \hline
CNN-LSTM Gumbel         & Response\_2 & 0.0081                  &
0.0951                & 95.1000 & 26.8608          \\ \hline
CNN-LSTM Gumbel         & Response\_3 & 0.3886                  &
0.5174                & 61.0625 & 26.8608          \\ \hline
CNN-LSTM Clayton-Gumbel & Response\_1 & 0.9159                  &
2.4122                & 40.3333 & 12.6967          \\ \hline
CNN-LSTM Clayton-Gumbel & Response\_2 & -0.0880                 &
0.1188                & 63.4000 & 12.6967          \\ \hline
CNN-LSTM Clayton-Gumbel & Response\_3 & 0.0456                  &
0.5057                & 61.0625 & 12.6967          \\ \hline
CNN-LSTM ReLU           & Response\_1 & 0.5227                  &
1.2212                & 31.2258 & 39.8169          \\ \hline
CNN-LSTM ReLU           & Response\_2 & 0.0100                  &
0.0995                & 95.1000 & 39.8169          \\ \hline
CNN-LSTM ReLU           & Response\_3 & 0.0585                  &
0.3712                & 21.9778 & 39.8169          \\ \hline
CNN-LSTM Clayton-ReLU   & Response\_1 & 0.4347                  &
1.7965                & 38.7200 & 11.1716          \\ \hline
CNN-LSTM Clayton-ReLU   & Response\_2 & -0.1551                 &
0.1303                & 50.0526 & 11.1716          \\ \hline
CNN-LSTM Clayton-ReLU   & Response\_3 & 0.0562                  &
0.4892                & 61.0625 & 11.1716          \\ \hline
CNN-LSTM Sigmoid        & Response\_1 & 1.2055                  &
2.4707                & 42.0870 & 26.8608          \\ \hline
CNN-LSTM Sigmoid        & Response\_2 & -0.0071                 &
0.1015                & 95.1000 & 26.8608          \\ \hline
CNN-LSTM Sigmoid        & Response\_3 & 0.3958                  &
0.5141                & 61.0625 & 26.8608          \\ \hline LSTM
Clayton            & Response\_1 & 0.4908                  & 2.3292
& 38.7843 & 10.6910          \\ \hline LSTM Clayton            &
Response\_2 & -0.0414                 & 0.1321                &
52.0526 & 10.6910          \\ \hline LSTM Clayton            &
Response\_3 & 0.0093                  & 0.5065                &
59.9394 & 10.6910          \\ \hline LSTM Gumbel             &
Response\_1 & 1.1114                  & 2.5938                &
39.5600 & 18.2995          \\ \hline LSTM Gumbel             &
Response\_2 & 0.0131                  & 0.1133                &
76.0769 & 18.2995          \\ \hline LSTM Gumbel             &
Response\_3 & 0.3825                  & 0.5190                &
59.9394 & 18.2995          \\ \hline LSTM Clayton-Gumbel     &
Response\_1 & 0.6306                  & 2.3563                &
38.7843 & 11.0577          \\ \hline LSTM Clayton-Gumbel     &
Response\_2 & -0.0153                 & 0.1307                &
54.9444 & 11.0577          \\ \hline LSTM Clayton-Gumbel     &
Response\_3 & 0.0337                  & 0.5074                &
59.9394 & 11.0577          \\ \hline LSTM ReLU               &
Response\_1 & 0.1190                  & 1.0148                &
24.0843 & 22.2052          \\ \hline LSTM ReLU               &
Response\_2 & 0.0160                  & 0.1255                &
61.8125 & 22.2052          \\ \hline LSTM ReLU               &
Response\_3 & 0.0316                  & 0.4009                &
22.6591 & 22.2052          \\ \hline LSTM Clayton-ReLU       &
Response\_1 & 0.5436                  & 2.3281                &
41.2083 & 9.4445           \\ \hline LSTM Clayton-ReLU       &
Response\_2 & -0.0573                 & 0.1350                &
48.4634 & 9.4445           \\ \hline LSTM Clayton-ReLU       &
Response\_3 & 0.0138                  & 0.5107                &
59.9394 & 9.4445           \\ \hline LSTM Sigmoid            &
Response\_1 & 1.1114                  & 2.5938                &
39.5600 & 12.3424          \\ \hline LSTM Sigmoid            &
Response\_2 & 0.0112                  & 0.1254                &
61.8125 & 12.3424          \\ \hline LSTM Sigmoid            &
Response\_3 & 0.3826                  & 0.5190                &
59.9394 & 12.3424          \\ \hline
\end{tabular}
}
\end{table}

Table \ref{table:1} presents a comparison of CNN-LSTM and LSTM
models using different activation functions and copula methods for
three response variables. The evaluation is based on four key
metrics: mean residual, standard deviation of residuals, mean ARL,
and standard deviation of ARL. The mean residual represents the
average difference between predicted and actual values, while the
standard deviation of residuals indicates the variability in
prediction errors. Mean ARL measures how often the model detects
changes, with higher values indicating fewer false alarms. The
standard deviation of ARL reflects the consistency of change
detection.

\begin{figure}[htp]
    \centering
    \begin{minipage}{0.8\textwidth}
        \centering
        \includegraphics[width=\linewidth]{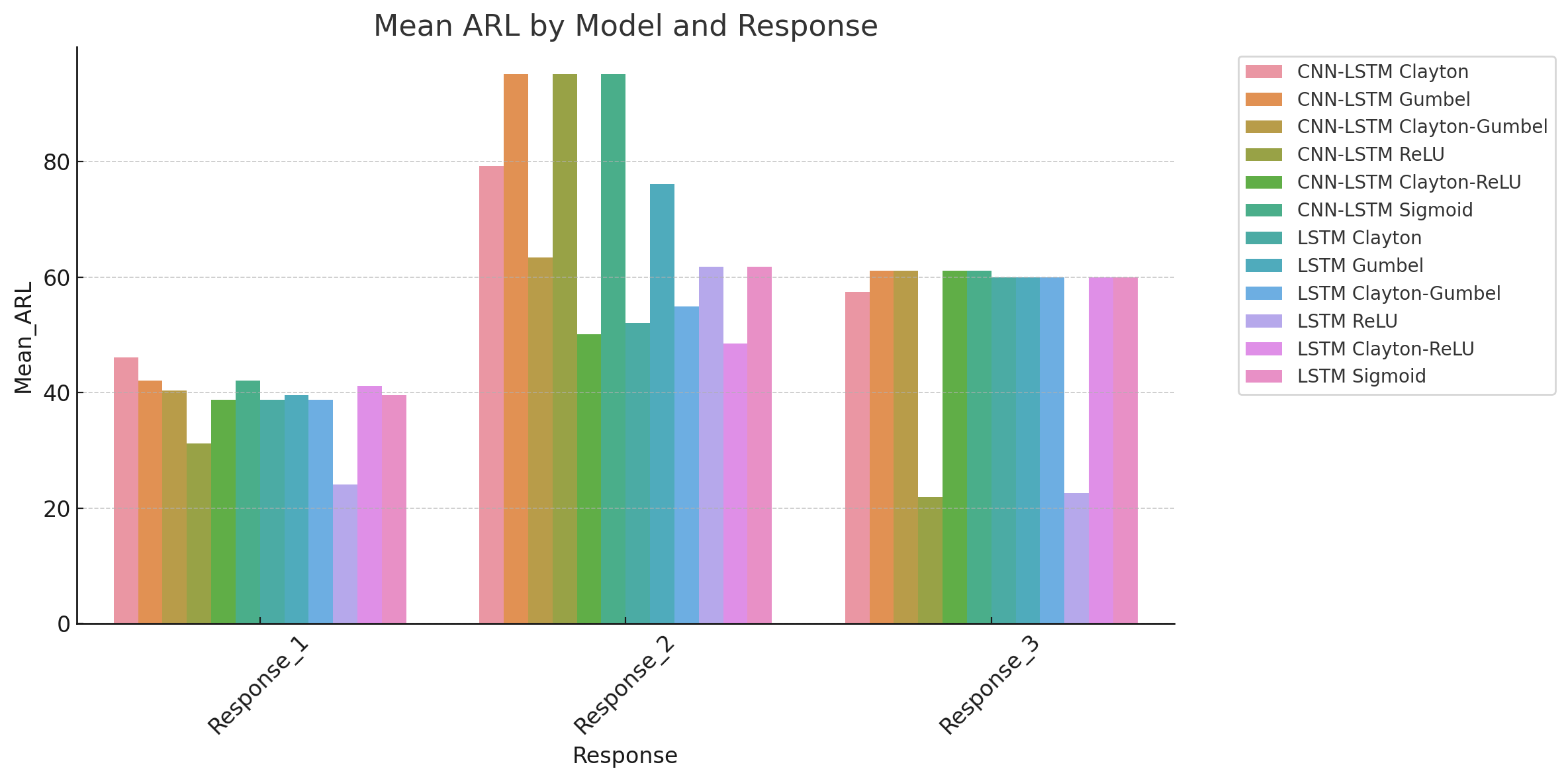}
    \end{minipage} \\[1ex]
    \begin{minipage}{0.8\textwidth}
        \centering
        \includegraphics[width=\linewidth]{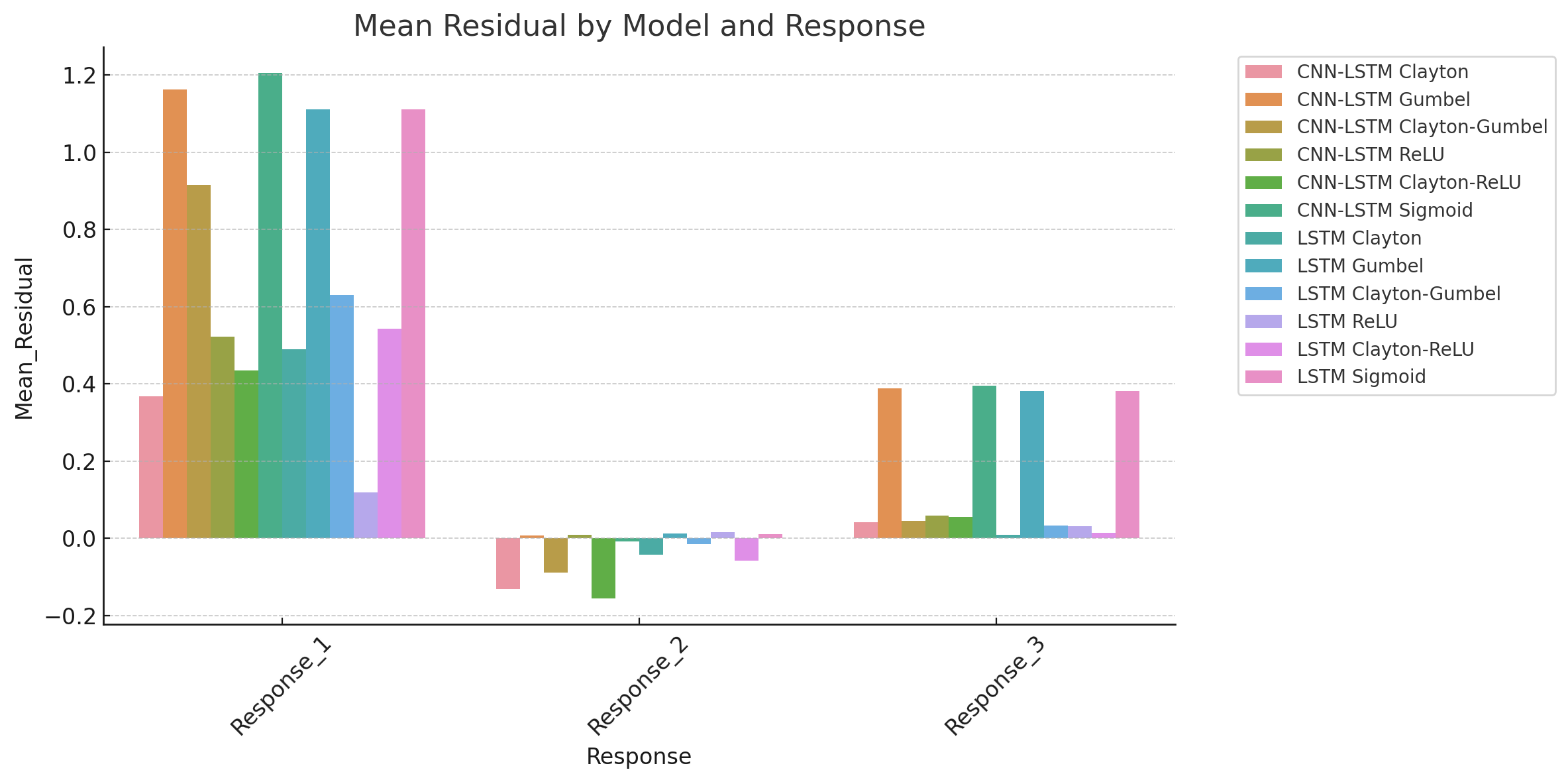}
    \end{minipage} \\[1ex]
    \begin{minipage}{0.8\textwidth}
        \centering
        \includegraphics[width=\linewidth]{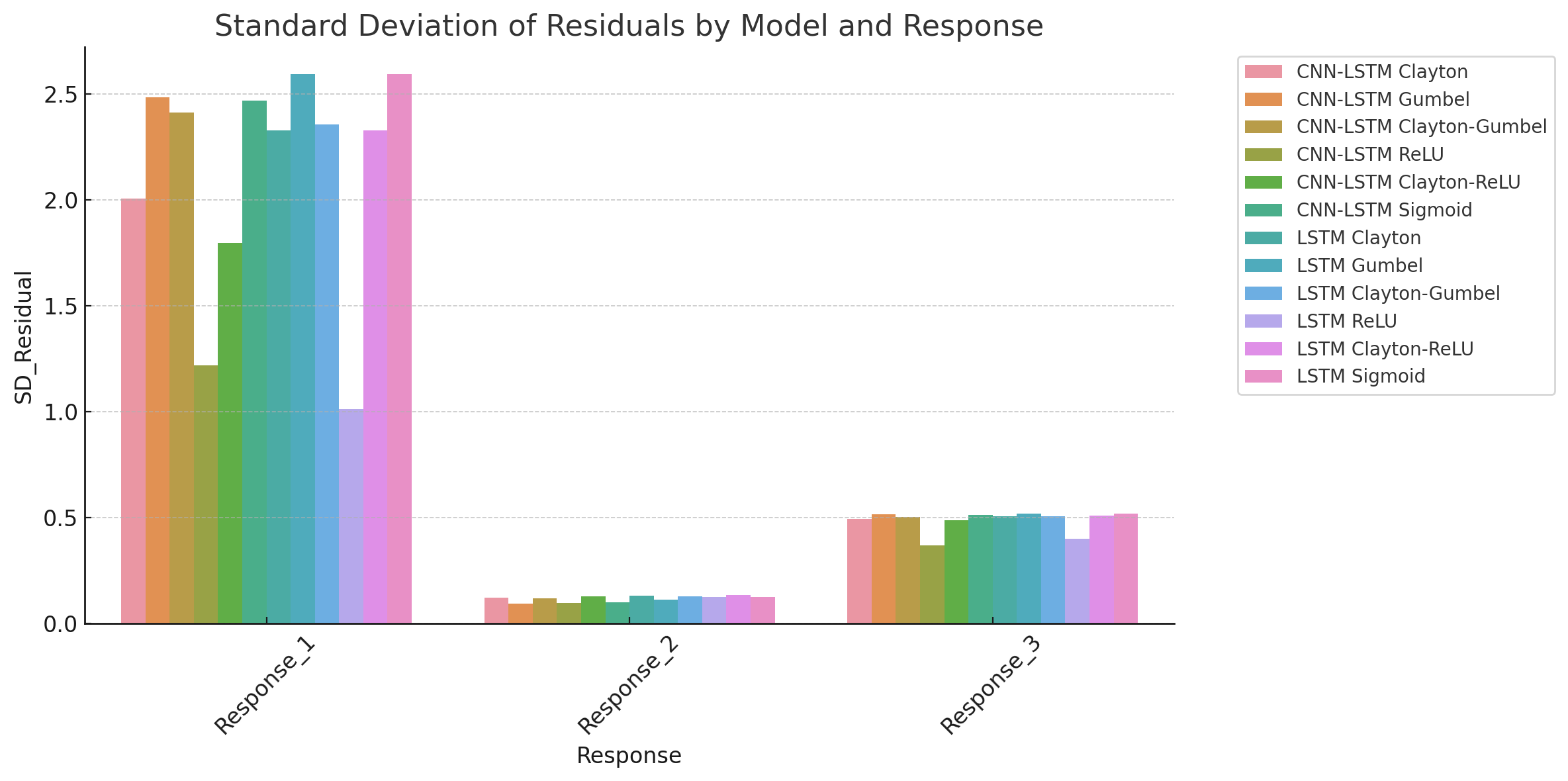}
    \end{minipage} \\[1ex]
    \caption{{Mean ARL, Mean Residual and Standard Deviation of Residuals by Model
and Response with Simulated Data.}}
    \label{fig:allfigtable1}
\end{figure}

Figure\ref{fig:allfigtable1} summarizing mean ARL values from Table
1 across different models and response types. This visualization
helps compare how each model performs across Response\_1,
Response\_2, and Response\_3 in terms of ARL. In addition, there are
two additional bar graphs for Mean Residual by Model and Response
and Standard Deviation of Residuals by Model and Response in
Figure\ref{fig:allfigtable1}. Mean Residual by Model and Response
highlights the average error for each model across the three
response variables. Standard Deviation of Residuals by Model and
Response indicates the spread or variability of the residuals for
each model and response. CNN-LSTM Gumbel and CNN-LSTM Sigmoid
exhibit particularly high residuals for Response\_1, exceeding 1.1,
which indicates challenges in capturing this response accurately.
For Response\_2, CNN-LSTM Clayton-ReLU and CNN-LSTM Clayton show the
lowest residual bias, with values of -0.1551 and -0.1311,
respectively. The standard deviation of residuals remains relatively
small, ranging between 0.1 and 0.13, reflecting stable predictions
across models. The lowest bias for Response\_3 is observed in
CNN-LSTM Clayton-Gumbel and LSTM Clayton, with values of 0.0456 and
0.0093, respectively. The standard deviation of residuals remains
below 0.52, indicating consistent performance across copula methods.

A higher mean ARL suggests that a model detects fewer false alarms.
CNN-LSTM ReLU and CNN-LSTM Sigmoid achieve the highest mean ARL
values for Response\_2, both at 95.1, indicating strong stability.
However, as shown in Figure \ref{fig:allfigures1} and Figure
\ref{fig:allfigures2}, both CNN-LSTM and LSTM models with ReLU and
Sigmoid activations exhibit instability in the binary variable
Response\_2 and categorical variable Response\_3. Therefore, we
conclude that the mean ARL values for Response\_2 with CNN-LSTM ReLU
and CNN-LSTM Sigmoid may not be trustworthy. In contrast, LSTM ReLU
has the lowest mean ARL for Response\_1 at 24.08, indicating it
reacts more quickly to changes. This is consistent with its
stability in the residual Shewhart chart in Figure
\ref{fig:allfigures2}. The ReLU activation enhances model stability,
particularly for CNN-LSTM models for Response\_1. Clayton-ReLU
Copulas reduce residuals, making them suitable choices for
Response\_2.

Overall, CNN-LSTM captures complex dependencies more effectively but
exhibits higher residual fluctuations, particularly for Response\_1.
LSTM models provide more stable predictions with lower residual
variance. For Response\_2 and Response\_3, Clayton-ReLU activation
functions yield the most stable performance.

\section{Breast Cancer Real Data Analysis}

Breast Cancer Gene Expression Profiles (METABRIC), likely includes
gene expression data related to breast cancer, where each sample
corresponds to a breast cancer patient, and the features represent
the expression levels of various genes. It likely contains a variety
of columns with features such as: Gene expression levels: These are
numerical values representing the expression levels of different
genes in each patient. Patient metadata: Includes attributes like
age, tumor stage, survival status, etc. Event time and censoring
information: For survival analysis, you might have columns for the
time until the event (e.g., death or recurrence) and a censoring
indicator (1 = event occurred, 0 = censored). With this data, we can
predict survival times using the gene expression data and clinical
features, classify patients into different groups (e.g., tumor
stage, survival status), identify genes whose expression levels
differ significantly between different subgroups (e.g., cancerous
vs. normal), and investigate the relationship between gene
expression levels and responses to specific cancer treatments. In
this research, we use the METABRIC (Molecular Taxonomy of Breast
Cancer International Consortium) dataset, which contains clinical
and genomic data for breast cancer patients. The preprocessing
pipeline is designed to prepare the data for survival analysis,
where the goal is to predict patient survival time and event status
(death or censored) using clinical features and gene expression
data. We downloaded the METABRIC dataset from \href{
https://www.kaggle.com/datasets/raghadalharbi/breast-cancer-gene-expression-profiles-metabric?resource=download}
{the Kaggle website}. The dataset contains both clinical variables
(e.g., survival times, tumor stage) and genomic features (e.g., gene
expression levels). The file path to the dataset must be specified,
and once loaded, the dataset is examined to understand its
structure, including the names of the columns.

\begin{table}[htbp]
\centering \caption{{METABRIC Data Summary Statistics}}
 \label{table:2} \resizebox{1.0\textwidth}{!}{
\begin{tabular}{|c|l|p{5.5cm}|p{5.5cm}|c|c|}
\hline
\textbf{No} & \textbf{Variable} & \textbf{Stats / Values} & \textbf{Freqs (\% of Valid)} & \textbf{Valid} & \textbf{Missing} \\
\hline 1 & \texttt{survival\_time} [numeric] & Mean (sd): 127.7
(78.5) \newline Min = Med = Max: 0.1 = 117.7 = 351 \newline IQR
(CV): 128.1 (0.6) \newline 1199 distinct values & -- &
1310 (100.0\%) & 0 (0.0\%) \\
\hline 2 & \texttt{event\_status} [numeric] & 1 distinct value & 0:
1310 (100.0\%) &
1310 (100.0\%) & 0 (0.0\%) \\
\hline 3 & \texttt{age} [numeric] & Mean (sd): 60.3 (13) \newline
Min = Med = Max: 21.9 = 61 = 96.3 \newline IQR (CV): 19 (0.2)
\newline 1141 distinct values & -- &
1310 (100.0\%) & 0 (0.0\%) \\
\hline 4 & \texttt{tumor\_stage} [numeric] & Mean (sd): 1.8 (0.6)
\newline Min = Med = Max: 1 = 2 = 4 \newline IQR (CV): 1 (0.4) & 1:
442 (33.7\%) \newline 2: 752 (57.4\%) \newline 3: 108 (8.2\%)
\newline 4: 8 (0.6\%) &
1310 (100.0\%) & 0 (0.0\%) \\
\hline 5 & \texttt{er\_status} [character] & 1. Negative \newline 2.
Positive & Negative: 303 (23.1\%) \newline Positive: 1007 (76.9\%) &
1310 (100.0\%) & 0 (0.0\%) \\
\hline 6 & \texttt{her2\_status} [character] & 1. Negative \newline
2. Positive & Negative: 1149 (87.7\%) \newline Positive: 161
(12.3\%) &
1310 (100.0\%) & 0 (0.0\%) \\
\hline
\end{tabular}
}
\end{table}

{Table 2 provides a detailed summary of six key variables
extracted from the METABRIC (Molecular Taxonomy of Breast Cancer
International Consortium) dataset. The dataset comprises information
for 1,310 patients, with no missing values for any of the selected
variables. The statistics presented include measures of central
tendency and variability for numeric variables, and frequency
distributions for categorical variables.}

{The variable \texttt{survival\_time} captures the follow-up
time (in months) until either death or censoring. The average
survival time is 127.7 months, with a standard deviation of 78.5
months, indicating substantial variability in patient outcomes. The
median survival is 117.7 months, which closely approximates the
interquartile range (IQR) of 128.1 months, suggesting a right-skewed
distribution. The minimum and maximum observed survival times are
0.1 and 351 months, respectively. The coefficient of variation (CV),
a standardized measure of dispersion, is 0.6, and there are 1,199
unique values, demonstrating a high level of granularity in this
continuous variable.}

{The \texttt{event\_status} variable is coded numerically,
though only one distinct value (0) appears in this subset,
indicating that all patients in this subset were censored and no
events (deaths) were recorded. All 1,310 entries are valid, with no
missing data. This suggests either a data coding issue or that the
sample was filtered to include only censored individuals.}

{The \texttt{age} variable captures the patient's age at
diagnosis. The average age is 60.3 years, with a standard deviation
of 13 years. The observed ages range from 21.9 to 96.3 years, with a
median of 61 years and an IQR of 19 years. The CV is 0.2, indicating
low relative variability. There are 1,141 unique age values,
suggesting minimal rounding or grouping of age measurements.}

{The \texttt{tumor\_stage} variable is a numeric representation
of disease progression, ranging from stage 1 to 4. The mean stage is
1.8, with a standard deviation of 0.6, and the median stage is 2.
The IQR is 1, and the CV is 0.4, pointing to moderate dispersion.
The distribution across tumor stages is as follows:}
\begin{itemize}
  \item Stage 1: 442 patients (33.7\%)
  \item Stage 2: 752 patients (57.4\%)
  \item Stage 3: 108 patients (8.2\%)
  \item Stage 4: 8 patients (0.6\%)
\end{itemize}
This indicates that the majority of patients were diagnosed at early
stages, particularly stage 2.

The \texttt{er\_status} is a binary categorical variable indicating
the presence (Positive) or absence (Negative) of estrogen receptors
in tumor cells. Among the 1,310 patients:
\begin{itemize}
  \item Positive ER status: 1,007 patients (76.9\%)
  \item Negative ER status: 303 patients (23.1\%)
\end{itemize}
This shows that the majority of tumors were ER-positive, which is
clinically significant for treatment decisions involving hormone
therapy.

Similarly, \texttt{her2\_status} categorizes the presence of human
epidermal growth factor receptor 2 (HER2). The distribution is as
follows:
\begin{itemize}
  \item Negative HER2 status: 1,149 patients (87.7\%)
  \item Positive HER2 status: 161 patients (12.3\%)
\end{itemize}
HER2-positive tumors often require different therapeutic strategies,
such as targeted therapies like trastuzumab.

The METABRIC dataset, as summarized in Table 2 is comprehensive and
clean, with no missing data across the selected variables. The
dataset exhibits rich variability in continuous measures such as
survival time and age, and includes clinically relevant categorical
variables such as ER and HER2 status. This robust data foundation
supports rigorous modeling and inferential analysis in the context
of breast cancer prognosis and treatment stratification.

To improve clarity and consistency, the columns of the dataset are
renamed with more descriptive names. Specifically, the survival time
column, overall\_survival\_months, is renamed to survival\_time, and
the event status column, overall\_survival, is renamed to
event\_status. Additional columns such as age\_at\_diagnosis,
tumor\_stage, er\_status, and her2\_status are also renamed for
clarity.

The event\_status column, which originally contains categorical
values (such as ``Dead'' or ``Alive''), is converted into a binary
format. This transformation allows the survival analysis model to
treat the event status as a binary outcome, where 1 indicates that
the patient has died (event occurred), and 0 indicates that the
patient is either alive or the data is censored (event not
observed).

The dataset may contain missing values for some columns. To ensure
that only complete cases are used for analysis, any rows with
missing data are removed from the dataset. This step is necessary to
ensure the integrity of the data, as missing values can introduce
biases or errors in survival models.

Several key parameters are defined for the model: The 'timesteps'
parameter defines the number of time steps for the model and is set
to a fixed value of 10 in this study. It can be adjusted based on
the specific requirements of the analysis. The number of features
corresponds to the number of columns in the dataset that represent
clinical or genomic variables (e.g., gene expression measurements).
This includes both the clinical features and any other variables
relevant to the survival analysis. The number of samples refers to
the number of rows in the dataset, where each row represents an
individual patient.

To feed the data into CNN-LSTM and LSTM models, which require
three-dimensional input, the dataset is reshaped into a 3D array
with dimensions representing (samples, time steps, features). This
transformation ensures that the LSTM model can handle sequential
data appropriately. In practice, this reshaped data would be
replaced with actual gene expression data or other time-series
features if available, rather than random values used for
illustration.

The input data is normalized to have zero mean and unit variance.
Normalization is an important step when working with deep learning
models, as it helps the model converge faster and avoids issues
caused by variables with different scales. This step ensures that
the features are standardized, making the model more efficient and
stable during training.

After normalization, the data is reshaped again to ensure that it is
in the correct format for input into the LSTM model. The reshaped
data maintains the dimensions required by the model, ensuring
compatibility with the network architecture.

The output labels, which include the survival time and event status,
are selected from the dataset. These labels represent the target
variables for the survival model. The survival\_time (Response\_1)
is a continuous variable representing the time until the event or
censoring, while the event\_status (Response\_2) is a binary
variable indicating whether the event (death) occurred. These labels
are then converted into a matrix format to be fed into the model
during training.

The preprocessing steps described above transform the METABRIC
dataset into a structured and ready-to-use format for survival
analysis. The dataset undergoes cleaning, renaming, normalization,
and reshaping to prepare it for deep learning models, particularly
LSTM networks. These steps ensure that the model receives
appropriately formatted and normalized data for predicting survival
outcomes, which will help in understanding patient prognosis and
improving clinical decision-making in breast cancer research.

\begin{figure}[htp]
    \centering
    \begin{minipage}{0.35\textwidth}
        \centering
        \includegraphics[width=\linewidth]{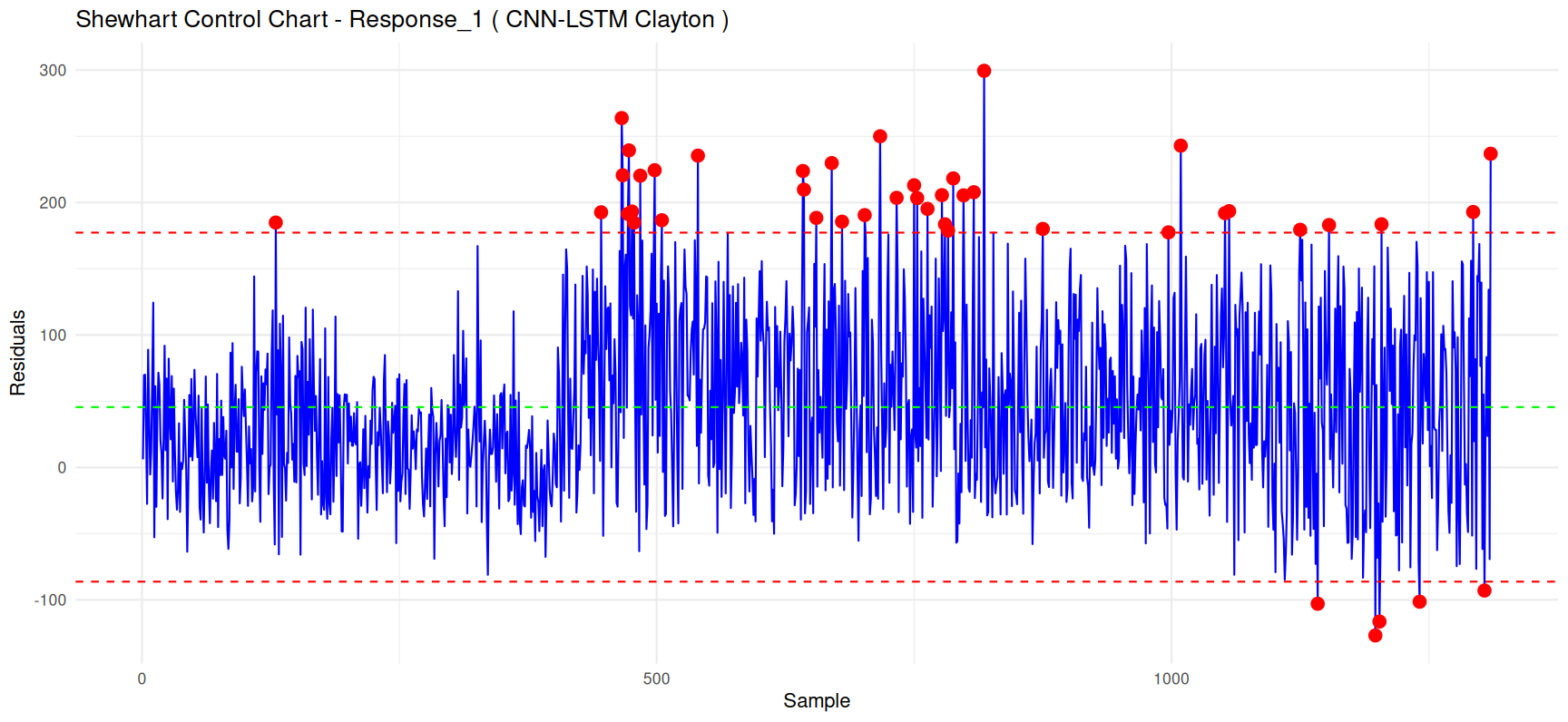}
        \subcaption{CNN-LSTM Clayton $Y_1$}\label{fig:fig1}
    \end{minipage}%
    \begin{minipage}{0.35\textwidth}
        \centering
        \includegraphics[width=\linewidth]{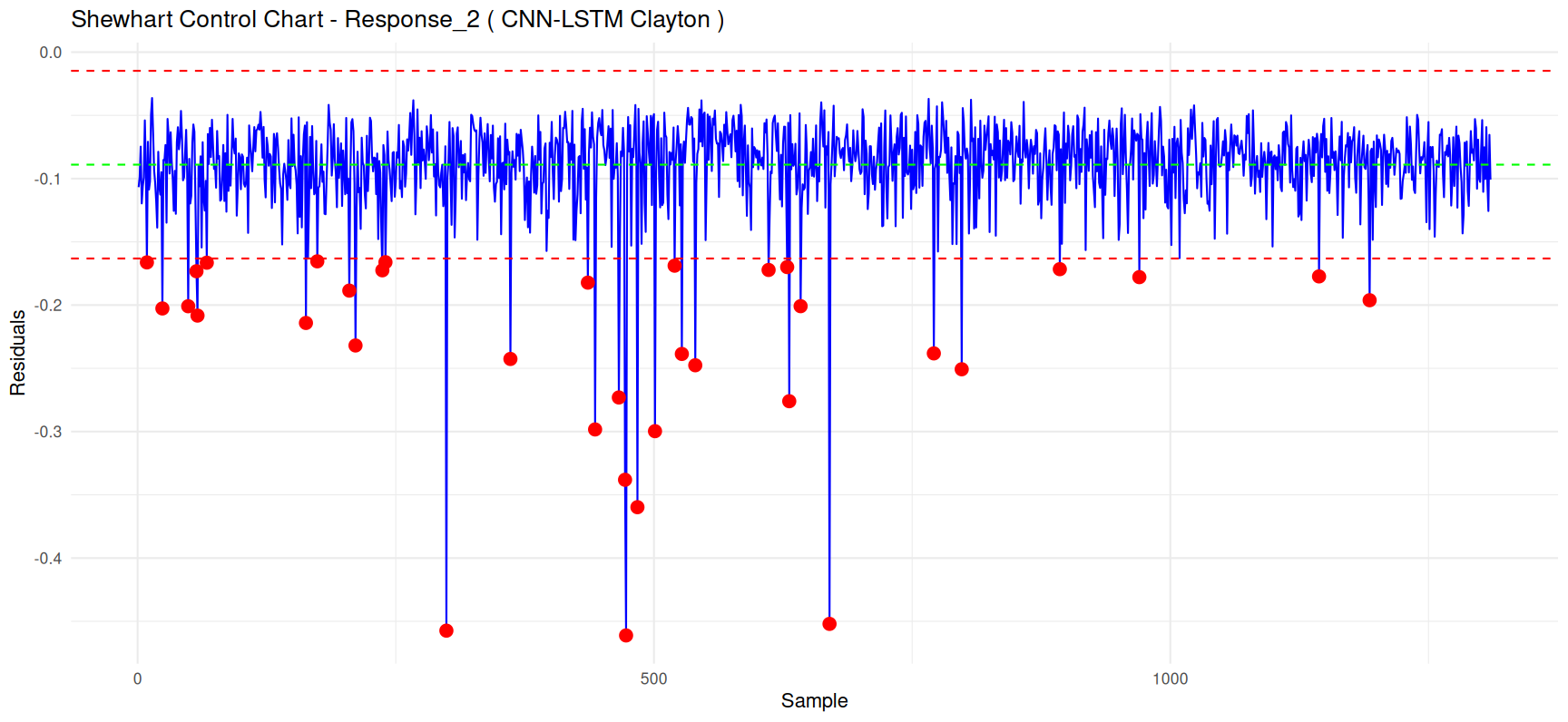}
        \subcaption{CNN-LSTM Clayton $Y_2$}\label{fig:fig2}
    \end{minipage} \\[1ex]
    \begin{minipage}{0.35\textwidth}
        \centering
        \includegraphics[width=\linewidth]{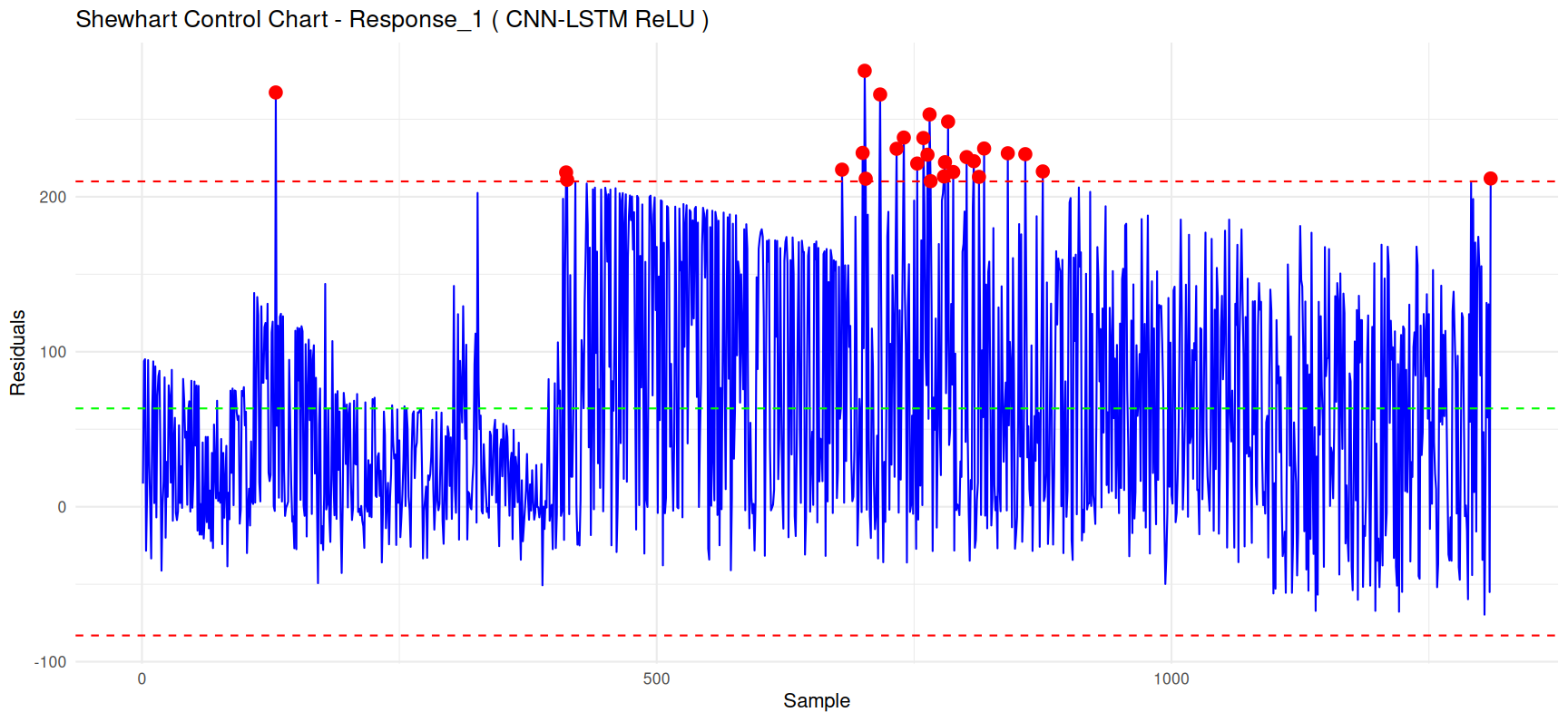}
        \subcaption{CNN-LSTM ReLU $Y_1$}\label{fig:fig4}
    \end{minipage}%
    \begin{minipage}{0.35\textwidth}
        \centering
        \includegraphics[width=\linewidth]{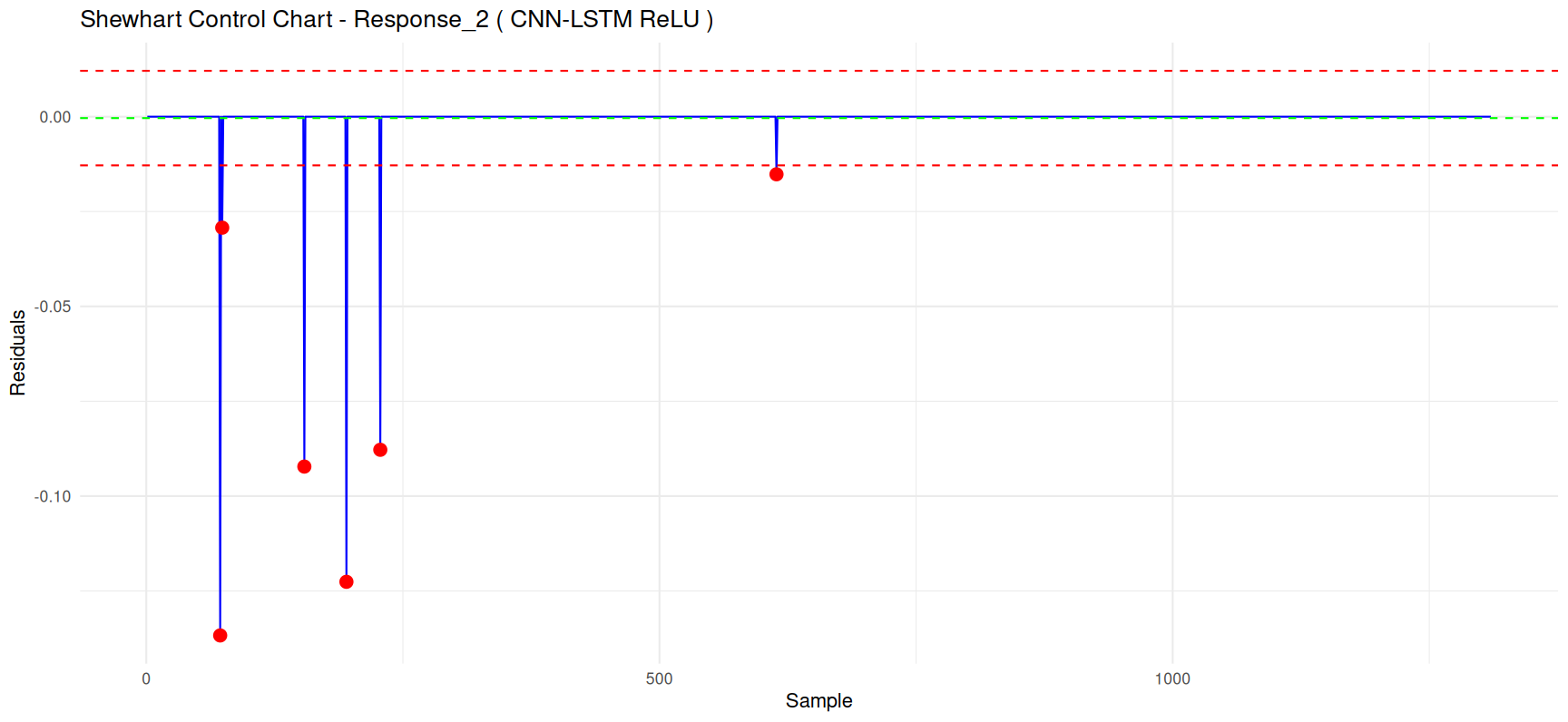}
        \subcaption{CNN-LSTM ReLU $Y_2$}\label{fig:fig5}
    \end{minipage}\\[1ex]
    \begin{minipage}{0.35\textwidth}
        \centering
        \includegraphics[width=\linewidth]{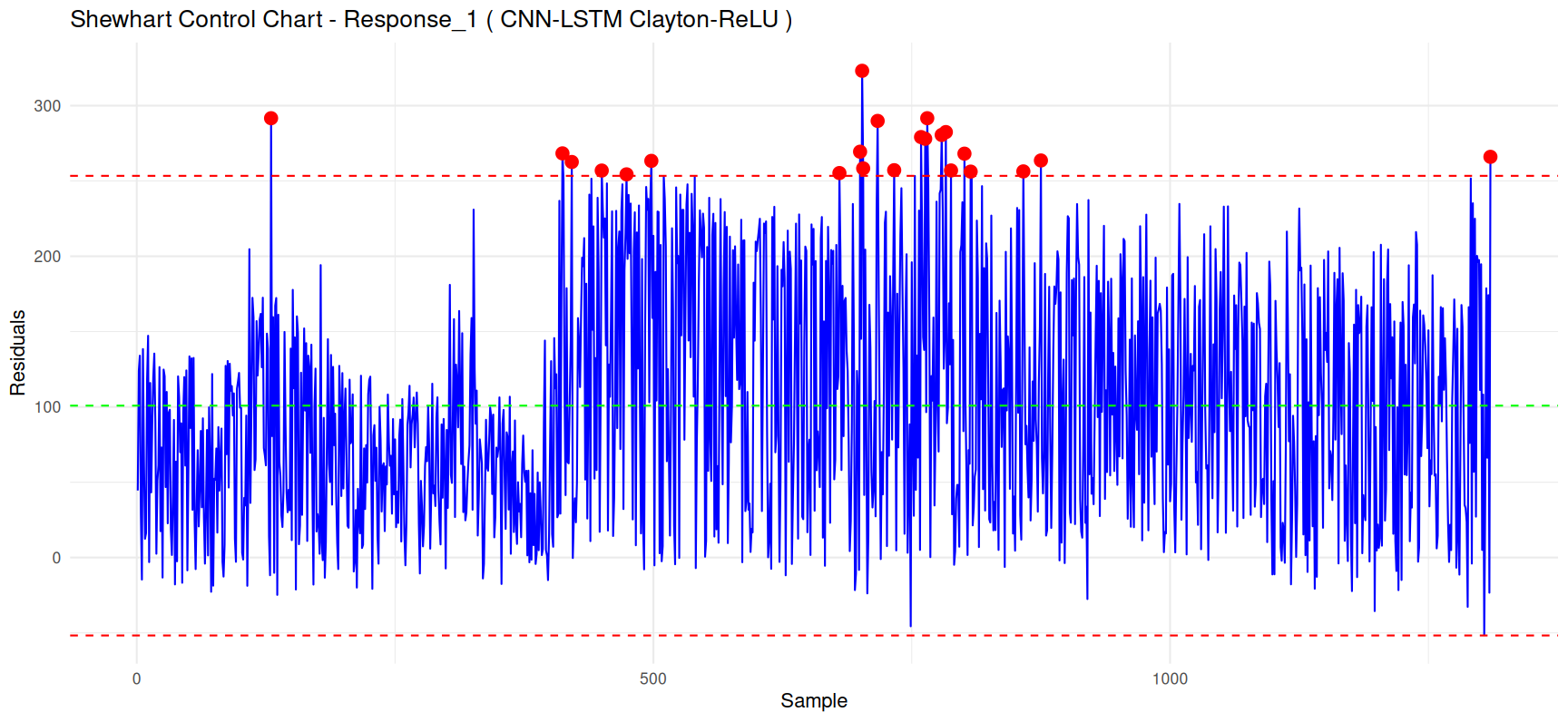}
        \subcaption{CNN-LSTM Clayton-ReLU $Y_1$}\label{fig:fig7}
    \end{minipage}%
    \begin{minipage}{0.35\textwidth}
        \centering
        \includegraphics[width=\linewidth]{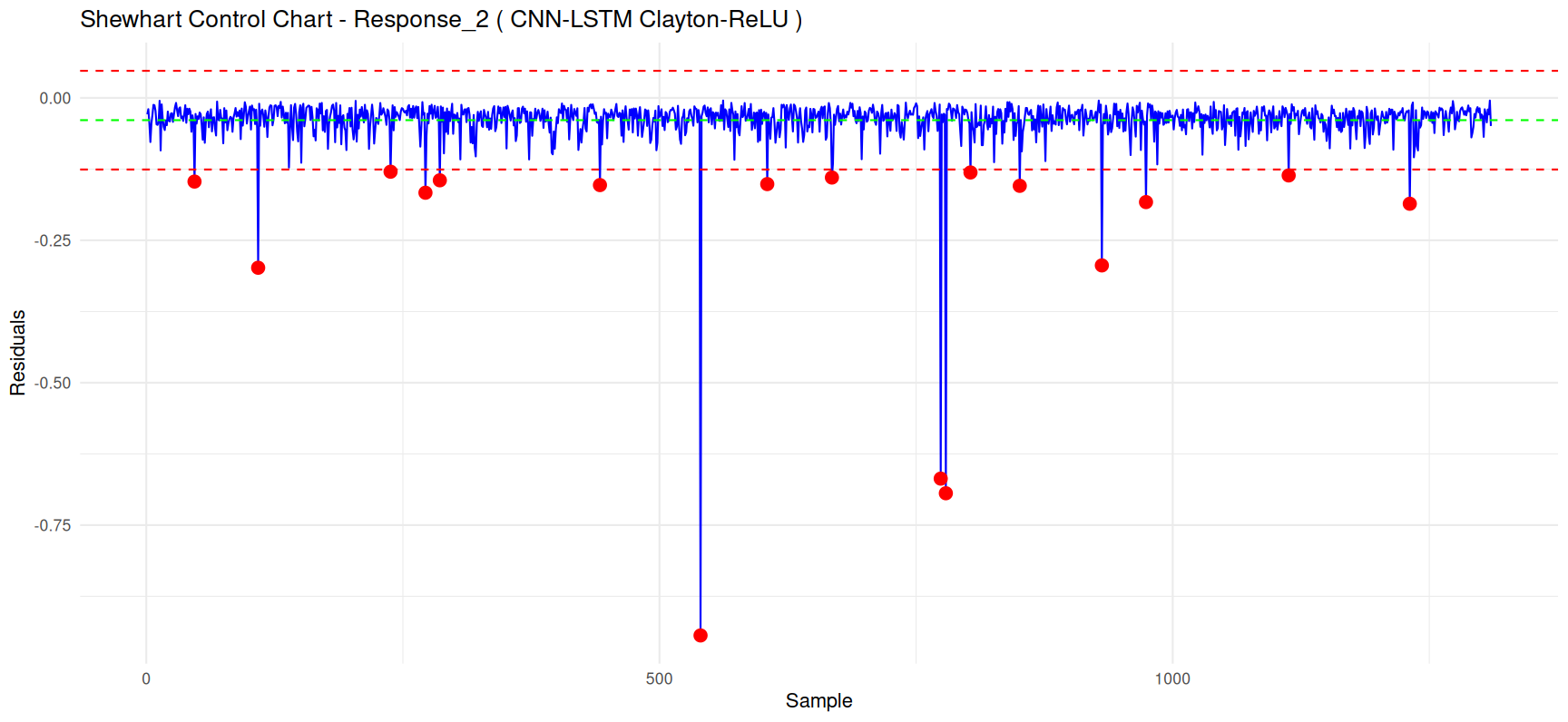}
        \subcaption{CNN-LSTM Clayton-ReLU $Y_2$}\label{fig:fig8}
    \end{minipage} \\[1ex]
    \begin{minipage}{0.35\textwidth}
        \centering
        \includegraphics[width=\linewidth]{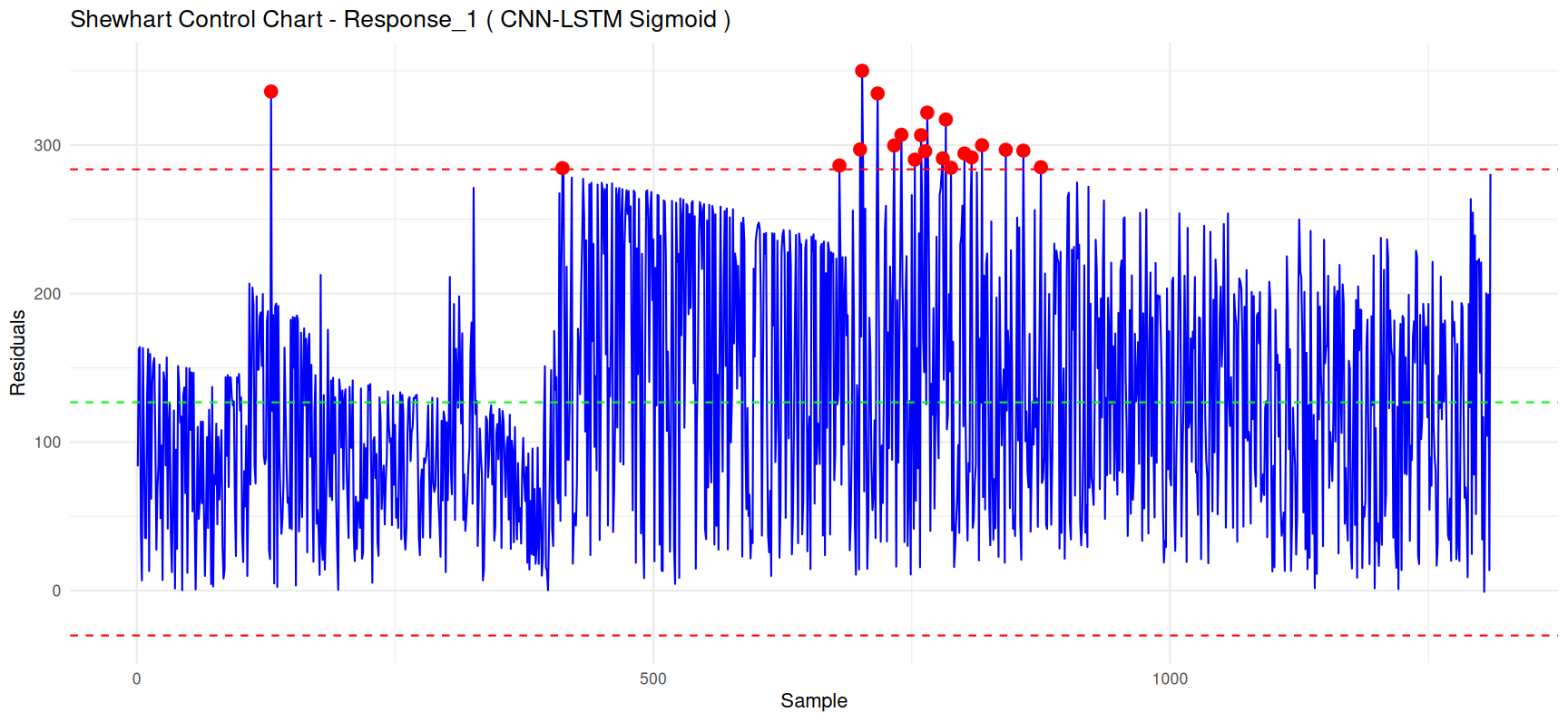}
        \subcaption{CNN-LSTM Sigmoid $Y_1$}\label{fig:fig10}
    \end{minipage}%
    \begin{minipage}{0.35\textwidth}
        \centering
        \includegraphics[width=\linewidth]{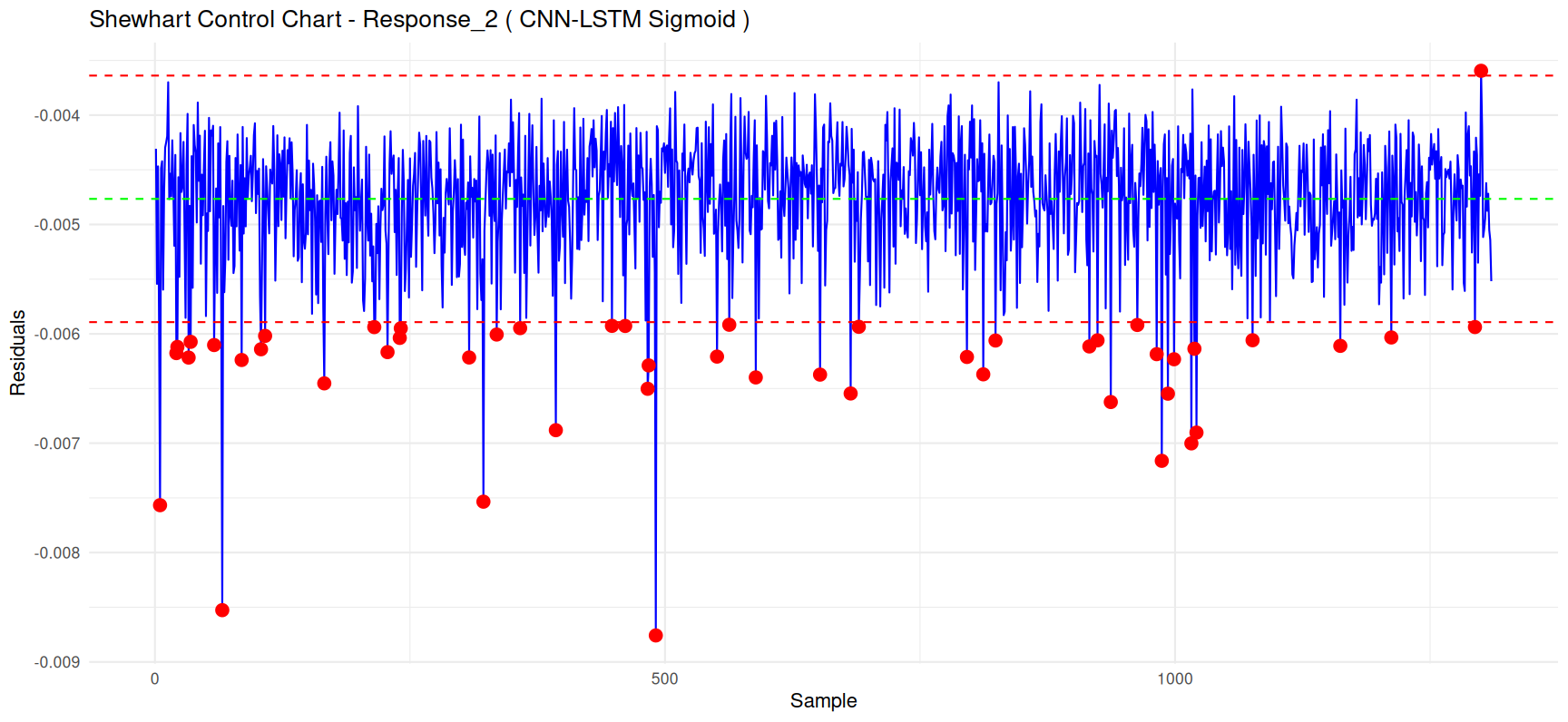}
        \subcaption{CNN-LSTM Sigmoid $Y_2$}\label{fig:fig11}
    \end{minipage} \\[1ex]
    \begin{minipage}{0.35\textwidth}
        \centering
        \includegraphics[width=\linewidth]{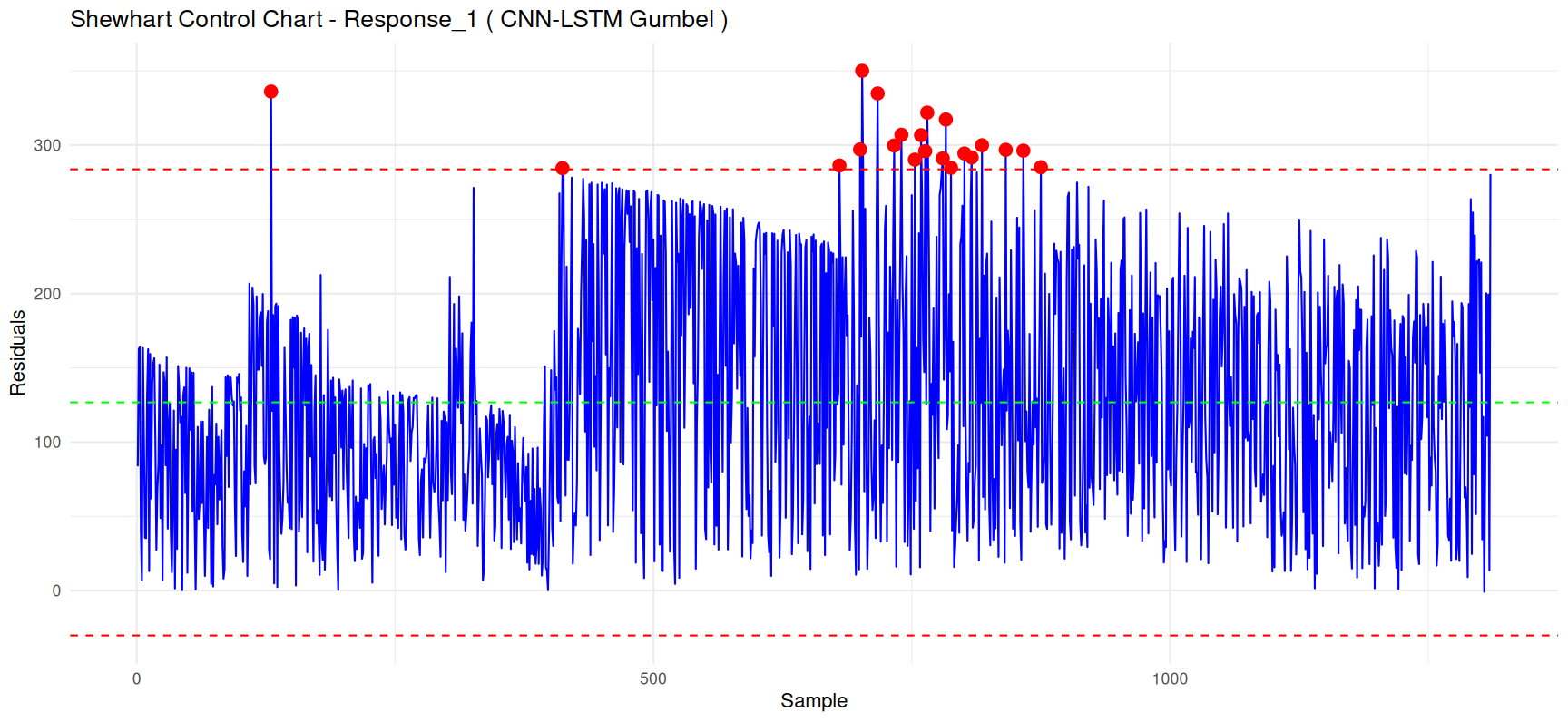}
        \subcaption{CNN-LSTM Gumbel $Y_1$}\label{fig:fig10}
    \end{minipage}%
    \begin{minipage}{0.35\textwidth}
        \centering
        \includegraphics[width=\linewidth]{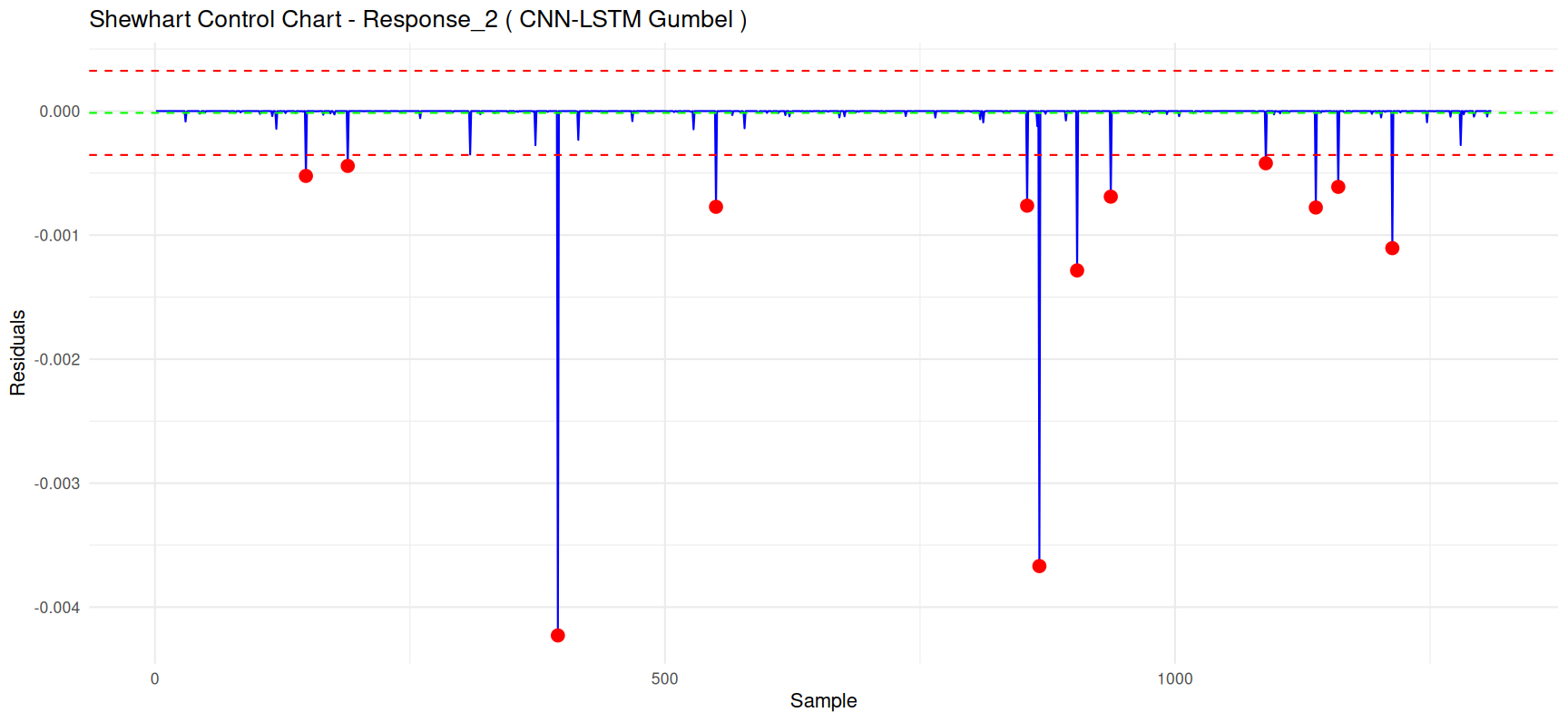}
        \subcaption{CNN-LSTM Gumbel $Y_2$}\label{fig:fig11}
    \end{minipage} \\[1ex]
    \caption{Residual Shewhart Control Charts of CNN-LSTM Models with METABRIC Data.}
    \label{fig:allfigures3}
\end{figure}

\begin{figure}[htp]
    \centering
    \begin{minipage}{0.35\textwidth}
        \centering
        \includegraphics[width=\linewidth]{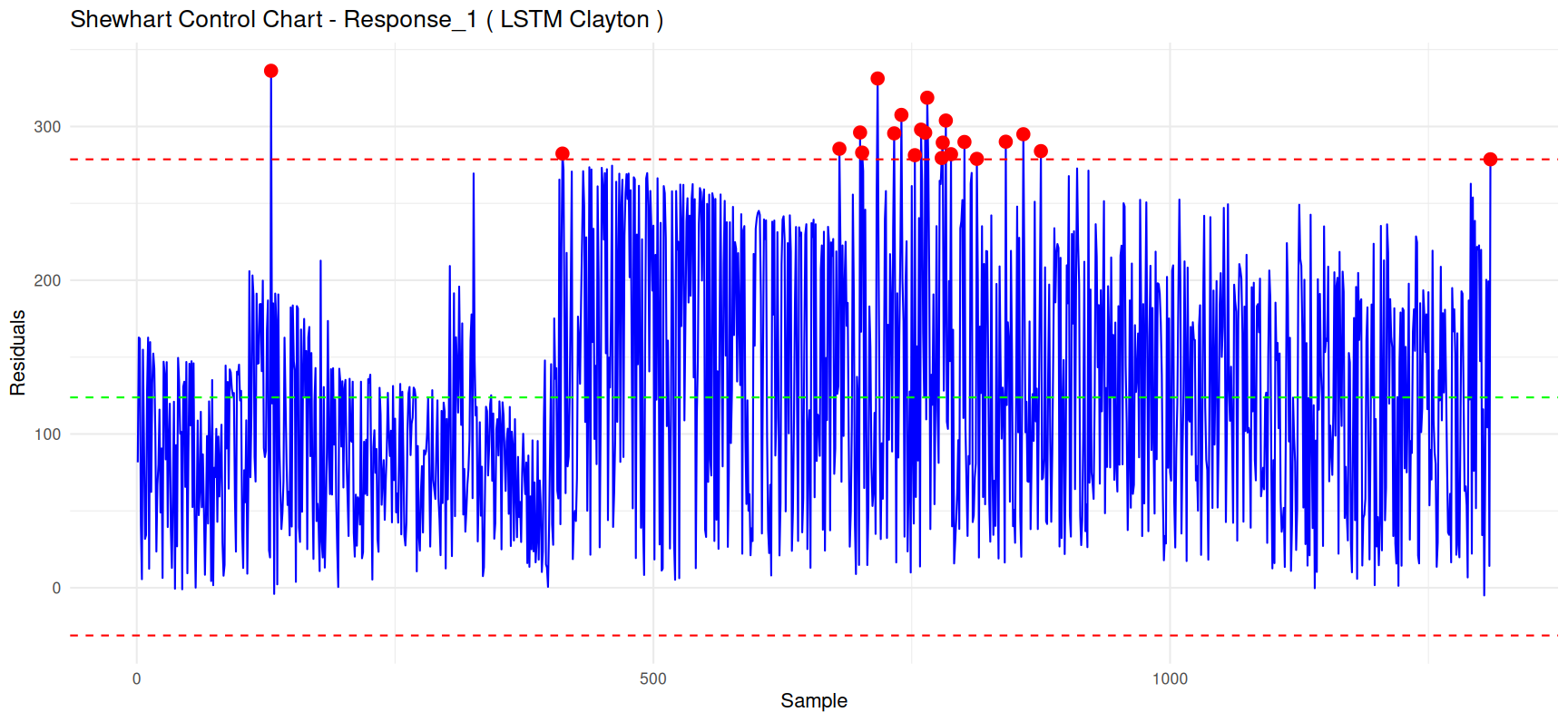}
        \subcaption{LSTM Clayton $Y_1$}\label{fig:fig1}
    \end{minipage}%
    \begin{minipage}{0.35\textwidth}
        \centering
        \includegraphics[width=\linewidth]{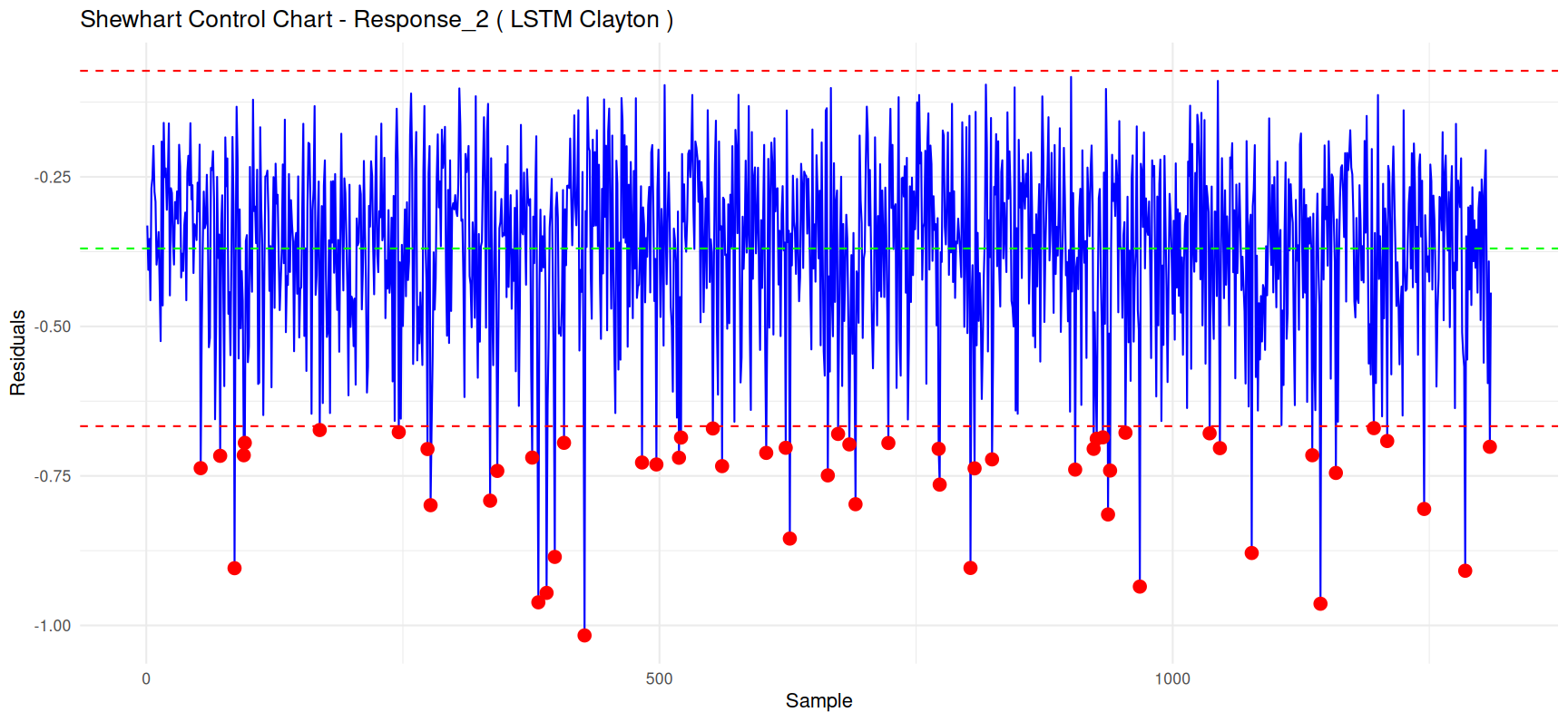}
        \subcaption{LSTM Clayton $Y_2$}\label{fig:fig2}
    \end{minipage} \\[1ex]
    \begin{minipage}{0.35\textwidth}
        \centering
        \includegraphics[width=\linewidth]{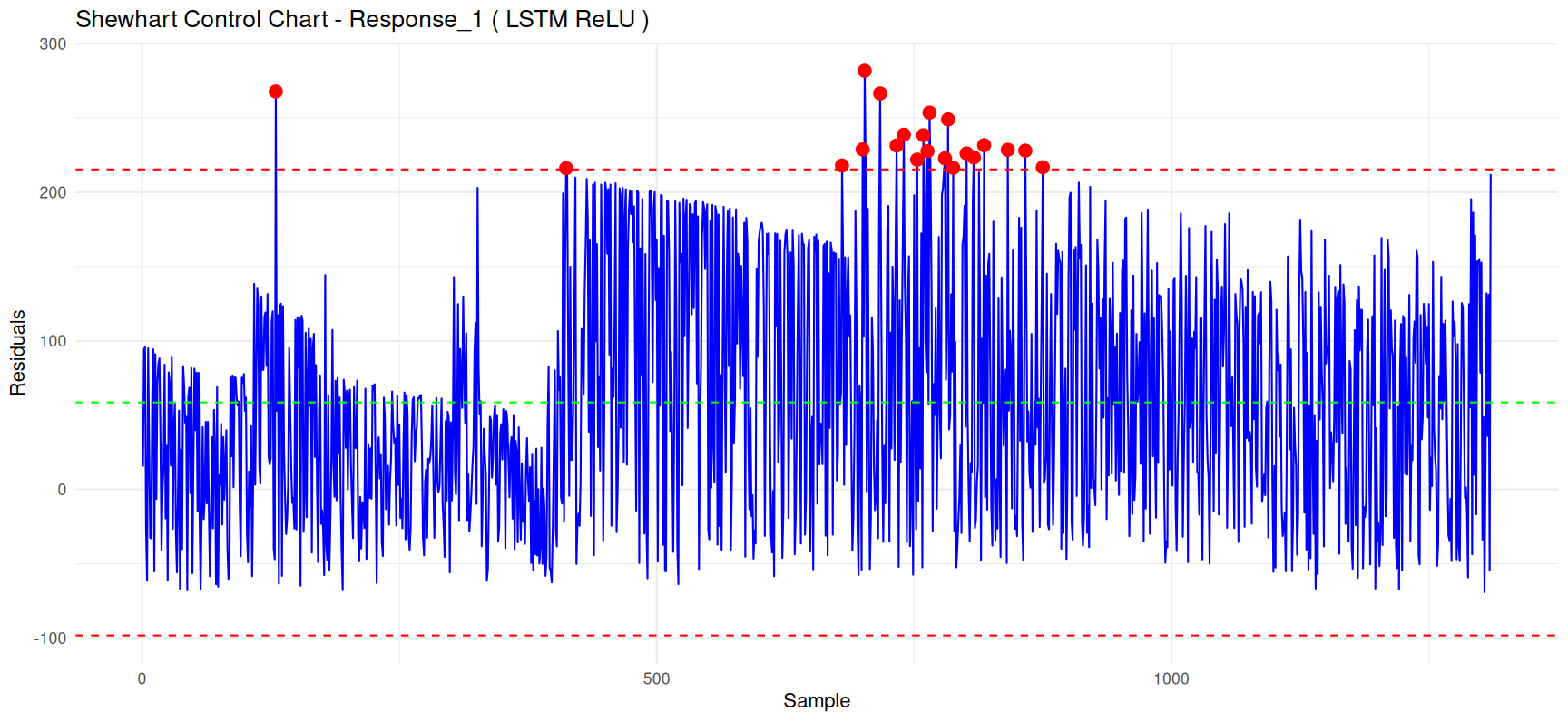}
        \subcaption{LSTM ReLU $Y_1$}\label{fig:fig4}
    \end{minipage}%
    \begin{minipage}{0.35\textwidth}
        \centering
        \includegraphics[width=\linewidth]{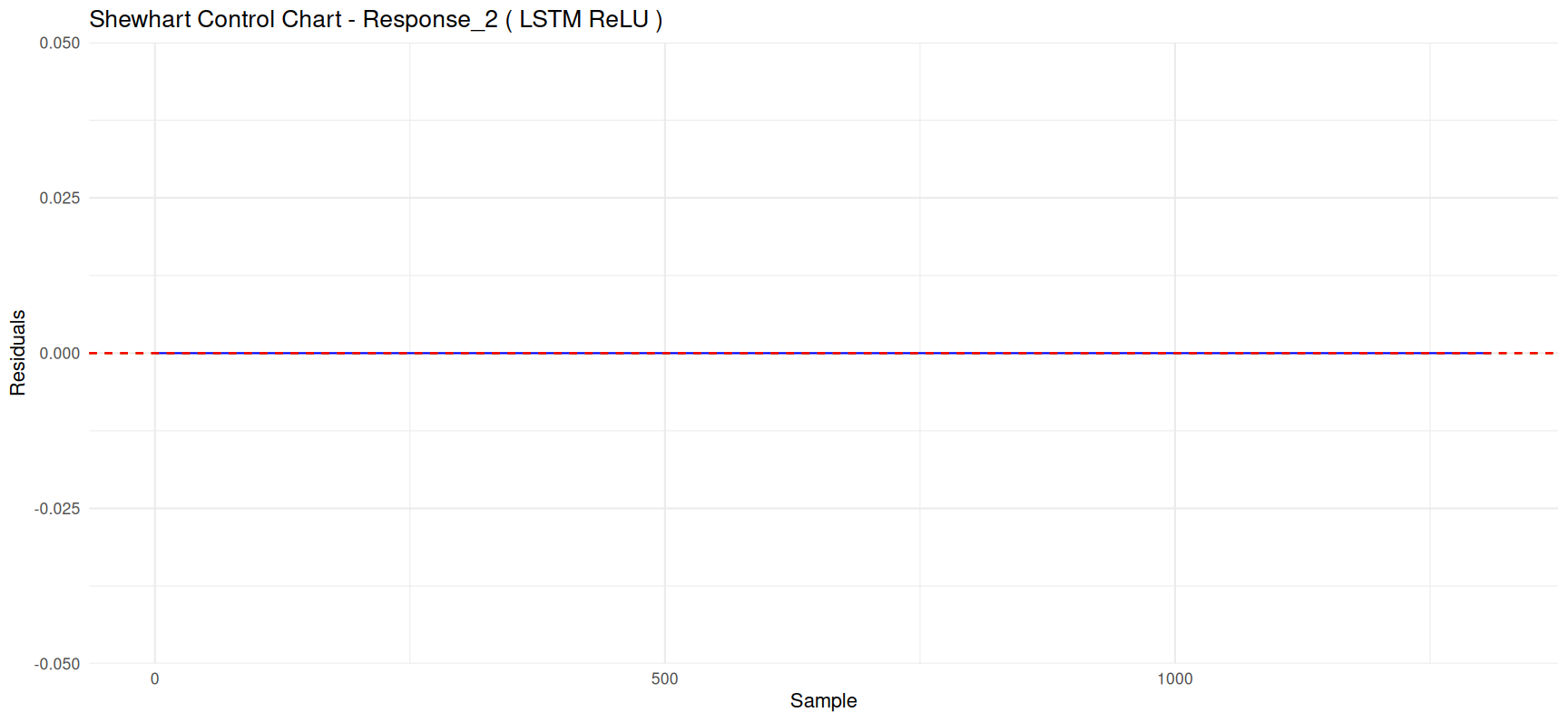}
        \subcaption{LSTM ReLU $Y_2$}\label{fig:fig5}
    \end{minipage}\\[1ex]
    \begin{minipage}{0.35\textwidth}
        \centering
        \includegraphics[width=\linewidth]{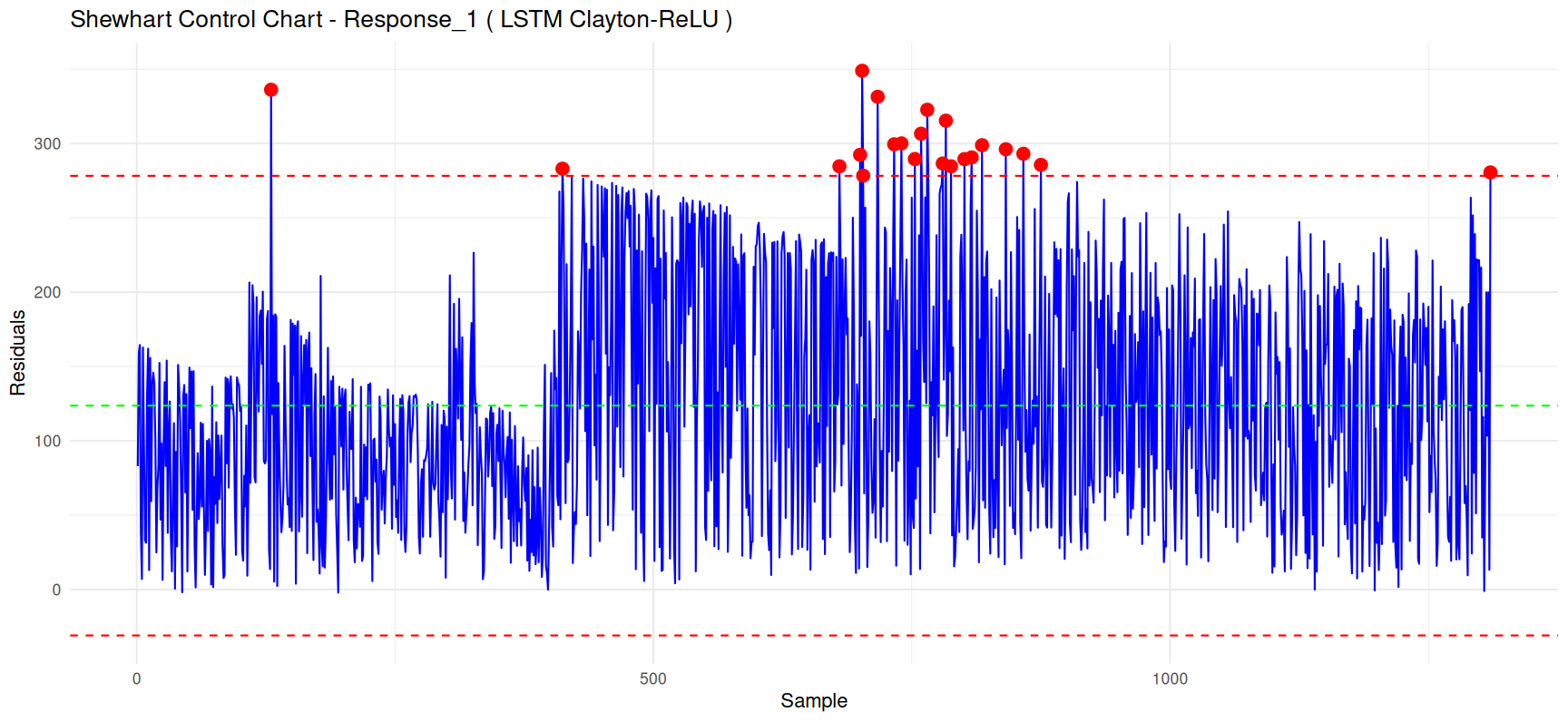}
        \subcaption{LSTM Clayton-ReLU $Y_1$}\label{fig:fig7}
    \end{minipage}%
    \begin{minipage}{0.35\textwidth}
        \centering
        \includegraphics[width=\linewidth]{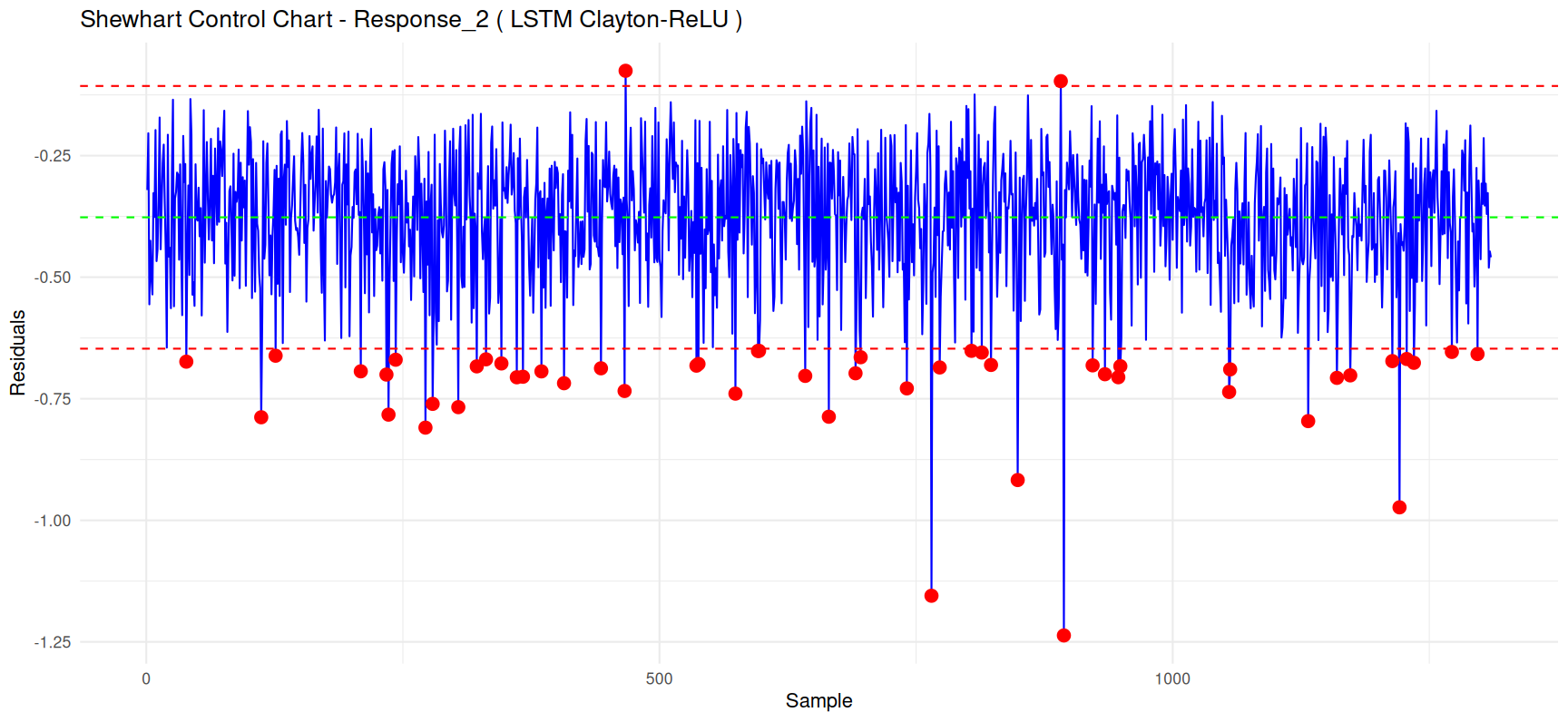}
        \subcaption{LSTM Clayton-ReLU $Y_2$}\label{fig:fig8}
    \end{minipage} \\[1ex]
    \begin{minipage}{0.35\textwidth}
        \centering
        \includegraphics[width=\linewidth]{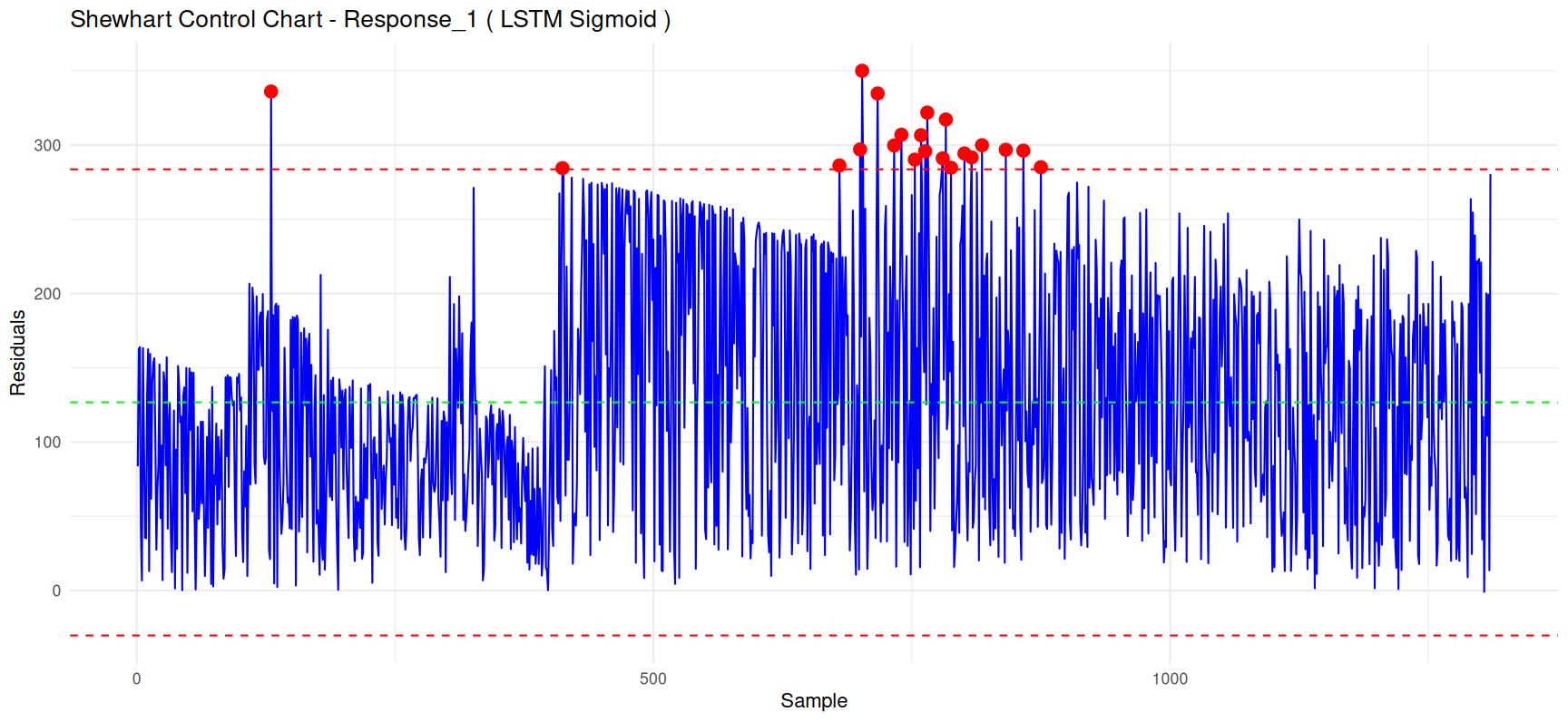}
        \subcaption{LSTM Sigmoid $Y_1$}\label{fig:fig10}
    \end{minipage}%
    \begin{minipage}{0.35\textwidth}
        \centering
        \includegraphics[width=\linewidth]{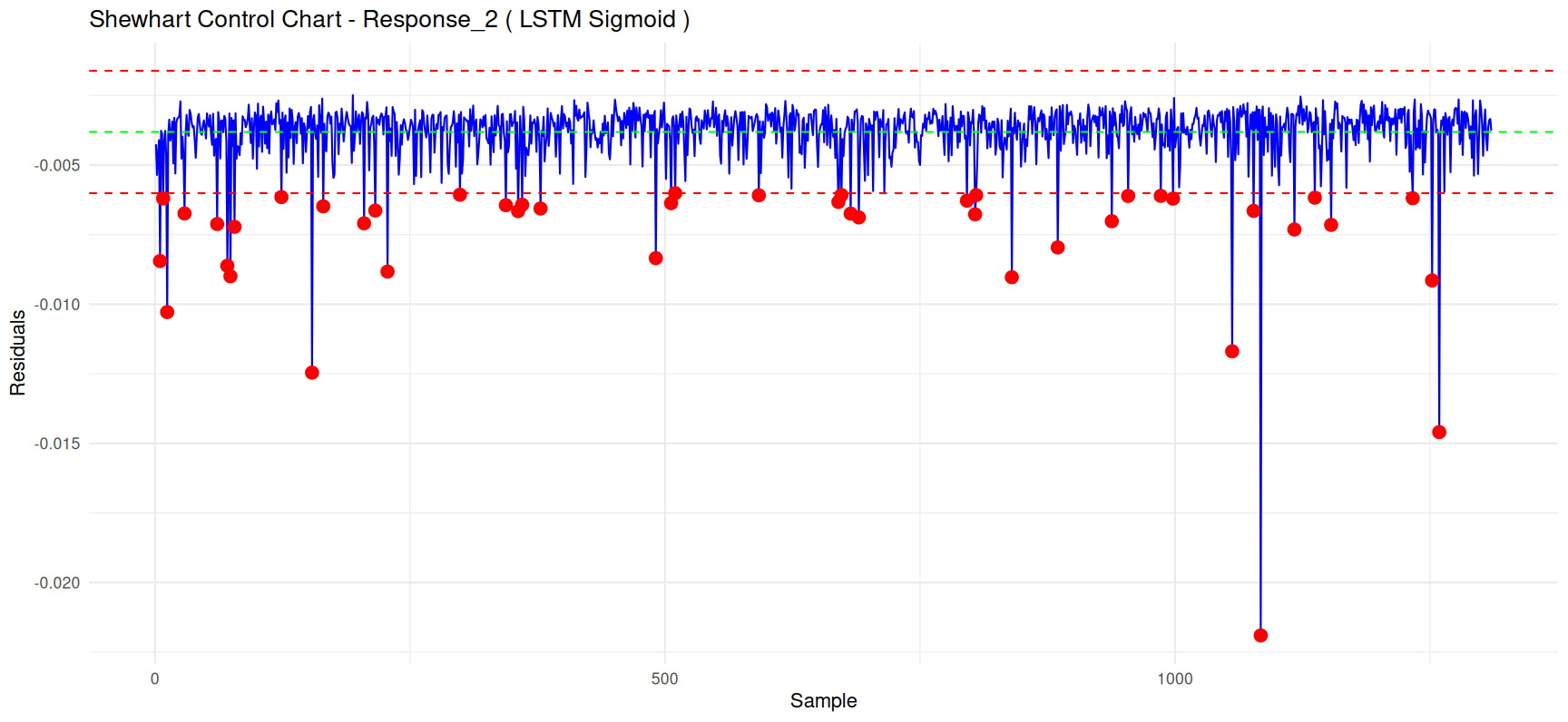}
        \subcaption{LSTM Sigmoid $Y_2$}\label{fig:fig11}
    \end{minipage} \\[1ex]
    \begin{minipage}{0.35\textwidth}
        \centering
        \includegraphics[width=\linewidth]{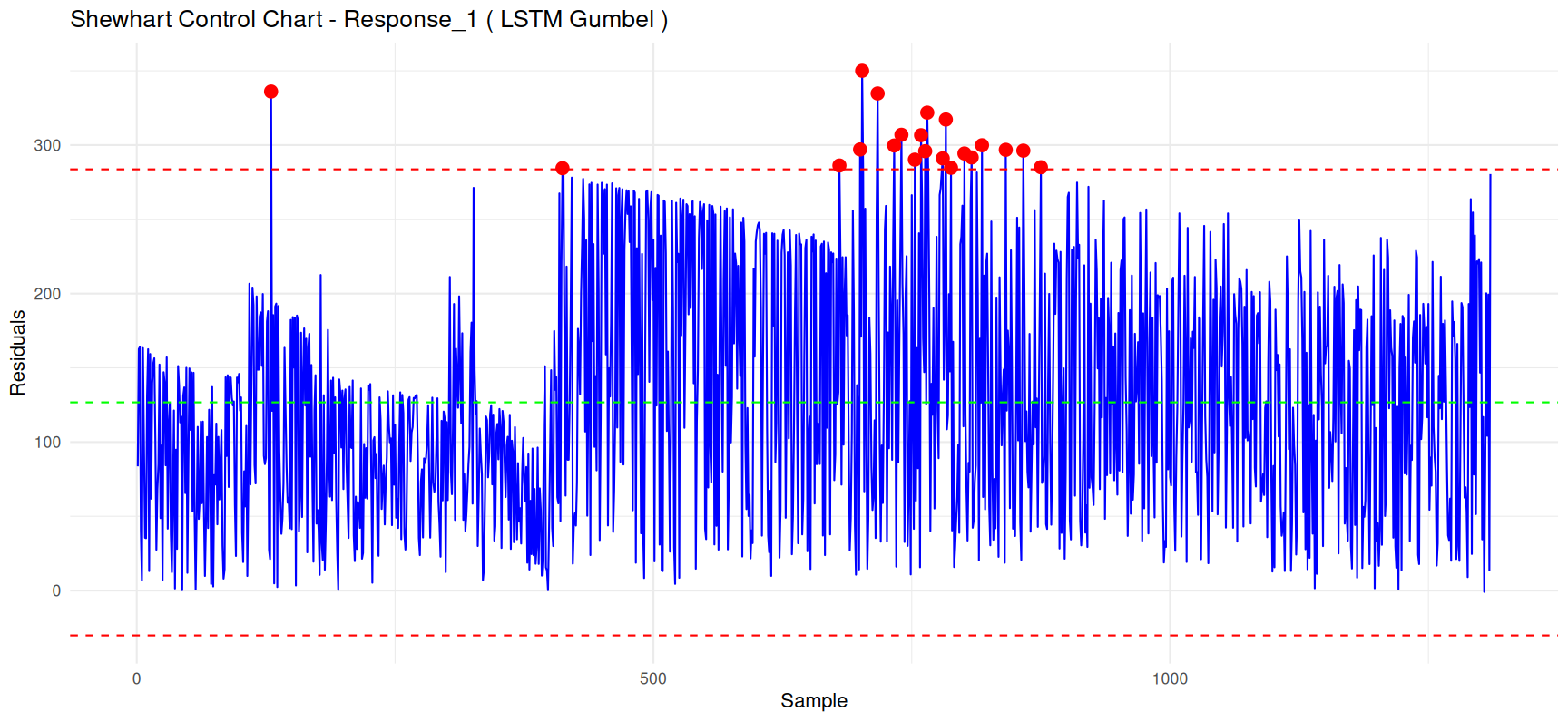}
        \subcaption{LSTM Gumbel $Y_1$}\label{fig:fig10}
    \end{minipage}%
    \begin{minipage}{0.35\textwidth}
        \centering
        \includegraphics[width=\linewidth]{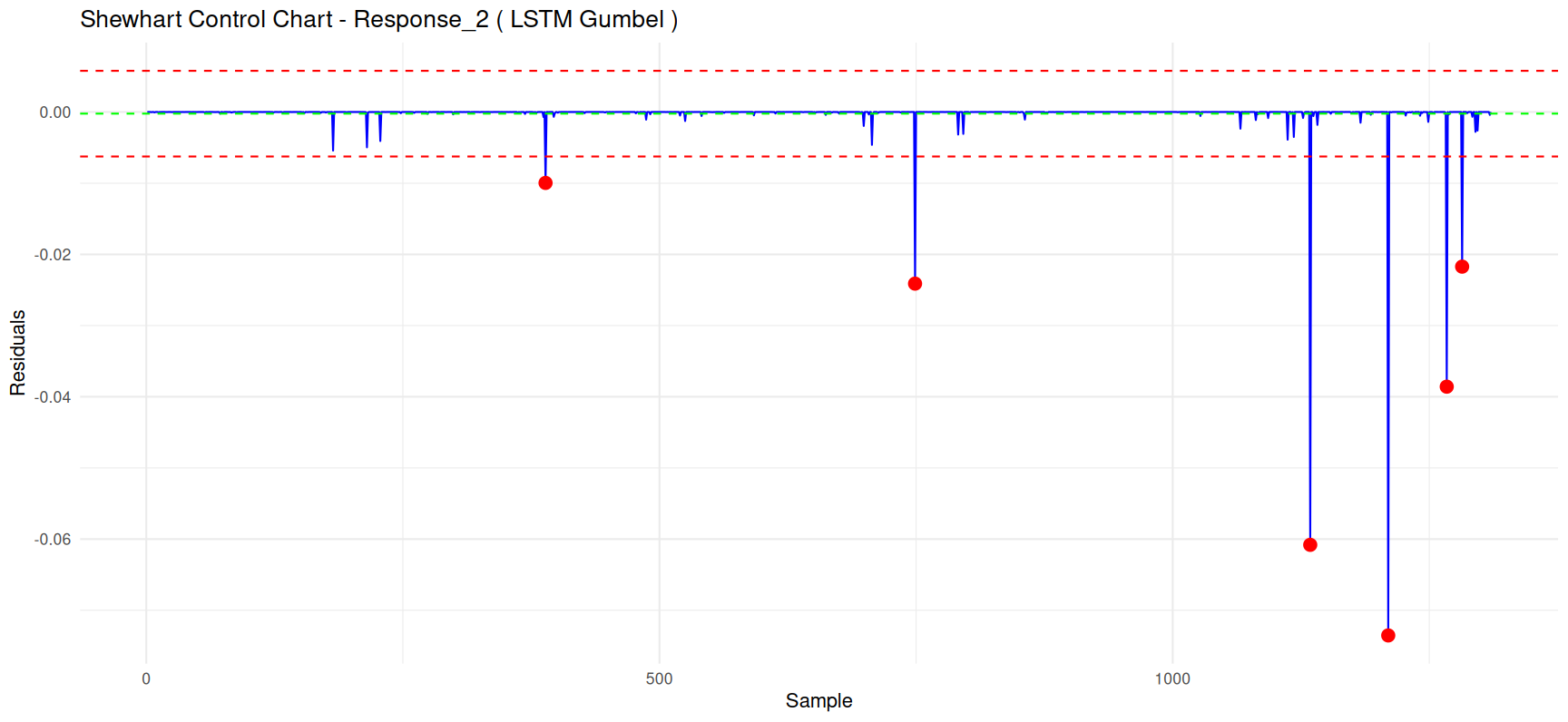}
        \subcaption{LSTM Gumbel $Y_2$}\label{fig:fig11}
    \end{minipage} \\[1ex]
    \caption{Residual Shewhart Control Charts of LSTM Models with METABRIC Data.}
    \label{fig:allfigures4}
\end{figure}

From Figures \ref{fig:allfigures3} and \ref{fig:allfigures4}, the
residual Shewhart control charts illustrate differing stability
levels across the CNN-LSTM and LSTM models. The CNN-LSTM and LSTM
models with the Clayton-ReLU activation function demonstrate
superior stability in the residuals for the binary variable
Response\_2, while the CNN-LSTM and LSTM models with ReLU, Sigmoid,
and Gumbel activation functions exhibit instability, particularly in
Response\_2. We can conclude that the CNN-LSTM model with
Clayton-based activation functions performs moderately well in
controlling residual fluctuations.

\begin{table}[]
\center     \caption{Model Comparison with METABRIC Data.}
 \label{table:3} \resizebox{1.0\textwidth}{!}{
\begin{tabular}{|c|c|c|c|c|c|}
\hline \textbf{Model}        & \textbf{Response} &
\textbf{Mean\_Residual} & \textbf{SD\_Residual} & \textbf{Mean\_ARL}
& \textbf{SD\_ARL} \\ \hline CNN-LSTM Clayton      & Response\_1 &
45.50912                & 65.87903              & 29.11111 & 3.51758
\\ \hline CNN-LSTM Clayton      & Response\_2 & -0.08897
& 0.03711               & 34.08571 & 3.51758          \\ \hline
CNN-LSTM ReLU         & Response\_1 & 63.42113                &
73.22987              & 48.51852 & 38.05282         \\ \hline
CNN-LSTM ReLU         & Response\_2 & -0.00037                &
0.00623               & 102.33333 & 38.05282         \\ \hline
CNN-LSTM Clayton-ReLU & Response\_1 & 100.82850               &
76.27779              & 56.95652 & 10.92851         \\ \hline
CNN-LSTM Clayton-ReLU & Response\_2 & -0.03900                &
0.04333               & 72.41176 & 10.92851         \\ \hline
CNN-LSTM Sigmoid      & Response\_1 & 126.72640               &
78.47299              & 41.66667 & 11.07801         \\ \hline
CNN-LSTM Sigmoid      & Response\_2 & -0.00477                &
0.00056               & 26.00000 & 11.07801         \\ \hline
CNN-LSTM Gumbel       & Response\_1 & 126.72640               &
78.47300              & 41.66667 & 42.01393         \\ \hline
CNN-LSTM Gumbel       & Response\_2 & -0.00001                &
0.00017               & 101.08333 & 42.01393         \\ \hline LSTM
Clayton          & Response\_1 & 123.79849               & 77.40638
& 59.54545 & 25.27585         \\ \hline LSTM Clayton          &
Response\_2 & -0.36971                & 0.14856               &
23.80000 & 25.27585         \\ \hline LSTM ReLU             &
Response\_1 & 58.5456                 & 78.3978               &
41.6667 & NA               \\ \hline LSTM ReLU             &
Response\_2 & 0                       & 0 & NA & NA               \\
\hline LSTM Clayton-ReLU     & Response\_1 & 123.6547 & 77.2923 &
59.5455 & 24.8009          \\ \hline LSTM Clayton-ReLU     &
Response\_2 & -0.3768                 & 0.1351                &
24.4717 & 24.8009
\\ \hline LSTM Sigmoid          & Response\_1 & 126.7264
& 78.4730               & 41.6667 & 9.6795           \\ \hline LSTM
Sigmoid          & Response\_2 & -0.0038                 & 0.0011 &
27.9778 & 9.6795           \\ \hline LSTM Gumbel           &
Response\_1 & 126.7264                & 78.4730               &
41.6667 & 121.6224         \\ \hline LSTM Gumbel           &
Response\_2 & -0.0002                 & 0.0030                &
213.6667 & 121.6224         \\ \hline
\end{tabular}
}
\end{table}

Table \ref{table:3} presents a comparison of CNN-LSTM and LSTM
models for breast cancer data using different activation functions
and copula-based dependency structures. The models are evaluated
based on four key metrics: mean residual, standard deviation of
residuals, mean ARL, and standard deviation of ARL. A lower mean
residual indicates better prediction accuracy, while a lower
standard deviation of residuals suggests more consistent
predictions. A higher mean ARL means fewer false alarms, and a lower
standard deviation of ARL indicates greater stability.

CNN-LSTM Clayton and CNN-LSTM ReLU exhibit moderate mean and
standard deviation residuals, with values of (45.51, 65.87) and
(63.42, 73.22), respectively. CNN-LSTM Sigmoid and CNN-LSTM Gumbel
show the highest residuals at (126.73, 78.47), indicating poor
prediction accuracy for Response\_1. LSTM models also exhibit high
residuals for Response\_1, with LSTM Clayton-ReLU at (123.65, 77.29)
and LSTM Gumbel at (126.73, 78.47), showing similar trends to
CNN-LSTM. Models using ReLU activation, both CNN-LSTM and LSTM, have
lower residuals than those using Sigmoid and Gumbel, making them
more stable.

For Response\_2, both CNN-LSTM and LSTM models generally perform
well, with mean residuals close to zero because the residual
Schwhart control charts does not work well in Figures
\ref{fig:allfigures3} and \ref{fig:allfigures4}. CNN-LSTM Clayton
and CNN-LSTM Clayton-ReLU have small residual values of (-0.08897,
0.03711) and (-0.03900, 0.04333), making them reliable choices for
stability. LSTM Gumbel and CNN-LSTM Gumbel display exceptionally low
residuals at (-0.0002, 0.0030) and (-0.00001, 0.00017),
respectively, indicating strong predictive performance because the
residual Schwhart control charts does not work well in Figures
\ref{fig:allfigures3} and \ref{fig:allfigures4}.

In terms of detecting changes, CNN-LSTM ReLU and CNN-LSTM Gumbel
achieve the highest mean ARL values for Response\_2, at 102.33 and
101.08, suggesting they are less likely to trigger false alarms.
CNN-LSTM Clayton, on the other hand, has lower ARL values for $Y_2$,
meaning it responds more frequently to changes.

Activation functions play a critical role in model performance.
Sigmoid activation results in higher residuals for $Y_1$, making it
less suitable for high-variance data. Clayton Copula improves $Y_2$
predictions.

Overall, CNN-LSTM models tend to have greater residual fluctuations
than LSTM models, particularly for Response\_1. Sigmoid and Gumbel
activations lead to higher instability in Response\_1, making them
less suitable for modeling this response.

\begin{figure}[htp]
    \centering
    \begin{minipage}{0.8\textwidth}
        \centering
        \includegraphics[width=\linewidth]{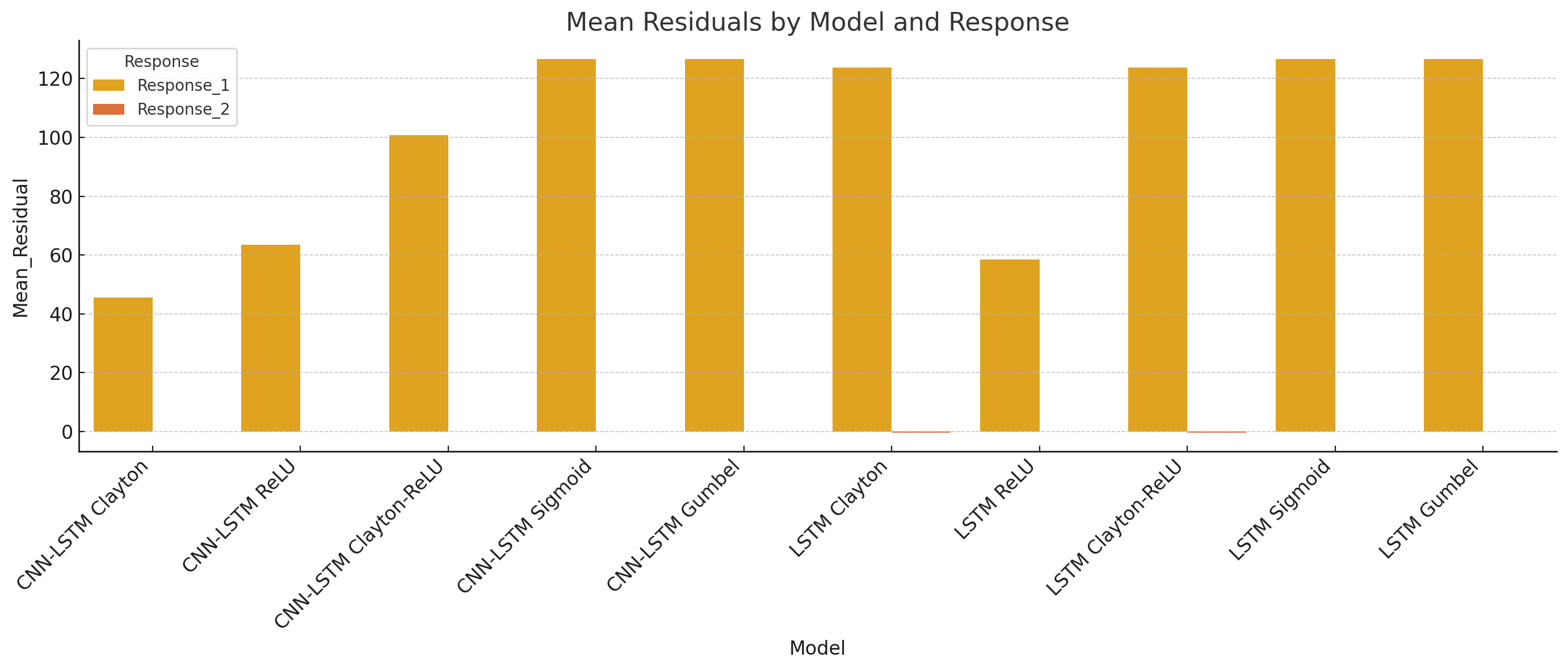}
    \end{minipage} \\[1ex]
    \begin{minipage}{0.8\textwidth}
        \centering
        \includegraphics[width=\linewidth]{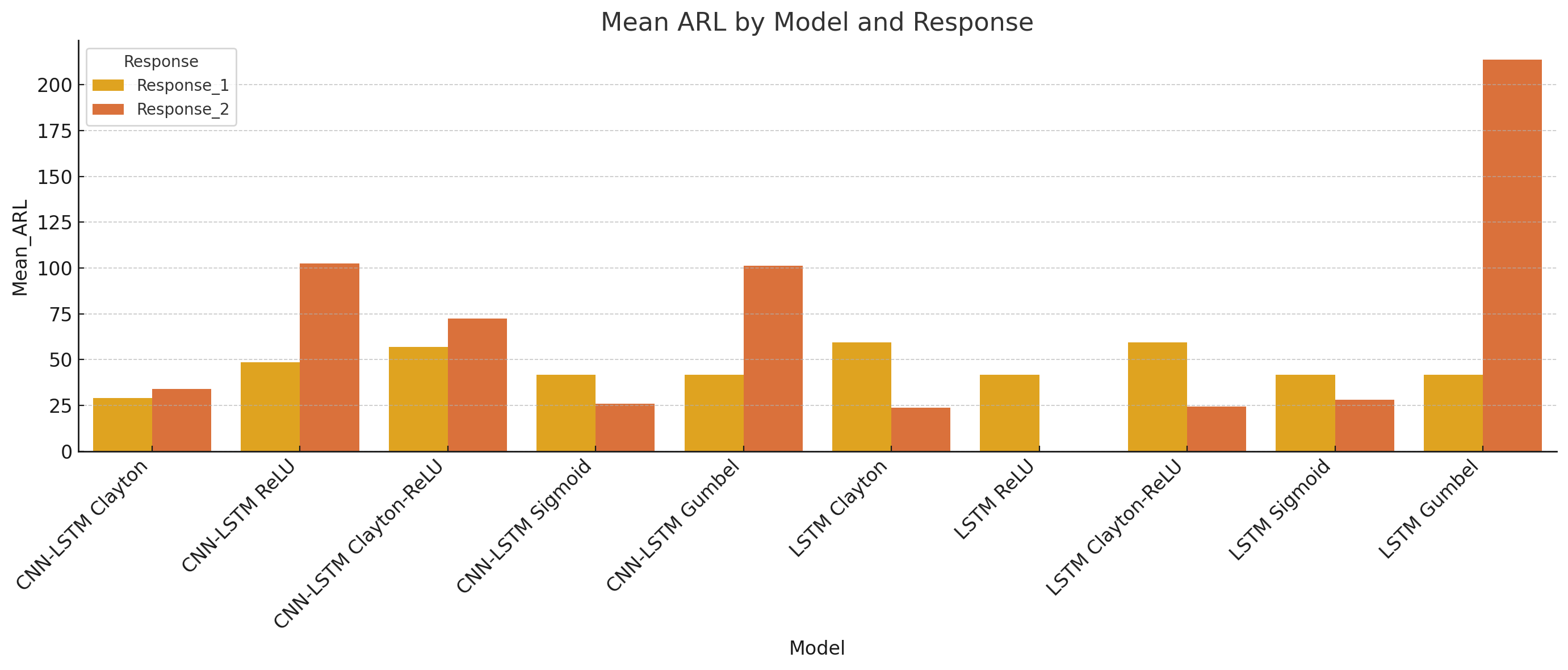}
    \end{minipage} \\[1ex]
    \caption{{Mean Residuals and Mean ARL by Model
and Response with METABRIC Data.}}
    \label{fig:allfigtable2}
\end{figure}

{Figure~\ref{fig:allfigtable2} summarizes mean residuals and
mean ARL by model and response with METABRIC Data. Mean residuals by
model and response shows how each model performs in terms of mean
residuals for Response\_1 and Response\_2. The mean ARLs by model
and response highlight how the models differ in their ARLs.}

\section{Conclusion}

In this research, we introduce copula-based activation functions
within CNN-LSTM survival analysis, demonstrating their effectiveness
in modeling complex dependencies in failure time data. This work pioneers
a hybrid framework that seamlessly integrates deep learning, copula
theory, and quality control methods, providing accurate predictions
while enhancing model reliability, interpretability, and robustness
for real-world applications. Our proposed method effectively handles
highly correlated, right-censored, multivariate survival response
data.

In conclusion, our framework overcomes the challenges posed by
censored data, captures intricate time dependencies, and models
dependency structures using copulas, offering distinct advantages
over traditional survival models. This approach holds significant
potential in fields such as medical research, reliability
engineering, and financial risk modeling, where precise survival
predictions and robust handling of censored data are essential. Our
future work will focus on competing-risks modeling for clustered
survival data, further extending the CNN-LSTM framework with
copula-based activation functions.

\section*{Acknowledgment}
This work was supported by the National Research Foundation of Korea
(NRF) grant funded by the Korea government(MSIT) (No.
NRF-2021R1G1A1094116 and No. RS-2023-00240794). We thank the two
respected referees, Associated Editor and Editor for constructive
and helpful suggestions which led to substantial improvement in the
revised version. {For the sake of transparency and
reproducibility, the R code for this study can be found in the
following GitHub repository:
\href{https://github.com/kjonomi/Rcode/blob/main/Copula_Based_Activation_Function}{R
code GitHub site}.}

\end{document}